\pdfoutput=1

\documentclass[10pt,letterpaper]{article}
\usepackage[top=0.85in,left=2.75in,footskip=0.75in]{geometry}

\usepackage{amsmath,amssymb}
\DeclareMathOperator{\atan}{atan}
\usepackage{stmaryrd}
\usepackage[ruled,vlined]{algorithm2e}

\usepackage{changepage}

\usepackage[utf8x]{inputenc}

\usepackage{textcomp,marvosym}

\usepackage{cite}

\usepackage{nameref}
\usepackage{siunitx}%
\newcommand{\ms}{\si{\milli\second}}%

\usepackage[right]{lineno}

\usepackage{makecell}
\usepackage{microtype}
\DisableLigatures[f]{encoding = *, family = * }

\usepackage[table]{xcolor}
\usepackage{multirow}

\usepackage{array}
\usepackage{tikz}
\usetikzlibrary{fit}

\newcolumntype{+}{!{\vrule width 2pt}}

\newcolumntype{L}[1]{>{\raggedright\let\newline\\\arraybackslash\hspace{0pt}}m{#1}}
\newcolumntype{C}[1]{>{\centering\let\newline\\\arraybackslash\hspace{0pt}}m{#1}}
\newcolumntype{R}[1]{>{\raggedleft\let\newline\\\arraybackslash\hspace{0pt}}m{#1}}
\newcolumntype{?}{!{\vrule width 1pt}}

\newcolumntype{V}[1]{>{\centering\arraybackslash} m{#1}}
\newlength\savedwidth

\raggedright
\usepackage[aboveskip=1pt,labelfont=bf,labelsep=period,justification=raggedright,singlelinecheck=off]{caption}

\makeatletter
\renewcommand{\@biblabel}[1]{\quad#1.}
\makeatother

\usepackage{lastpage,fancyhdr,graphicx}
\usepackage{epstopdf}
\fancyheadoffset[L]{2.25in}
\fancyfootoffset[L]{2.25in}
\makeatother
\def\mQuad{\hskip.61\fontdimen6\font}

\makeatletter
\newcommand{\algorithmfootnote}[2][\footnotesize]{%
  \let\old@algocf@finish\@algocf@finish
  \def\@algocf@finish{\old@algocf@finish
    \leavevmode\rlap{\begin{minipage}{\linewidth}
    #1#2
    \end{minipage}}%
  }%
}

\newcommand{\plotfig}{1}

\begin{document}
\vspace*{0.2in}

\begin{flushleft}
{\Large
\textbf\newline{
Sparse Deep Predictive Coding captures contour integration capabilities of the early visual system
}
}




Victor Boutin\textsuperscript{1,2*},
Angelo Franciosini\textsuperscript{1},
Frederic Chavane\textsuperscript{1},
Franck Ruffier\textsuperscript{2},
Laurent Perrinet\textsuperscript{1}
\\
\bigskip
\textbf{1} Aix Marseille Univ, CNRS, INT,  Inst Neurosci Timone, Marseille, France
\\
\textbf{2} Aix Marseille Univ, CNRS, ISM, Marseille, France
\\
\bigskip

%
%





* corresponding author: victor.boutin@univ-amu.fr

\end{flushleft}
\section*{Abstract}

Both neurophysiological and psychophysical experiments have pointed out the crucial role of recurrent and feedback connections to process context-dependent information in the early visual cortex. While numerous models have accounted for feedback effects at either neural or representational level, none of them were able to bind those two levels of analysis. Is it possible to describe feedback effects at both levels using the same model? We answer this question by combining Predictive Coding (PC) and Sparse Coding (SC) into a hierarchical and convolutional framework. In this Sparse Deep Predictive Coding (SDPC) model, the SC component models the internal recurrent processing within each layer, and the PC component describes the interactions between layers using feedforward and feedback connections. Here, we train a 2-layered SDPC on two different databases of images, and we interpret it as a model of the early visual system (V1~\&~V2). We first demonstrate that once the training has converged, SDPC exhibits oriented and localized receptive fields in V1 and more complex features in V2. Second, we analyze the effects of feedback on the neural organization beyond the classical receptive field of V1 neurons using interaction maps. These maps are similar to association fields and reflect the Gestalt principle of good continuation. We demonstrate that feedback signals reorganize interaction maps and modulate neural activity to promote contour integration. Third, we demonstrate at the representational level that the SDPC feedback connections are able to overcome noise in input images. Therefore, the SDPC captures the association field principle at the neural level which results in better disambiguation of blurred images at the representational level.

%
%
\section*{Author summary} 
One often compares biological vision to a camera-like system where an image would be processed according to a sequence of successive transformations. In particular, this ``feedforward'' view has been the building-block of popular architectures in deep-learning. However, neuroscientists have long stressed that more complex information flow is necessary to reach natural vision efficiency. In particular, recurrent and feedback connections in the visual cortex allow to integrate contextual information in our representation of visual stimuli. These modulations have been observed both at the low-level of neural activity and at the higher level of perception. In this study, we present an architecture that describes biological vision at both levels of analysis. It suggests that the brain uses feedforward and feedback connections to compare the sensory stimulus with its own internal representation. In contrast to classical deep learning approaches, our model learns interpretable features. Moreover, it demonstrates that feedback signals modulate neural activity to promote good continuity of contours. Finally, the same model can disambiguate images corrupted by noise. To the best of our knowledge, this is the first time that the same model describes the effect of recurrent and feedback modulations at both neural and representational levels.

\section*{Introduction}
Visual processing of objects and textures has been traditionally described as a pure feedforward process that extracts local features. These features become increasingly more complex and task-specific along the hierarchy of the ventral visual pathway~\cite{riesenhuber1999hierarchical, dicarlo2012does}. This view is supported by the very short latency of evoked activity observed in monkeys ($\approx~90~\ms$) in higher-order visual areas~\cite{bullier2001integrated, lamme2000distinct}. This feed-forward flow of information is sufficient to account for object categorization in the IT cortical area~\cite{fabre1998rapid}. Although this feedforward view of the visual cortex was able to account for a large scope of electrophysiological~\cite{freedman2003comparison, rust2006mt} and psychophysical~\cite{serre2007robust} findings, it does not take advantage of the high density ($\approx 20\%$) and diversity of feedback connections observed in the anatomy~\cite{felleman1991distributed, sporns2004small, markov2010weight}.

Feedback connections, but also horizontal intra-cortical connections are known to integrate contextual modulations in the early visual cortex~\cite{gilbert2013top,roelfsema2006cortical, muller2018cortical}. At the neurophysiological level, it was observed that activity in the 'silent' regions surrounding the classical Receptive Field (RF) tends to either suppress or facilitate the neural activity in the center of the RF. These so-called 'Center/Surround' modulations are known to be highly stimulus specific~\cite{series2003silent}. For example, when gratings are presented to the visual system, feedback signals tend to suppress horizontal connectivity which is thought to better segregate the shape of the perceived object from the ground (figure-ground segregation)~\cite{nurminen2018top, nassi2013corticocortical}. In contrast, when co-linear and co-oriented lines are presented, feedback signals facilitate horizontal connections such that local edges are grouped towards better shape coherence (contour integration)~\cite{liang2017interactions}. Interestingly, both figure-ground segregation and contour integration are directly derived from the Gestalt principle of perception. In particular, contour integration is known to follow the Gestalt rule of good continuation as mathematically formalized by the concept of association field~\cite{field1993contour}. This association field suggests that local edges tend to align toward a co-circular geometry. Besides being central in natural image organization~\cite{geisler2001edge}, association fields might also be implemented in the connectivity within the V1 area~\cite{gilbert2012adult, gerard2016synaptic} and play a crucial role in contour perception~\cite{field1993contour, polat1993lateral}. At the psychophysical level, the temporal decoupling between feedforward and feedback connections allowed experimenters to investigate the effect of feedback connection~\cite{boehler2008rapid, fahrenfort2008spatiotemporal}. In particular, it was demonstrated that short-range feedback connections (originating in the ventral visual area and targeting V1) play a crucial role in the recognition of degraded images~\cite{wyatte2012limits}. These pieces of biological evidence suggest that feedforward models are not sufficient to account for the context-dependent behavior of the early visual cortex and urge us to look for more complex, recurrent models.

From a neurophysiological perspective, Predictive Coding (PC) might be a good candidate to describe inter-layer interaction between feedforward and feedback connections. Rao \& Ballard~\cite{rao1999predictive} were the first to leverage Predictive Coding (PC) into a hierarchical framework and to combine it with Sparse Coding. On one hand, PC describes the brain as a  Bayesian 'machine' that consistently updates its internal model of the world to infer the possible physical causes of a given sensory input~\cite{knill2004bayesian}. PC suggests that top-down connections convey predictions about the activity in the lower level while bottom-up processes transmit prediction error to the next higher level. In particular, PC models were able to describe center-surround antagonism in the retina~\cite{srinivasan1982predictive} and extra-classical RFs effects observed in the early visual cortex~\cite{rao1999predictive}. In addition, studies have investigated the correspondence between cortical micro-circuitry and the connectivity implied by the PC theory~\cite{bastos2012canonical, shipp2016neural}. Later, PC has been generalized into a unified brain theory for perception and action~\cite{friston2008hierarchical, friston2010free,lee2003hierarchical}.
On the other hand, Sparse Coding (SC) might be considered as a framework to describe local computations in the early visual cortex. Olshausen \& Field demonstrated that a SC strategy was sufficient to account for the emergence of features similar to the Receptive Fields (RFs) of simple cells in the mammalian primary visual cortex~\cite{olshausen1997sparse}. These RFs are spatially localized, oriented band-pass filters~\cite{hubel1968receptive}.  Furthermore, SC could also be considered as a result of a competition mechanism that explains sensory input in terms of a small number of possible causes. SC implements an 'explaining away' strategy~\cite{spratling2012unsupervised} by suppressing alternative explanations and selecting only the dominant causes.

At the higher representational level, several models have accounted for feedback effects in psychophysical experiments. O'Reilly and colleagues presented a model, called LVis, to describe recurrent processing during object recognition~\cite{o2013recurrent}. The LVis model was used to describe the results of psychophysical experiments in which subjects had to recognize objects degraded by occlusion or contrast reduction under a backward-masking setting~\cite{wyatte2012limits}. Interestingly, the LVis model fitted the psychophysical findings: feedback connections greatly improved object recognition in degraded images and exhibited no significant improvement in classification performance when images of objects were not degraded~\cite{wyatte2012limits, wyatte2014early}. Another study uses recurrent convolutional neural networks to dissociate the role of feedforward, feedback, and horizontal connectivity in different kinds of degraded object recognition tasks (cluttered and noisy stimuli)~\cite{spoerer2017recurrent}. The authors demonstrated that while horizontal and feedforward connectivity were enough to account for good classification under modest degradation of stimuli, the feedback connection was necessary to reach higher performance for heavily degraded inputs.

To the best of our knowledge, there is no model accounting for feedback effect at both neural level and representational level. In this paper, we use a Sparse Deep Predictive Coding (SDPC) model that combines Predictive Coding and Sparse Coding in a hierarchical and convolutional framework. We first briefly introduce the 2-layered SDPC network used to conduct all the experiments of the paper, and we show the results of the training of the SDPC on two different databases. Next, we investigate the feedback effects at the 'neural' level. We show how feedback signals in SDPC account for a reshaping of V1 neural population both in terms of topographic organization and activity level. Then, we probe the effect of feedback at the representational level. In particular, we investigate the ability of the feedback connection to denoise corrupted input. Finally, we discuss the results obtained with the SDPC model in the light of the psychophysical and neurophysiological findings observed in neuroscience.

\section*{Results}
In our mathematical description of the proposed model, italic letters are used as symbols for \textit{scalars}, bold lowercase letters for column $\boldsymbol{vectors}$ and bold uppercase letters for \textbf{MATRICES}. $\textit{j}$ refers to the complex number such that $\textit{j}^2 = -1$.
\subsection*{Brief description of the SDPC}
Given a hierarchical generative model for the formation of images, the core objective of Hierarchical Sparse Coding (HSC) is to retrieve the parameters and the internal states variables that best explain the input stimulus. As any perceptual inference model, HSC attempts to solve an inverse problem (Eq.~\ref{eq:eq0}), where the forward model is a hierarchical linear model:
 \begin{equation}
    \label{eq:eq0}
    \left\{
    \begin{array}{lll}
        \boldsymbol{x} = \mathbf{D}_{1}^{T}\boldsymbol{\gamma}_{1} + \boldsymbol{\epsilon}_{1} & \textnormal{s.t.} \mQuad \Vert \boldsymbol{\gamma}_{1} \Vert_{0}< \alpha_{1} & \textnormal{and} \mQuad  \boldsymbol{\gamma}_{1}>0 \\
    \boldsymbol{\gamma}_{1} = \mathbf{D}_{2}^{T}\boldsymbol{\gamma}_{2} + \boldsymbol{\epsilon}_{2} & \textnormal{s.t.} \mQuad \Vert \boldsymbol{\gamma}_{2} \Vert_{0}< \alpha_{2} & \textnormal{and} \mQuad  \boldsymbol{\gamma}_{2}>0 \\
    ..\\
    \boldsymbol{\gamma}_{L-1} = \mathbf{D}_{L}^{T}\boldsymbol{\gamma}_{L} + \boldsymbol{\epsilon}_{L} & \textnormal{s.t.} \mQuad \Vert \boldsymbol{\gamma}_{L} \Vert_{0}< \alpha_{L} & \textnormal{and} \mQuad  \boldsymbol{\gamma}_{L}>0
     \end{array}
     \right.
\end{equation}
The number of layers of our model is denoted $L$ and $\boldsymbol{x}$ is the sensory input (i.e. image).
The sparsity at each layer is enforced by a constraint on the $\ell_{0}$ pseudo-norm of the internal state variable $\boldsymbol{\gamma}_{i}$. Finally, $\boldsymbol{\epsilon}_{i}$  and $\mathbf{D}_{i}$ are respectively the prediction error and the weights (i.e. the parameters) at each layer $i$.

To tighten the link with neuroscience, we impose $\boldsymbol{\gamma}_{i}$ to be non-negative and we force $\mathbf{D}_{i}$ to have a convolutional structure. It allows us to interpret $\boldsymbol{\gamma}_{i}$ as a retinotopic map describing the neural activity at layer $i$. In addition, $\mathbf{D}_{i}$ could be viewed as the synaptic weights between $2$ layers whose activity is represented by $\boldsymbol{\gamma}_{i-1}$ and $\boldsymbol{\gamma}_{i}$. When projected into the visual space (see Eq.~\ref{eq:Effective_Dictionaries}), dictionaries could also be interpreted as a set of Receptive Fields (RFs). We call  $\mathbf{D}_{i}^\mathrm{eff}$ the back-projection of $\mathbf{D}_{i}$ into the visual space (see Fig.~\ref{fig:fig_backprojection}). Note that RFs in the visual space get bigger for neurons located in deeper layers (that is, on layers further away from the sensory layer). To visualize the information represented by each layer, we back-project $\boldsymbol{\gamma}_{i}$ into the visual space (see Eq.~\ref{eq:Reconstruction}). We call this projection a 'representation' and it is denoted by $\boldsymbol{\gamma}_{i}^\mathrm{eff}$.

One possibility to solve the problem defined by Eq.~\ref{eq:eq0} in a neuro-plausible way is to use the Sparse Deep Predictive Coding (SDPC) model (see~\nameref{Sec:Methods}). The SDPC model combines local computational mechanisms to learn the weights and infer internal state variables. It leverages recurrent and bi-directional connections (feedback and feedforward) through the Predictive Coding (PC) theory. SDPC consists in minimizing, for each layer the loss function defined by Eq.~\ref{eq:eq1} :
 \begin{equation}
    \label{eq:eq1}
    \mathcal{L}  =  \frac{\displaystyle 1}{\displaystyle 2} \Vert \boldsymbol{\gamma}_{i-1} - \mathbf{D}_{i}^{T}\boldsymbol{\gamma}_{i} \Vert_{2}^{2} + \frac{\displaystyle k_\mathrm{FB}}{\displaystyle 2} \Vert \boldsymbol{\gamma}_{i} - \mathbf{D}_{i+1}^{T}\boldsymbol{\gamma}_{i+1} \Vert_{2}^{2} + \lambda_{i} \Vert \boldsymbol{\gamma}_{i} \Vert_{1}
\end{equation}
In Eq.~\ref{eq:eq1}, $k_\mathrm{FB}$ is a parameter we introduce to tune the strength of the feedback connection and $\lambda_{i}$ controls the sparsity level within each layer.

In this paper, we aim at modeling the early visual cortex using a $2$-layered version of SDPC (see Fig.~\ref{fig:fig_SDPC}). Consequently, we denote the first and second layer of the SDPC as the V1 and V2 model, respectively. In particular, $\boldsymbol{\gamma}_{1}$ corresponds to the activity-map in V1 and $\boldsymbol{\gamma}_{2}$ to V2's activity-map. We refer to the V1 space as the retinotopic space described by $\boldsymbol{\gamma}_{1}$, and it is symbolized with a small coordinate system centered in $O_{V1}$ (see Fig.~\ref{fig:fig_SDPC}).

\ifnum \plotfig=1
\begin{figure}
	\centering
\begin{tikzpicture}
\draw [anchor=north west] (0.05\linewidth, 1\linewidth) node {\includegraphics[width=1\linewidth]{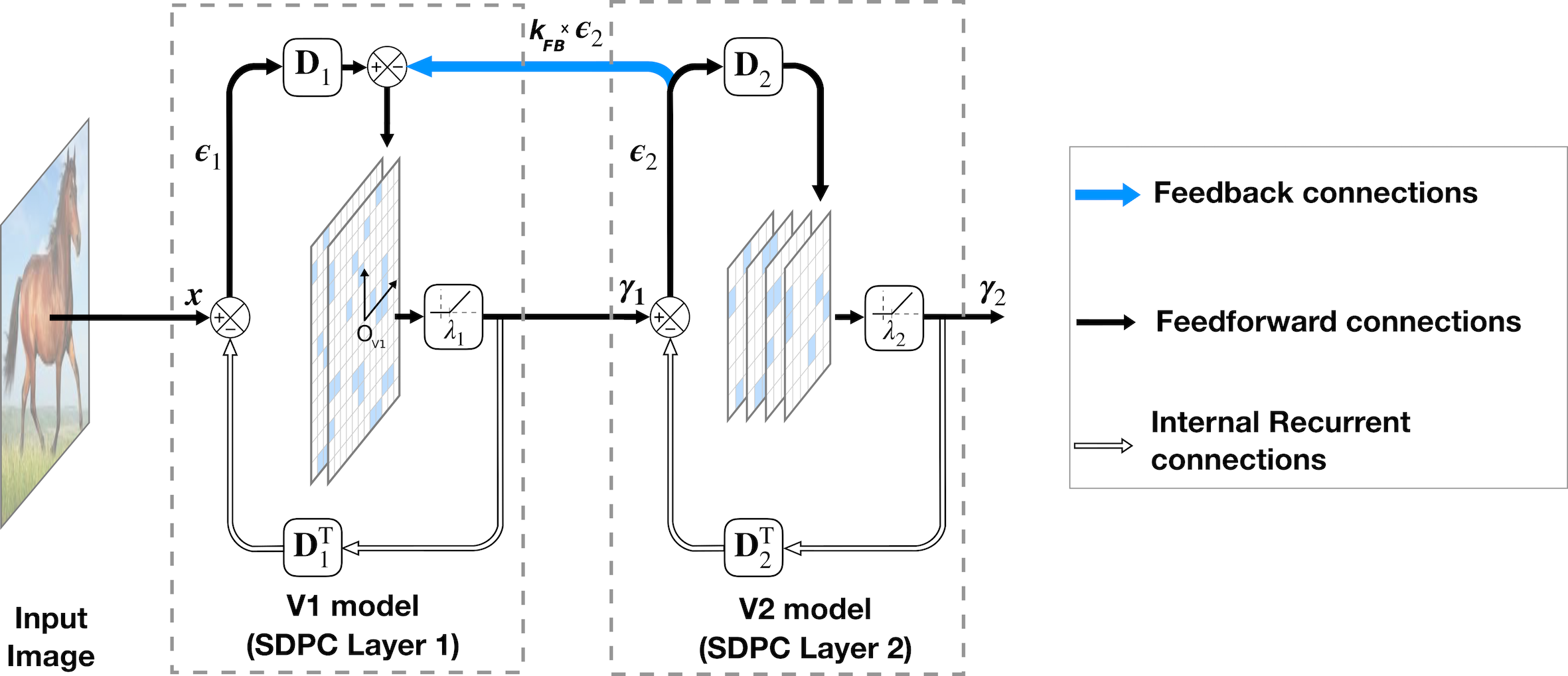}};
\end{tikzpicture}
\caption{{\bf Architecture of a 2-layered SDPC model.} In this model $\boldsymbol{\gamma}_{i}$ represents the activity of the neural population and $\boldsymbol{\epsilon}_{i}$  is the representation error (also called prediction error) at layer $i$. The synaptic weights of the feedback and feedforward connection at each layer ($\boldsymbol{D}_{i}^{T}$ and $\boldsymbol{D}_{i}$ respectively) are reciprocal. The level of sparseness is tuned with the soft thresholding parameter $\lambda_{i}$. The scalar $k_\mathrm{FB}$ controls the strength of the feedback connection represented with a blue arrow.}
\label{fig:fig_SDPC}
\end{figure}
\fi

We train the SDPC on $2$ different datasets: a face database (CFD) and a natural images database (STL-10). In this paper, all presented results are obtained with a SDPC network trained with a feedback strength equal to $1$ (i.e $k_\mathrm{FB}$=1). Once trained, and when specified, we vary the feedback strength to evaluate its effect on the inference process. Note that we have also experimented to equate the feedback strength during learning and inference, and the results obtained are extremely similar to those obtained when the feedback strength was set to $1$ during the SDPC training. For both databases, all the presented results are obtained on a testing set that is different from the training set (except when we describe the training in the section entitled~'\nameref{Sec:training}'. All network parameters and database specifications are listed in the~'\nameref{Sec:Methods}' section.

\subsection*{SDPC learns localized edge-like RFs in V1 and more specific RFs in V2}\label{Sec:training}
In this subsection, we present the results of the training of the SDPC model on both STL-10 and CFD databases with a feedback strength  $k_\mathrm{FB}$ equal to $1$ (Fig.~\ref{fig:fig2}). First-layer Receptive Fields (RFs) exhibit two different types of filters: low-frequency filters, and higher frequency filters that are localized, band-pass and similar to Gabor filters (Fig.~\ref{fig:fig2}-B and Fig.~\ref{fig:fig2}-G). The low-frequency filters are mainly encoding for textures and colors whereas the higher frequency ones describe contours. Second layer RFs (Fig.~\ref{fig:fig2}-D and Fig.~\ref{fig:fig2}-I) are built from a linear combination of the first layer RFs. For both databases, the second layer RFs are bigger than those in the first layer (approximately $3$ times bigger for both databases). We note that for the CFD database the second layer RFs present curvatures and specific face features, whereas on the STL-10 database they only exhibit longer oriented edges. This difference is mainly coming from the higher variety of images composing the STL-10 database: the identity of objects, their distance, and their angle of view are more diverse than in the CFD database. On the contrary, as the CFD database is composed only of well-calibrated centered faces, the SDPC model is able to extract curvatures and features that are common to all faces. In particular, we observe on the CFD database the emergence of face-specific features such as eyes, nose and mouth that are often selected by the model to describe the input (second layer RFs are ranked by their activation probability in a descending order in Fig.~\ref{fig:fig2}-D  and Fig.~\ref{fig:fig2}-I). All the $64$ first layer RFs and the $128$ second layer RFs learned by the SDPC on both databases are available in the Supporting information section (Fig.~\ref{fig:figSD1} for STL-10 and  Fig.~\ref{fig:figSD2} for CFD). The first layer reconstruction on both databases (Fig.~\ref{fig:fig2}-C and Fig.~\ref{fig:fig2}-H) are highly similar to the input image (Fig.~\ref{fig:fig2}-A and Fig.~\ref{fig:fig2}-F). In the second layer reconstructions (Fig.~\ref{fig:fig2}-E and Fig.~\ref{fig:fig2}-J), the details like textures and colors are faded and smoothed in favor of more pronounced contours. In particular, the contours of the STL-10 images reconstructed by the second layer of the SDPC are sketched with a few oriented lines.

\ifnum \plotfig=1
\begin{figure}[p]
	\centering
\begin{tikzpicture}
\draw  [line width=0.5mm,   black] (0.50\linewidth, 1\linewidth) -- (0.50\linewidth, -0.06\linewidth);
\draw [anchor=north west] (0.0\linewidth, 0.96\linewidth) node
{\includegraphics[width=0.48\linewidth]{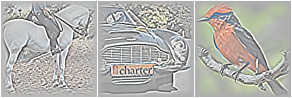}};

\draw [anchor=north west] (0.0\linewidth, 0.76\linewidth) node
{\includegraphics[width=0.48\linewidth]{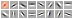}};

\draw [anchor=north west] (0.0\linewidth, 0.60\linewidth) node
{\includegraphics[width=0.48\linewidth]{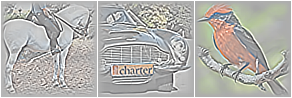}};

\draw [anchor=north west] (0.0\linewidth, 0.40\linewidth) node
{\includegraphics[width=0.48\linewidth]{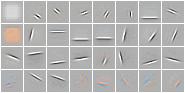}};

\draw [anchor=north west] (0.0\linewidth, 0.12\linewidth) node
{\includegraphics[width=0.48\linewidth]{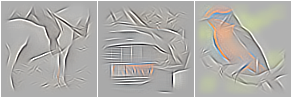}};

\draw [anchor=north west] (0.50\linewidth, 0.94\linewidth) node
{\includegraphics[width=0.48\linewidth]{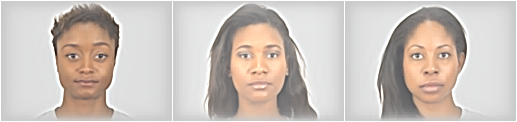}};

\draw [anchor=north west] (0.50\linewidth, .76\linewidth) node
{\includegraphics[width=0.48\linewidth]{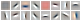}};

\draw [anchor=north west] (0.50\linewidth, .58\linewidth) node {\includegraphics[width=0.48\linewidth]{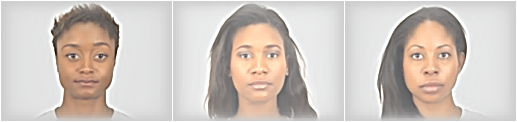}};
\draw [anchor=north west] (0.50\linewidth, .40\linewidth) node {\includegraphics[width=0.48\linewidth]{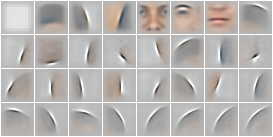}};
\draw [anchor=north west] (0.50\linewidth, .10\linewidth) node {\includegraphics[width=0.48\linewidth]{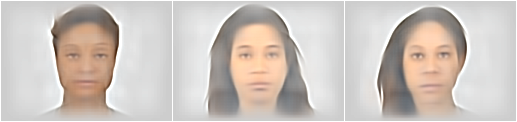}};
\begin{scope}
\draw [anchor=west,fill=white] (0.13\linewidth, 1\linewidth) node {{\bf STL-10 database}};
\draw [anchor=west,fill=white] (0.65\linewidth, 1\linewidth) node {{\bf CFD database}};

\draw [anchor=west,fill=white] (0.0\linewidth, 0.97\linewidth) node {{\bf A}};
\draw [anchor=west,fill=white] (0.0\linewidth, .77\linewidth) node  {{\bf B}};
\draw [anchor=west,fill=white] (0.0\linewidth, .61\linewidth) node  {{\bf C}};
\draw [anchor=west,fill=white] (0.0\linewidth, .41\linewidth) node  {{\bf D}};
\draw [anchor=west,fill=white] (0.0\linewidth, .13\linewidth) node  {{\bf E}};
\draw [anchor=west,fill=white] (0.52\linewidth, 0.97\linewidth) node {{\bf F}};
\draw [anchor=west,fill=white] (0.52\linewidth, .77\linewidth) node  {{\bf G}};
\draw [anchor=west,fill=white] (0.52\linewidth, .61\linewidth) node  {{\bf H}};
\draw [anchor=west,fill=white] (0.52\linewidth, .41\linewidth) node  {{\bf I}};
\draw [anchor=west,fill=white] (0.52\linewidth, .13\linewidth) node  {{\bf J}};
\end{scope}
\end{tikzpicture}
\caption{{\bf Results of training SDPC on the STL-10 database (right column) and on the CFD database (left column) with a feedback strength $\mathbf{k_\mathrm{FB}=1}$.}
{\bf (A)} \& {\bf (F)}:~Randomly selected input images. 
Both databases are pre-processed with Local Contrast Normalization~\cite{jarrett2009best} and whitening.
{\bf (B)} \& {\bf (G)}:~16 randomly selected first-layer RFs from the 64 RFs composing $\boldsymbol{D}_{1}^\mathrm{eff}$ (see Eq.~\ref{eq:Effective_Dictionaries}). 
The RFs are ranked by their activation probability in a descending order. 
The RF size of neurons located on the first layer is 8$\times$8 px on the STL-10 database {\bf (B)} and 9$\times$9 px on the CFD database {\bf (G)}.
{\bf (C)} \& {\bf (H)}:~Reconstruction of images corresponding to the input images shown in {\bf (A)} \& {\bf (F)} from the representation in the first layer, denoted $\boldsymbol{\gamma}_{1}^\mathrm{eff}$ (see Eq.~\ref{eq:Reconstruction}).
{\bf (D)} \& {\bf (I)}:~32 sub-sampled RFs out 128 RFs composing  $\boldsymbol{D}_{2}^\mathrm{eff}$ (see Eq.~\ref{eq:Effective_Dictionaries}), ranked by their activation probability in descending order. The size of the  RF from neurons located on the second layer is $22\times 22$ px on the STL-10 database {\bf (D)} and 33$\times$33 px on the CFD database {\bf (I)}.
{\bf (E)} \&\ {\bf (J)}:~Reconstruction of images corresponding to the input images shown in {\bf (A)} \& {\bf (F)} from the representation in the second layer, denoted $\boldsymbol{\gamma}_{2}^\mathrm{eff}$ (see Eq.~\ref{eq:Reconstruction}).
}
\label{fig:fig2}
\end{figure}
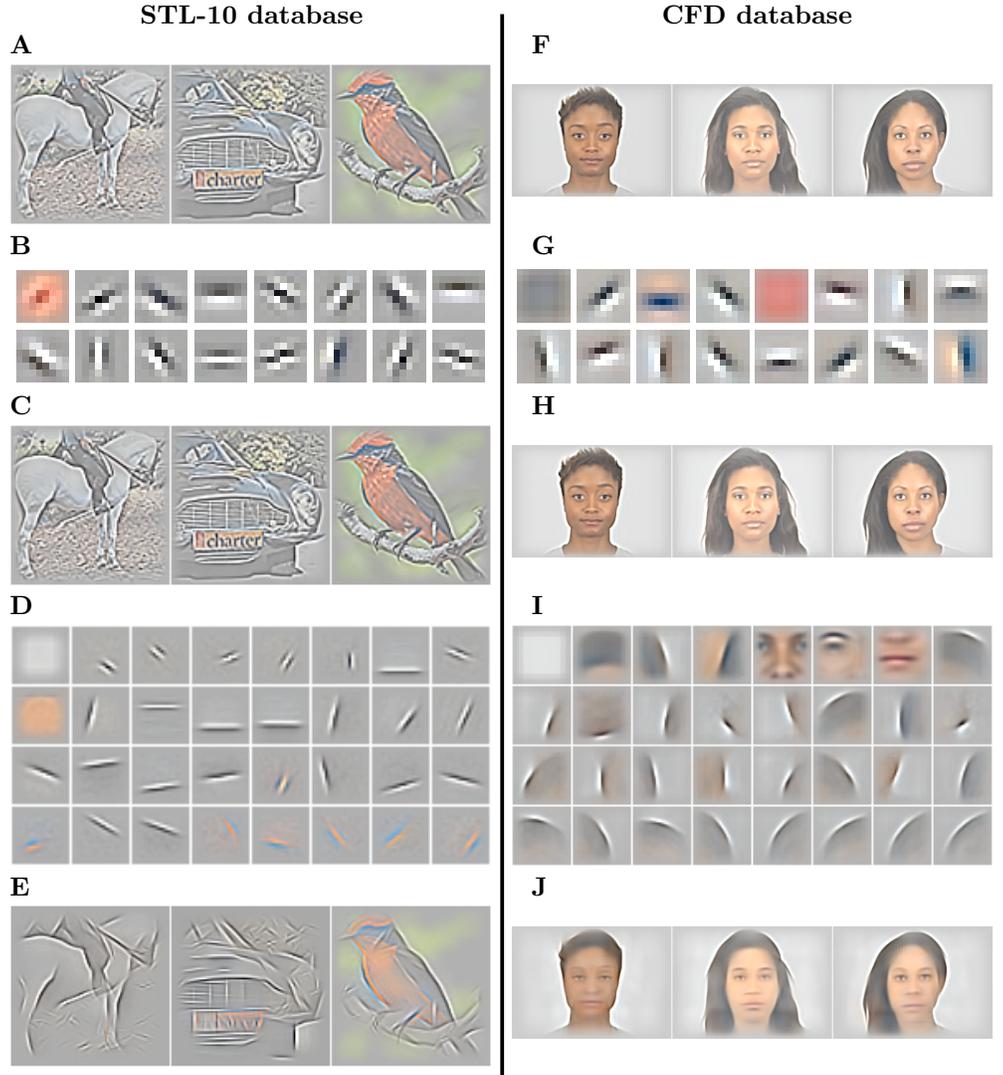

\fi
\subsection*{Effect of the feedback at the neural level}
We now vary the strength of the feedback connection to assess its impact on neural representations when an image is presented as a stimulus. The strength of the feedback, $k_\mathrm{FB}$, is a scalar ranging from $0$ to $4$. When $k_\mathrm{FB}$ is set to $0$, the feedback connection is suppressed, in other words, the neural activity in the first layer is independent of the neural activity at the second layer. Inversely, when $k_\mathrm{FB}=4$ feedback signals are strongly amplified such that it reinforces the interdependence between the neural activities of both layers. As a consequence, varying the feedback strength should affect the first layer activity. The objective of this subsection is to study this effect on the organization of V1 neurons (i.e. the first layer of the SDPC).

\subsubsection*{SDPC feedback recruits more neurons in the V1 model}
In the first experiment, we monitor the median number of active neurons in our V1 model when varying the feedback strength on both databases. The medians are computed over $1200$ images of STL-10 database (Fig.~\ref{fig:fig3}-A) and $400$ images of CFD database (Fig.~\ref{fig:fig3}-B). In this paper, we use the median $\pm$ Median Absolute Deviation (MAD) instead of the classical mean $\pm$ standard deviation to avoid assuming that samples are normally distributed~\cite{pham2001mean}.  For the same reason, all the statistical tests are performed using the Wilcoxon signed-rank test. It will be denoted $\mathrm{WT}(N=1200, p<0.01)$ when the null hypothesis is rejected. In this notation $N$ is the number of samples and $p$ is the corresponding probability value (p-value). 
In contrast, we will formalize the test by $\mathrm{WT}(N=1200, p=0.3)$ when the null hypothesis cannot be rejected. 

For both databases, we observe that the percentage of active neurons increases with the strength of the feedback. In particular, we note a strong increase in the number of activated neurons when we restore the feedback connection (from $k_\mathrm{FB}=0$ to $k_\mathrm{FB}=1$): +$8.7\%$ and +$4.7\%$ for STL-10 and CFD databases, respectively. Incrementally amplifying the feedback strength above $1$ further increases the number of active neurons in the first layer even if the effect is sublinear. All the increases in the percentage of recruited neurons with the feedback strength are significant as quantified with statistical tests between all pairs of feedback strength: $\mathrm{WT}(N=1200, p<0.01)$ for STL-10 database and  $\mathrm{WT}(N=400, p<0.01)$ for CFD database. For each database, we notice that the inter-stimuli variability, as illustrated by error-bars, is lower when the feedback connection is removed: $0.80\%$ with $k_\mathrm{FB}=0$ versus $2.55\%$ with $k_\mathrm{FB}=1$ for the STL-10 database and $0.25\%$ with $k_\mathrm{FB}=0$ versus $0.85\%$ with $k_\mathrm{FB}=1$ for the CFD database. The results of this first experiment suggest that as the feedback gets stronger, the number of recruited neurons becomes larger. 

\ifnum \plotfig=1
\begin{figure}[!ht]
	\centering
\begin{tikzpicture}
\draw [anchor=north west] (0.0\linewidth, 0.98\linewidth) node {\includegraphics[width=0.47\linewidth]{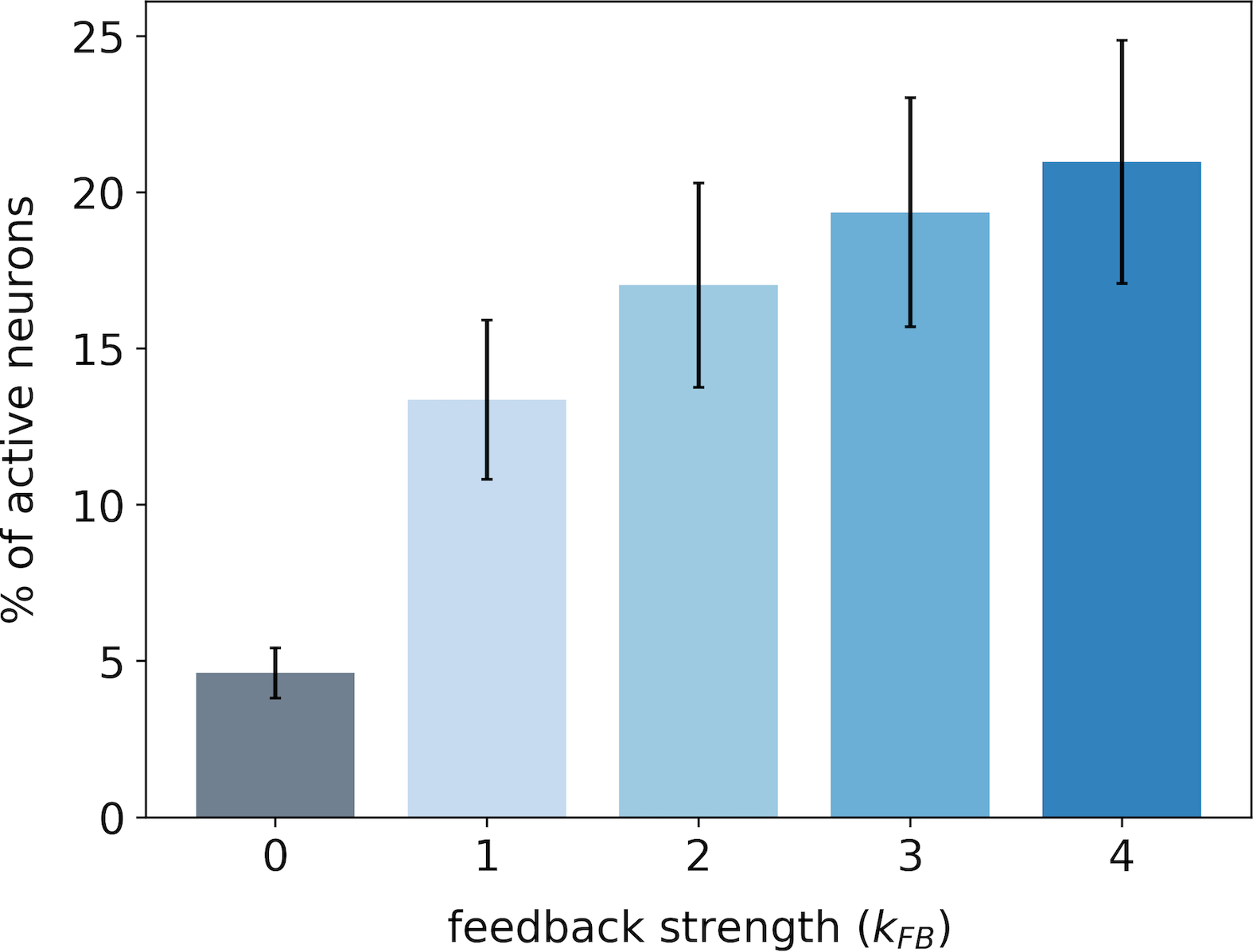}};
\draw [anchor=north west] (0.50\linewidth, 0.98\linewidth) node {\includegraphics[width=0.47\linewidth]{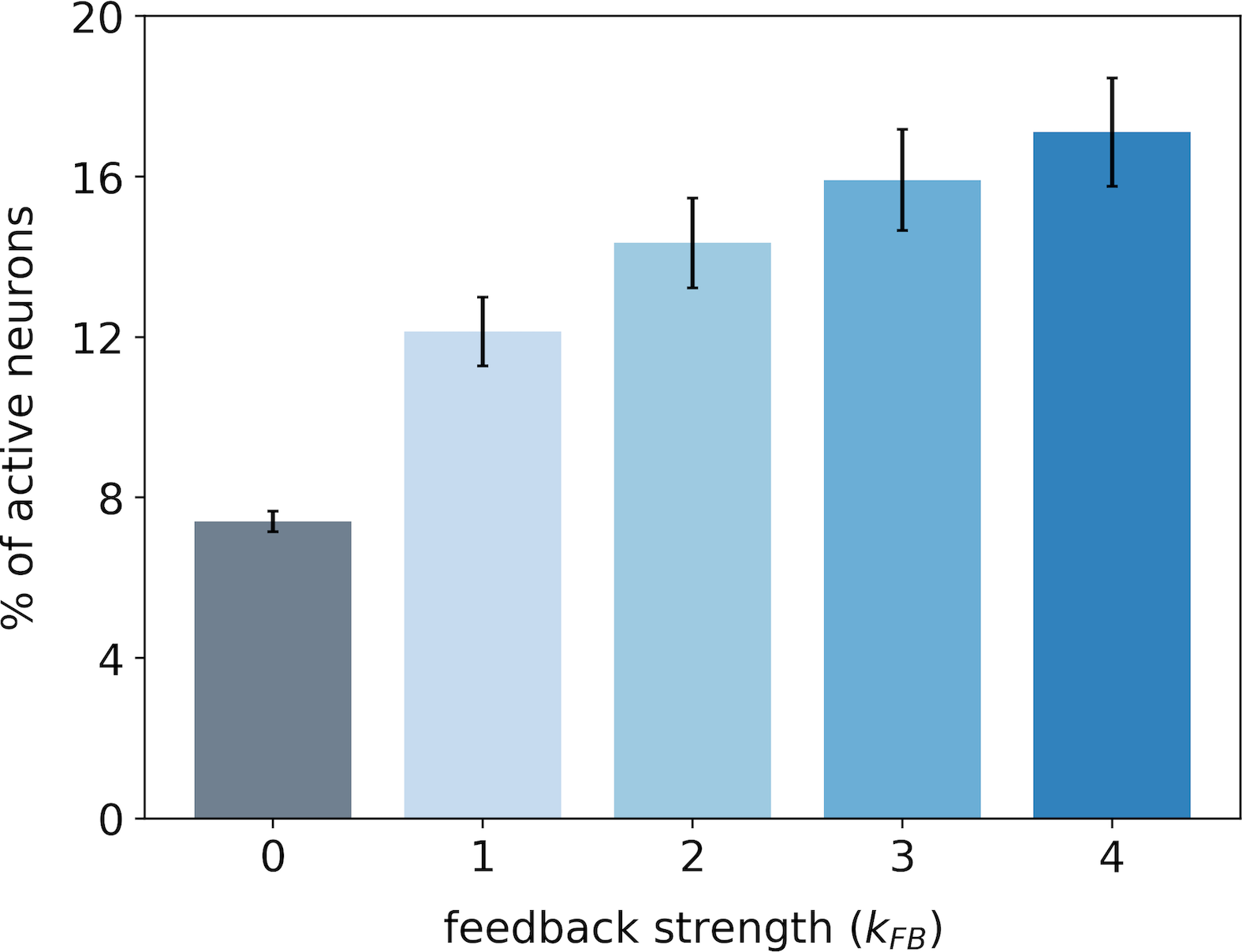}};
\begin{scope}
\draw [anchor=west,fill=white] (0.0\linewidth, 1\linewidth) node {{\bf A}};
\draw [anchor=west,fill=white] (0.5\linewidth, 1\linewidth) node {{\bf B}};
\end{scope}
\end{tikzpicture}
\caption
{{\bf Percentage of active neurons in the first layer of the SDPC model}. {\bf (A)} On the STL-10 database. {\bf (B)} On the CFD database.
We record the percentage of active neurons with a feedback strength $k_\mathrm{FB}$ varying from 0 (no feedback) to 4 (strong feedback). The height of the bars represent the median percentage of active neurons and the error bars are computed using the Median Absolute Deviation (MAD) over $1200$ and $400$ images of the testing set for STL-10 and CFD, respectively.}
\label{fig:fig3}
\end{figure}
\fi

\subsubsection*{SDPC feedback signals reorganize the interaction map of the V1 model}
We investigate the effect of feedback on the neural organization in our V1 model when the SDPC is trained on natural images (i.e.  STL-10 database).  The V1 activity-map ($\boldsymbol{\gamma}_{1}$) being a high-dimensional tensor, it is a priori difficult to visualize its internal organization. We define the notion of interaction map to reduce the neural activity to two state variables at every position on the V1 space: the resulting orientation and activity. Using interaction map allows us to represent in 2D the state of the network surrounding a given central position. We choose the location of the center of the interaction map such that neurons, at this position, are strongly responsive to a given orientation. This orientation is called the central preferred orientation and denoted $\theta_c$. Interaction maps are denoted $\boldsymbol{\bar{a}}$, the maps of resulting orientations and activities are denoted $\boldsymbol{\bar{\theta}}$ and $\big \vert \boldsymbol{\bar{a}} \big \vert$, respectively. The computation of the interaction maps is detailed in the subsection~'\nameref{sec:InteractionMap}' of the section~'\nameref{Sec:Methods}' .

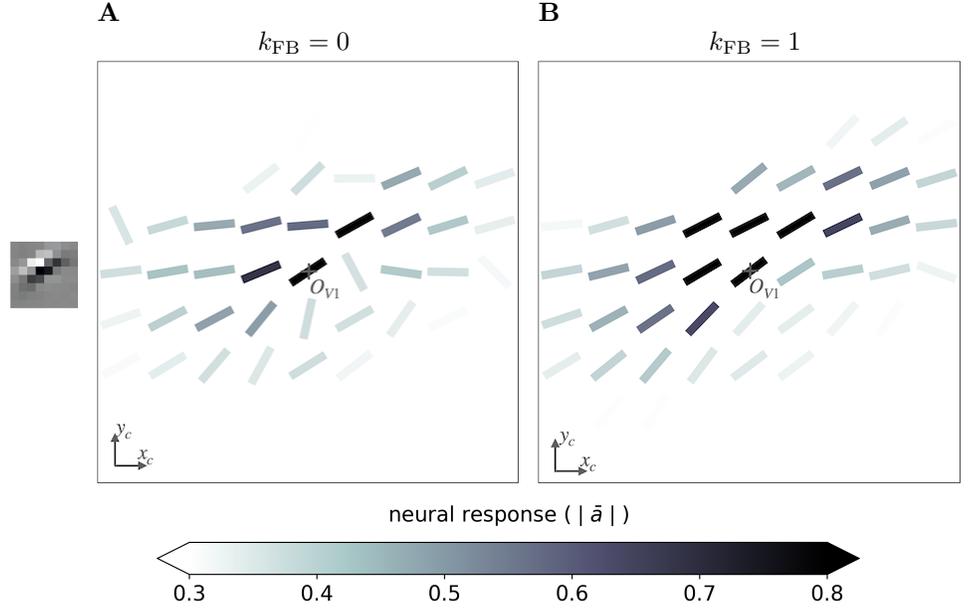
\begin{figure}[h]
    \centering
\begin{tikzpicture}
\draw [anchor=north west] (0.0\linewidth, 0.78\linewidth) node {\includegraphics[width=0.07\linewidth]{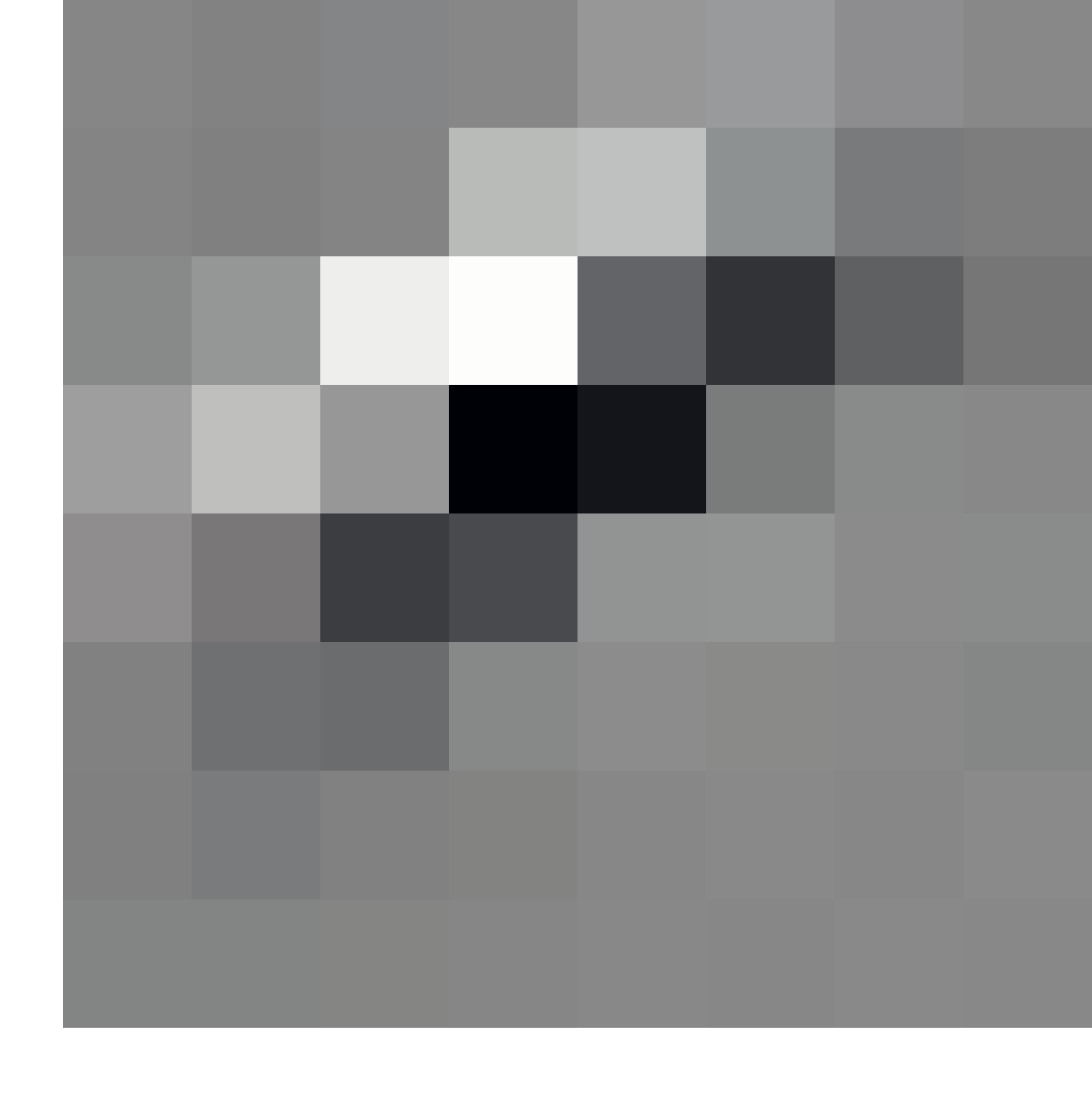} };
\draw [anchor=north west] (0.09\linewidth, 0.96\linewidth) node {\includegraphics[width=0.42\linewidth]{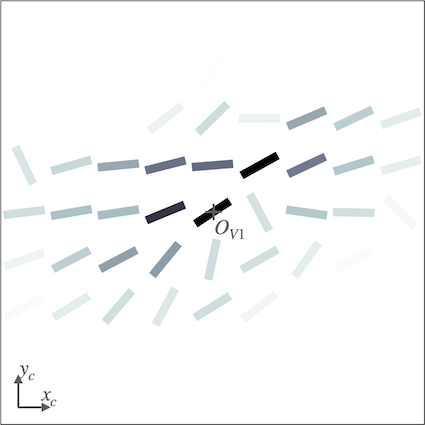}};
\draw [anchor=north west] (0.53\linewidth, 0.96\linewidth) node {\includegraphics[width=0.42\linewidth]{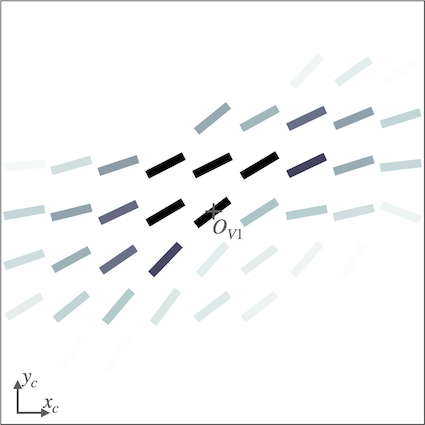}};
\draw [anchor=north west] (0.15\linewidth, 0.52\linewidth) node  {\includegraphics[width=0.7\linewidth]{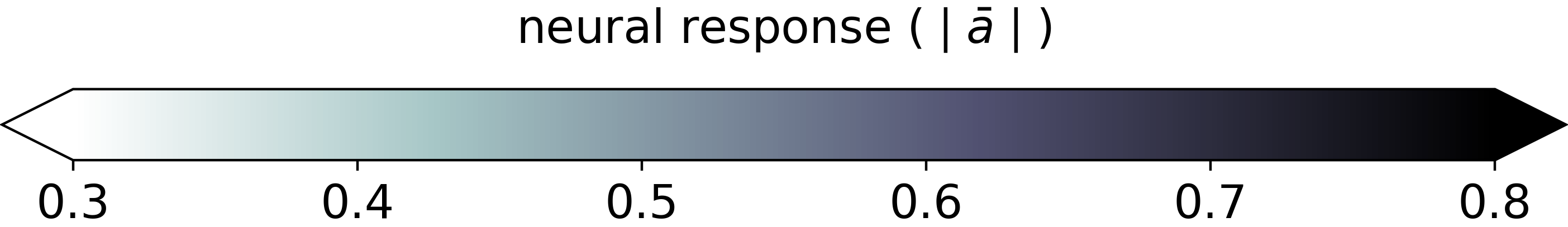}};
\begin{scope}
\draw [anchor=west,fill=white] (0.09\linewidth, 1\linewidth) node {{\bf A}};
\draw [anchor=west,fill=white] (0.53\linewidth, 1\linewidth) node {{\bf B}};
\draw [anchor=west,fill=white] (0.25\linewidth, 0.97\linewidth) node {$k_\mathrm{FB}=0$};
\draw [anchor=west,fill=white] (0.70\linewidth, 0.97\linewidth) node {$k_\mathrm{FB}=1$};
\end{scope}
\end{tikzpicture}
\caption
{{\bf Example of a 9$\times$9 interaction map of a V1 area centered on neurons strongly responding to a central preferred orientation of \ang{30}}. {\bf(A)} Without feedback. {\bf (B)} With a feedback strength equal to $1$.  These interaction maps are obtained when the SDPC is trained on natural images (i.e. STL-10). At each location identified by the coordinates $(x_{c},y_{c})$ the angle is $\boldsymbol{\bar{\theta}}[x_{c},y_{c}]$ (see Eq.~\ref{eq:eq6}) and the color scale is $\big \vert \boldsymbol{\bar{a}}[x_{c},y_{c} ] \big \vert$ (see Eq.~\ref{eq:eq7}). The color scale being saturated toward both maximum and minimum activity, all the activities above $0.8$ or below $0.3$ have the same color, respectively dark or white.}
\label{fig:fig4}
\end{figure}

For all feedback strengths and different central preferred orientations, we observe that the interaction maps are highly similar to association fields~\cite{field1993contour}: most of the orientations of the interaction map are co-linear and/or co-circular to the central preferred orientation (see Fig.~\ref{fig:fig4} for one example of this phenomenon and Fig.~\ref{fig:figSD3} for more examples with $k_\mathrm{FB}$=1). In addition, interaction maps exhibit a strong activity in the center and towards the end-zone of the central preferred orientation. We define the end-zone as the region covering the axis of the central preferred orientation, and the side-zone as the area covering the orthogonal axis of the central preferred orientation. The activity of the interaction map in the side-zone is lower compared to the activity in the end-zone. We notice qualitatively that the orientations of the interaction maps are less co-linear to the central preferred orientation when feedback is suppressed~(i.e. $k_\mathrm{FB}=0$). In other words, when feedback is active, the interaction map looks more organized compared to the interaction map generated without feedback (see Fig.~\ref{fig:fig4} for a striking example of this phenomenon).

We next quantify this organizational difference for different levels of feedback strength by estimating the co-linearity and the co-circularity deviation of the interaction map in both regions: the end-zone and the side-zone. To perform such an analysis, we measure the co-linearity deviation of the interaction map from the central preferred orientation, denoted $\boldsymbol{\theta}_{co-lin}$ (see Eq.~\ref{eq:eq8} in section~\nameref{Sec:Methods}). In addition, we measure the co-circularity deviation of the interaction map from a map of orientations that are co-circular to the central preferred orientation $\boldsymbol{\theta}_{co-cir}$ (see Eq.~\ref{eq:eq9} in section~\nameref{Sec:Methods}).

When feedback is suppressed (ie $k_\mathrm{FB}=0$), we measure a median co-linearity deviation equal to $\ang{9}$ and  $\ang{13}$ in the end-zone and in the side-zone of the interaction map, respectively (the medians are computed over all the central preferred orientations). These measures are to be compared with the marginal co-linearity, i.e the median co-linearity deviation outside the interaction map. We report a marginal co-linearity of $\ang{43}$ in both regions (as computed with a spatially shuffled version of the V1 activity-map). Consequently, orientations inside the interaction map are more co-linear to the central preferred orientation than orientations located outside of the interaction map. Still without feedback, we measure a median co-circularity deviation equal to $\ang{26}$ and $\ang{39}$ in the end-zone and the side-zone of the interaction map, respectively. The marginal co-circularity, i.e the median co-circularity deviation outside the interaction map is equal to $\ang{37}$ in both regions. Thus, orientations located in the end-zone of the interaction map are more co-circular to the central preferred orientation than orientations located outside the interaction map. In contrast, side-zone orientations in the interaction map are equally co-circular to the central preferred orientation that orientations located outside of the interaction map.

For a given feedback strength $k_\mathrm{FB}$, we synthesize these results by introducing two ratios to compare the respective precisions in co-linearity ($r_{co-lin}^{k_\mathrm{FB}}$ in Eq.~\ref{eq:eq10} in section~\nameref{Sec:Methods})  and co-circularity ($r_{co-cir}^{k_\mathrm{FB}}$ in Eq.~\ref{eq:eq11} in section~\nameref{Sec:Methods}). These are derived by dividing the marginal co-linearity/co-circularity deviation by the co-linearity/co-circularity deviation in the interaction map.

\begin{figure}[h]
	\centering
\begin{tikzpicture}
\draw [anchor=north west] (0.00\linewidth, 0.99\linewidth) node {\includegraphics[width=0.47\linewidth]{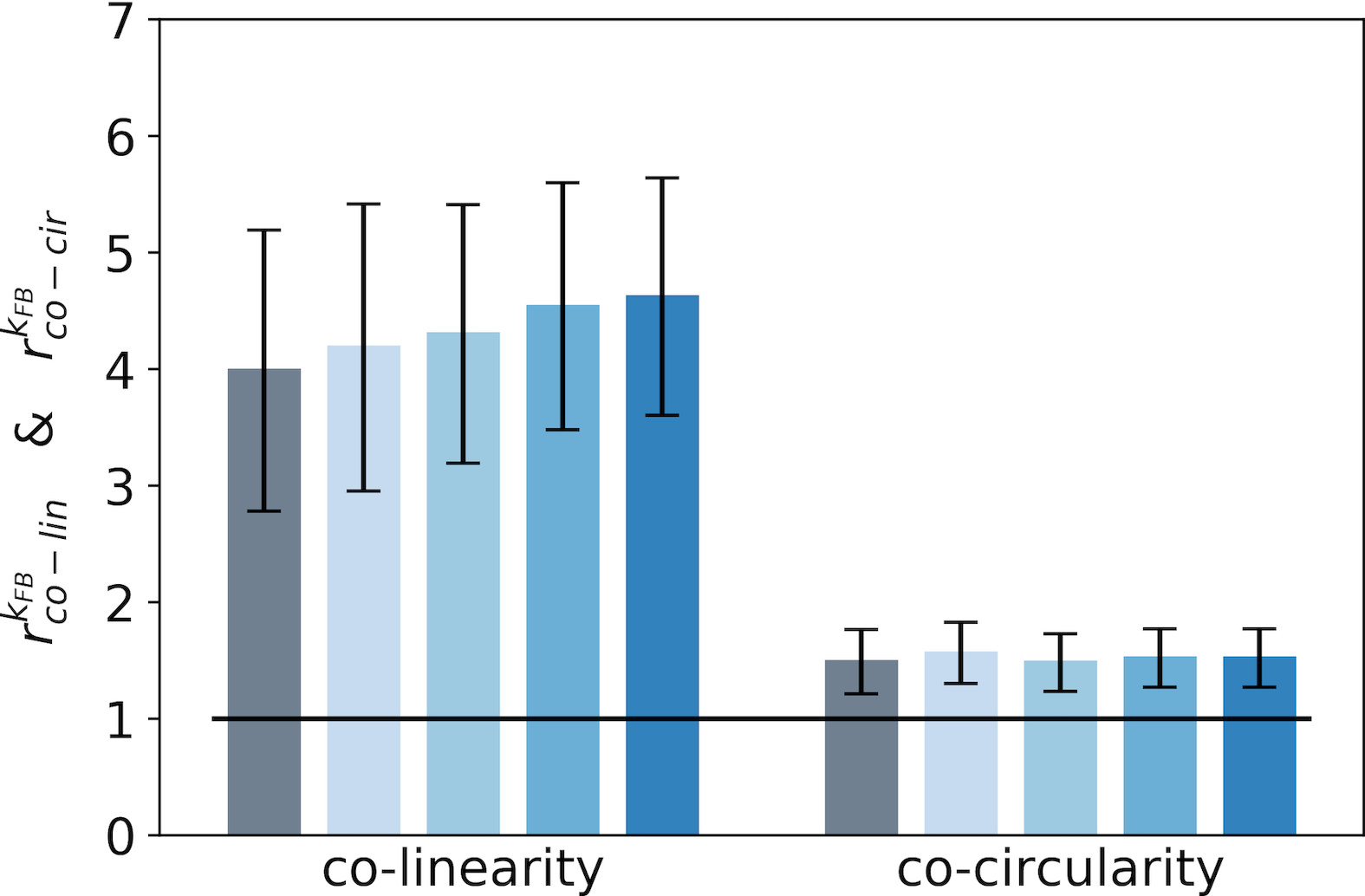}};
\draw [anchor=north west] (0.50\linewidth, 0.99\linewidth) node {\includegraphics[width=0.47\linewidth]{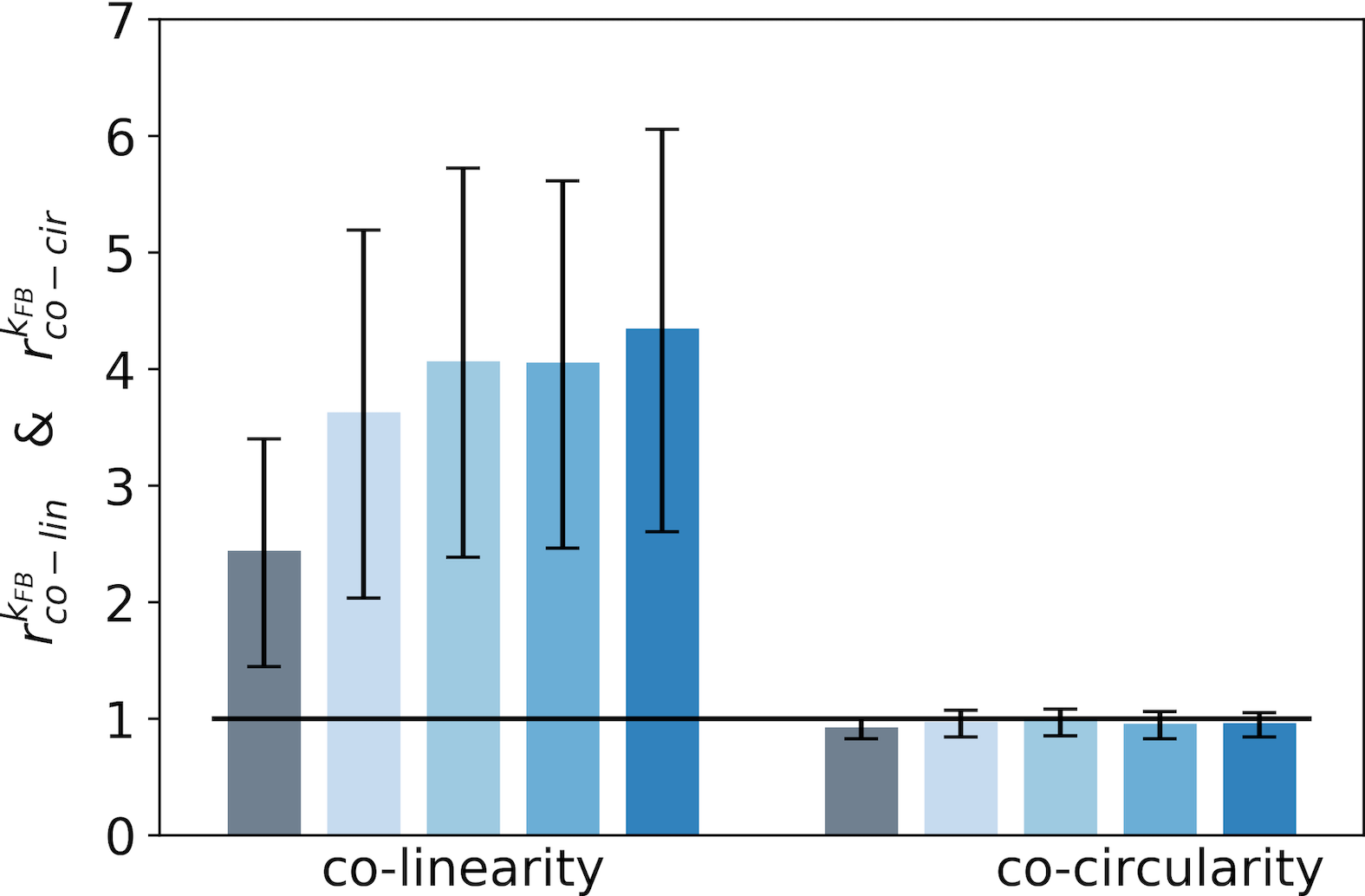}};
\draw [anchor=north west] (0.385\linewidth, 0.98\linewidth) node {\includegraphics[width=0.08\linewidth]{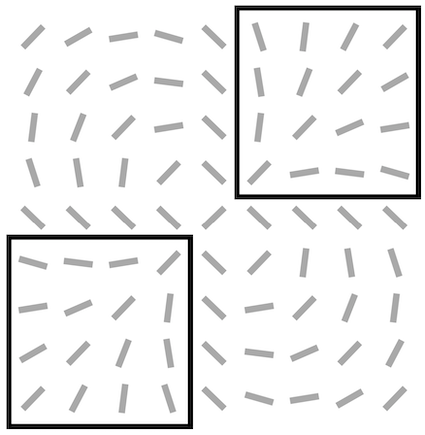} };
\draw [anchor=north west] (0.885\linewidth, 0.98\linewidth) node {\includegraphics[width=0.08\linewidth]{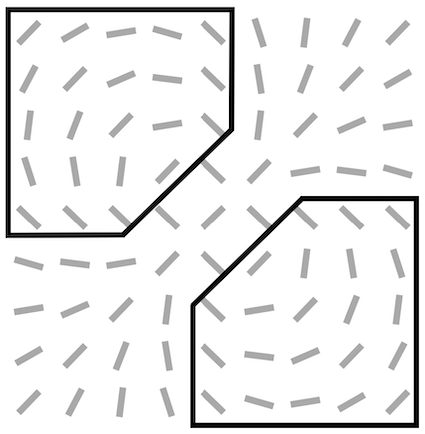} };
\draw [anchor=north west] (0\linewidth, 0.65\linewidth) node  {\includegraphics[width=1\linewidth]{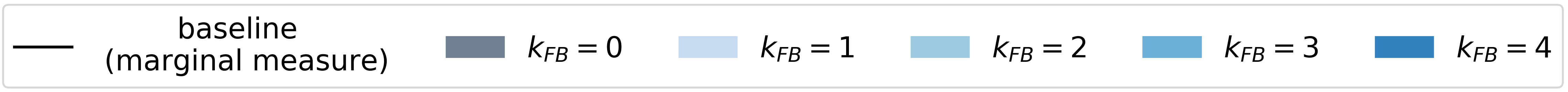}};
\begin{scope}
\draw [anchor=west,fill=white] (0.0\linewidth, 1\linewidth) node {{\bf A}};
\draw [anchor=west,fill=white] (0.5\linewidth, 1\linewidth) node {{\bf B}};
\end{scope}
\end{tikzpicture}
\caption
{{\bf Relative co-linearity and co-circularity of the V1 interaction map w.r.t. marginal co-linearity/co-circularity}. {\bf (A)} In the end-zone. {\bf (B)} In the side-zone. For each plot, the left block of bars represents the relative co-linearity as quantified with $r_{co-lin}^{k_\mathrm{FB}}$ (see Eq.~\ref{eq:eq10}) . The right block of bars in each plot represents the relative co-circularity as quantified with $r_{co-cir}^{k_\mathrm{FB}}$ (see Eq.~\ref{eq:eq10}). Bars' heights represent the median over all the orientations, and error bar are computed as the Median Absolute Deviation. The baseline represents the co-linearity and co-circularity without feedback.}
\label{fig:fig5}
\end{figure}

We report these two ratios for the end-zone (Fig.~\ref{fig:fig5}-A) and the side-zone (Fig.~\ref{fig:fig5}-B). As a first observation, for all feedback strengths the end-zone orientations in the interaction map are both more co-linear and co-circular to the central preferred orientation than orientations located outside the interaction map (as computed by the marginal baseline, see Fig.~\ref{fig:fig5}-A). This increase of co-linearity and co-circularity compared to the baseline in the end-zone is significant for all feedback strengths ($\mathrm{WT}(N=51, p<0.01)$). All increases of co-linearity between $k_\mathrm{FB}=0$ and $k_\mathrm{FB}> 0$ are also significant as reported by pair-wise statistical test ($\mathrm{WT}(N=51, p<0.01)$). In contrast, the increases in co-linearity between the rest of the feedback strength are not significant as measured by the minimum p-value of all pair-wise statistical tests ($\mathrm{WT}(N=51, p=0.09)$). Concerning the side-zone and for all feedback strengths, orientations in the interaction map are significantly more co-linear to the central preferred orientation compared to the marginal co-linearity (see Fig.~\ref{fig:fig5}-B) ($\mathrm{WT}(N=51, p<0.01)$). Interestingly, the increases of co-linearity between $k_\mathrm{FB}=0$ and $k_\mathrm{FB}> 0$  in the side-zone are also significant as quantified by pair-wise statistical tests ($\mathrm{WT}(N=51, p<0.01)$). Besides, the co-circularity in the side-zone is at the marginal level or below. However, this observation is not significant as reported by the maximum p-value of all pair-wise statistical test ($\mathrm{WT}(N=51, p=0.93)$). In addition, we observe that the feedback strength has a strong effect on the co-linearity in the side-zone and a smoother effect on the co-linearity in the end-zone. While the relative co-linearity w.r.t. marginal co-linearity is strong in the end-zone it is weaker in the side-zone when feedback is suppressed. The feedback signals in the side-zone tend to catch-up with the co-linearity level in the end-zone. Our analysis suggests that interaction maps without feedback are both co-linear and co-circular in the end-zone of the interaction map, but only slightly co-linear and not co-circular in the side-zone. Interestingly, increasing the feedback strength tends to orientate the side-zone of the interaction map co-linearly to the central preferred orientation.

\subsubsection*{SDPC feedback signals modulate the activity within the interaction map}
\begin{figure}[h]
	\centering
\begin{tikzpicture}
\draw [anchor=north west] (0.10\linewidth, 0.99\linewidth) node {\includegraphics[width=0.42\linewidth]{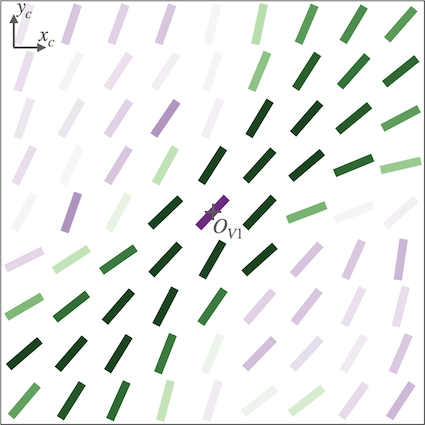}};
\draw [anchor=north west] (0.\linewidth, 0.82\linewidth) node {\includegraphics[width=0.07\linewidth]{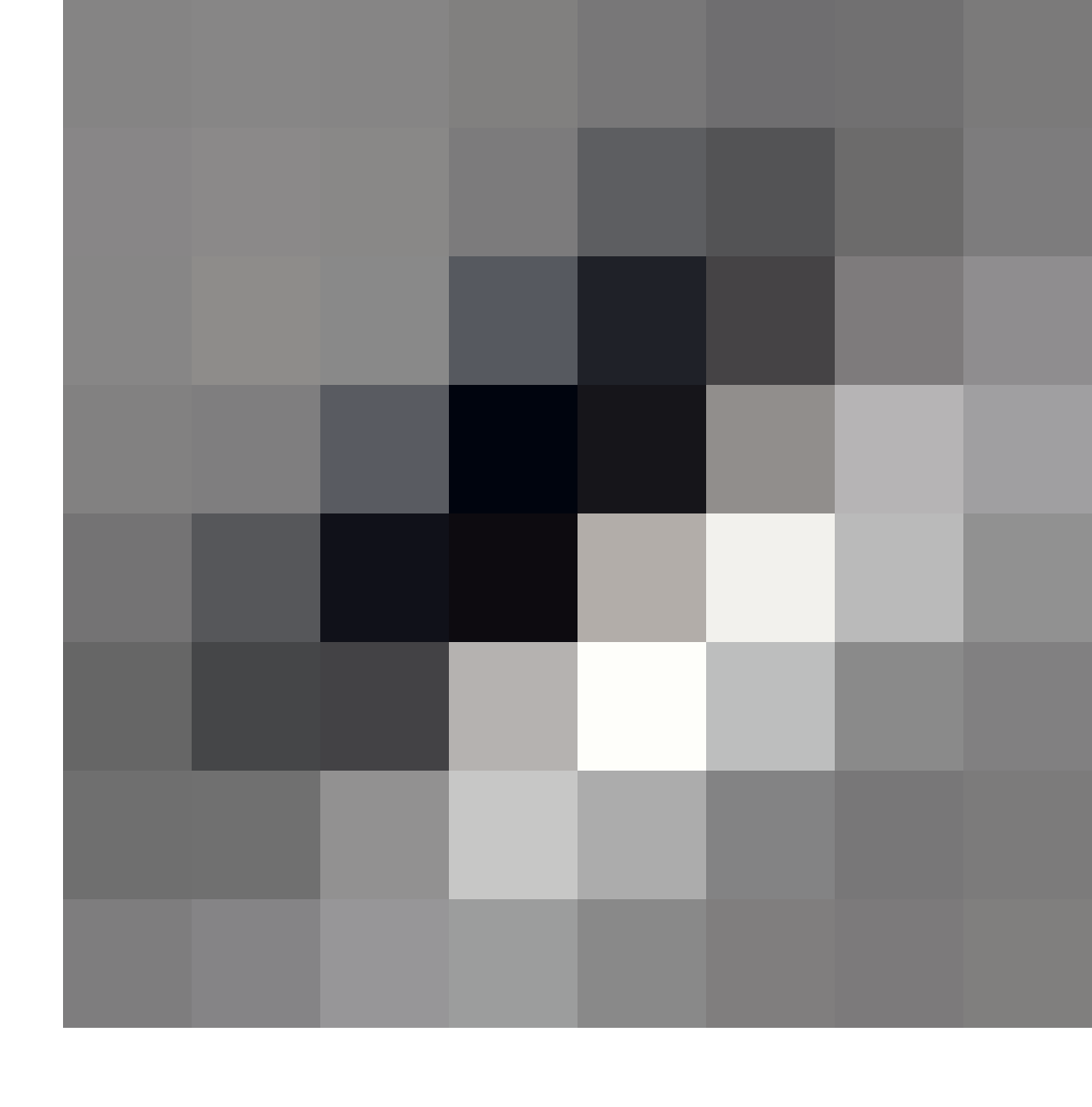}};
\draw [anchor=north west] (0.55\linewidth, 1\linewidth) node {\includegraphics[width=0.105\linewidth]{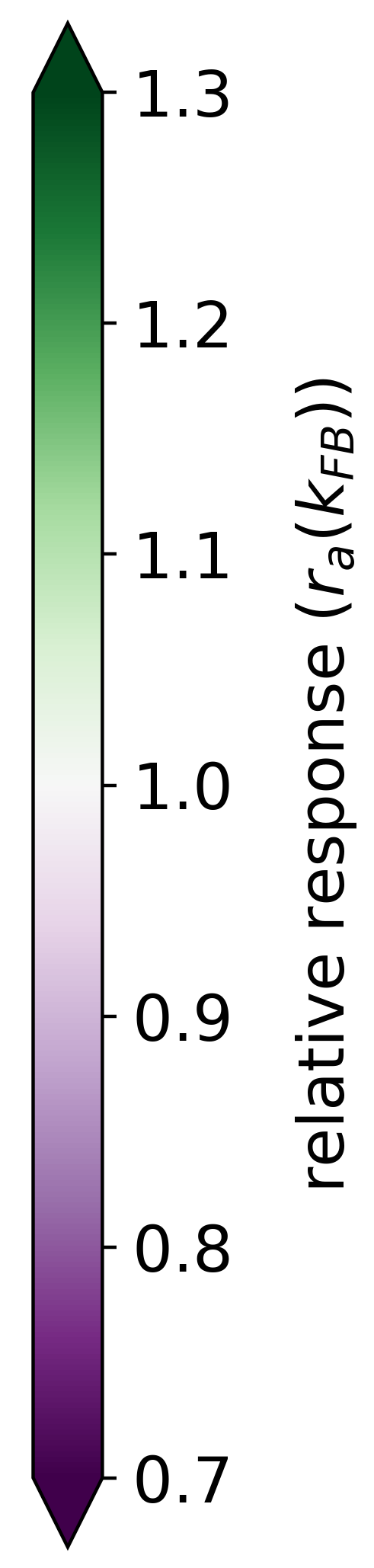}};
\end{tikzpicture}
\caption
{{\bf Example of a 9$\times$9 interaction map of a V1 area centered on neurons strongly responding to a central preferred orientation of \ang{45}, and colored with the relative response w.r.t. no feedback.} The feedback strength is set to $1$ and the SDPC is trained on natural images (i.e. STL-10). At each location identified by the coordinates $(x_{c},y_{c})$ the angle is $\boldsymbol{\bar{\theta}}[x_{c},y_{c}]$ (see Eq.~\ref{eq:eq6}) and the color scale is proportional to $\boldsymbol{r_{a}}(k_\mathrm{FB})$ (see Eq.~\ref{eq:eq12}). The color scale being saturated toward both maximum and minimum activity, all the activities above $1.3$ or below $0.5$ have the same color, respectively dark green or purple.}
\label{fig:fig6}
\end{figure}

To study the effect of the feedback on the level of activity within the interaction map, we introduce the ratio $\boldsymbol{r_{a}}(k_\mathrm{FB})$ between the activity with a certain feedback strength and the activity when the feedback is suppressed (see Eq.~\ref{eq:eq12} in section~\nameref{Sec:Methods}). Coloring the interaction map using a color scale proportional to $\boldsymbol{r_{a}}(k_\mathrm{FB})$ allows us to identify which part of the map is more activated with the feedback. First, we observe qualitatively that the interaction map in the end-zone is more strongly activated when the feedback connection is active. On the contrary, the side-zone exhibited weaker activities when feedback is turned on (see Fig.~\ref{fig:fig6} and Fig.~\ref{fig:figSD4} for examples of this phenomenon with $k_\mathrm{FB}$=1). Note also that the activity in the center of the interaction map, which corresponds to the classical RF area, is lowered when feedback is active.

\begin{figure}[h]
	\centering
\begin{tikzpicture}

\draw [anchor=north west] (0.00\linewidth, 0.99\linewidth) node {\includegraphics[width=0.30\linewidth]{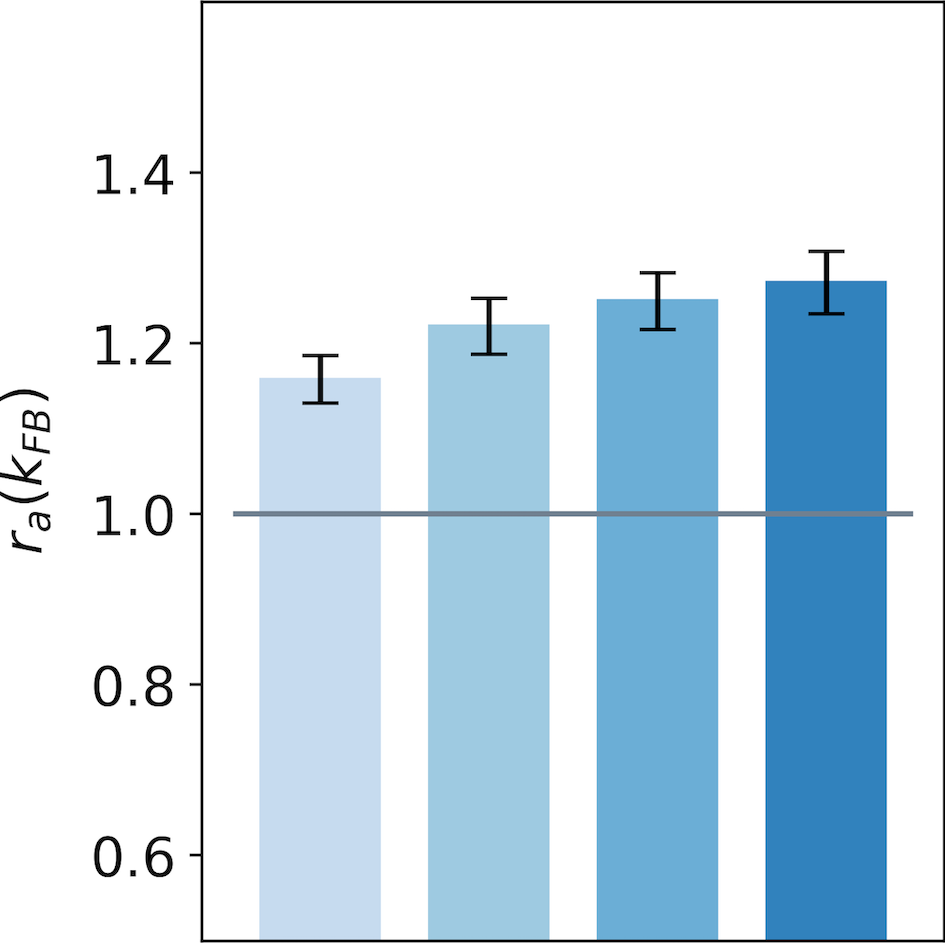}};
\draw [anchor=north west] (0.33\linewidth, 0.99\linewidth) node {\includegraphics[width=0.30\linewidth]{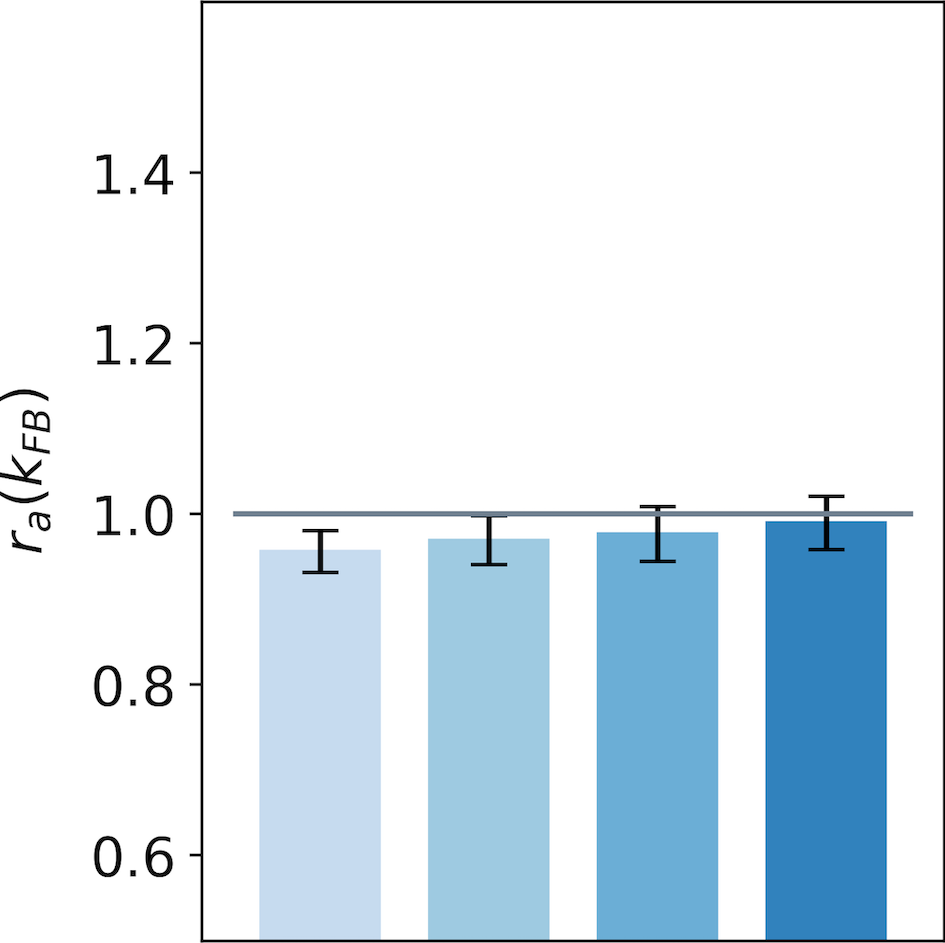}};
\draw [anchor=north west] (0.66\linewidth, 0.99\linewidth) node {\includegraphics[width=0.30\linewidth]{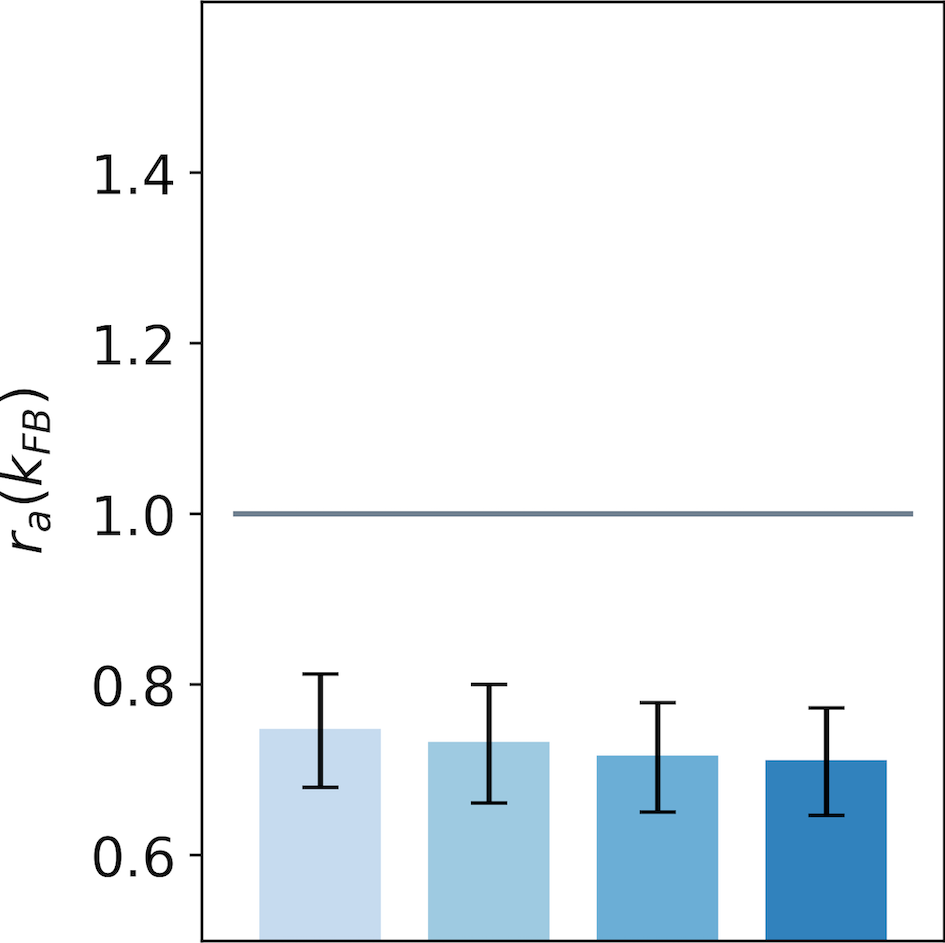}};
\draw [anchor=north west] (0.07\linewidth, 0.985\linewidth) node {\includegraphics[width=0.08\linewidth]{Figure/Fig5/Region_R1.png} };
\draw [anchor=north west] (0.40\linewidth, 0.985\linewidth) node {\includegraphics[width=0.08\linewidth]{Figure/Fig5/Region_R2.png} };
\draw [anchor=north west] (0.73\linewidth, 0.985\linewidth) node {\includegraphics[width=0.08\linewidth]{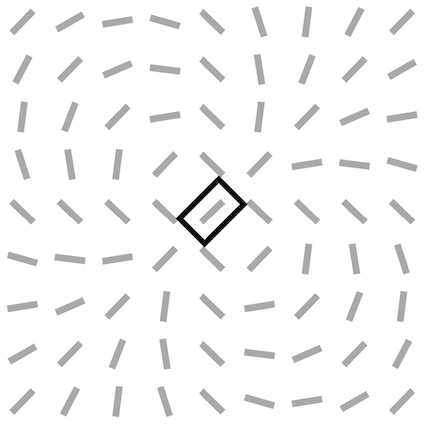} };
\draw [anchor=north west] (0.10\linewidth, 0.65\linewidth) node  {\includegraphics[width=0.8\linewidth]{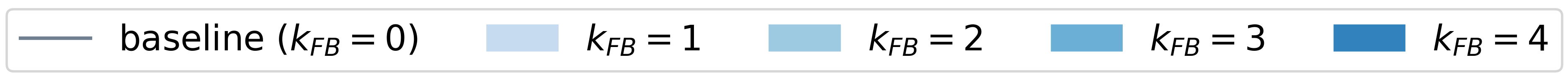}};
\begin{scope}
\draw [anchor=west,fill=white] (0.0\linewidth, 1\linewidth) node {{\bf A}};
\draw [anchor=west,fill=white] (0.33\linewidth, 1\linewidth) node {{\bf B}};
\draw [anchor=west,fill=white] (0.66\linewidth, 1\linewidth) node {{\bf C}};
\end{scope}
\end{tikzpicture}
\caption
{{\bf Relative response of V1 interaction map w.r.t. no feedback for all central preferred orientations}. {\bf (A)} In the end-zone. {\bf (B)} In the side-zone. {\bf (C)} In the center (cRF). Bars' height represent the median over all the central preferred orientations, and error bar are computed as the Median Absolute Deviation. The computation of the relative response, denoted $\boldsymbol{r_{a}}(k_\mathrm{FB})$, is detailed in Eq.~\ref{eq:eq12}. The baseline represents the relative response without feedback.}
\label{fig:fig7}
\end{figure}

We now generalize, refine and quantify these qualitative observations. We include a third region of interest, the center of the interaction map, to confirm the decreasing activity observed qualitatively at this location.  We report the median of the ratio $\boldsymbol{r_{a}}(k_\mathrm{FB})$ over all central preferred orientations, for the end-zone, the side-zone and the center. This analysis is repeated for a feedback strength ranging from $1$ to $4$  (see Fig.~\ref{fig:fig7}).
We observe an increase of the activity in the end-zone of the interaction map with feedback compared to the end-zone of the interaction without feedback (see Fig.~\ref{fig:fig7}-A). This increase is significant as quantified by all pair-wise statistical tests with the baseline ($\mathrm{WT}(N=51, p<0.01)$). For larger feedback strengths, we observe a higher activity in the end-zone which is also significant (all pair-wise statistical tests between all feedback strengths ($\mathrm{WT}(N=51, p<0.01)$). For example, in the end-zone, the median activity over all the central preferred orientations is 16\% and  25\% higher with a respective feedback strength of $1$ and $4$ compared to the median when feedback is suppressed. This suggests that the feedback signals excite neurons in the end-zone of the interaction map.
In contrast, we observe a slight decrease of activity in the side-zone of the interaction map with feedback active compared to when feedback is suppressed (see Fig.~\ref{fig:fig7}-B). The decrease compared to the baseline is significant as well as the increase between the different feedback strengths as quantified by all pair-wise statistical tests ($\mathrm{WT}(N=51, p<0.01)$).  The center of the interaction map exhibits a significant decrease in activity compared to the baseline ($\mathrm{WT}(N=51, p<0.01)$). In addition, the larger the feedback strength, the weaker the activity in the center of the interaction map ($\mathrm{WT}(N=51, p<0.01)$). For example, we report a decrease from -28\% for $k_\mathrm{FB} = 1$ to -34\% for $k_\mathrm{FB} = 4$ compared to the activity in the center of the interaction map without feedback (see Fig.~\ref{fig:fig7}-C).

\begin{figure}[h]
\centering
\begin{tikzpicture}
	\draw [anchor=north west] (0.0\linewidth, 0.99\linewidth) node {\includegraphics[width=0.8\linewidth]{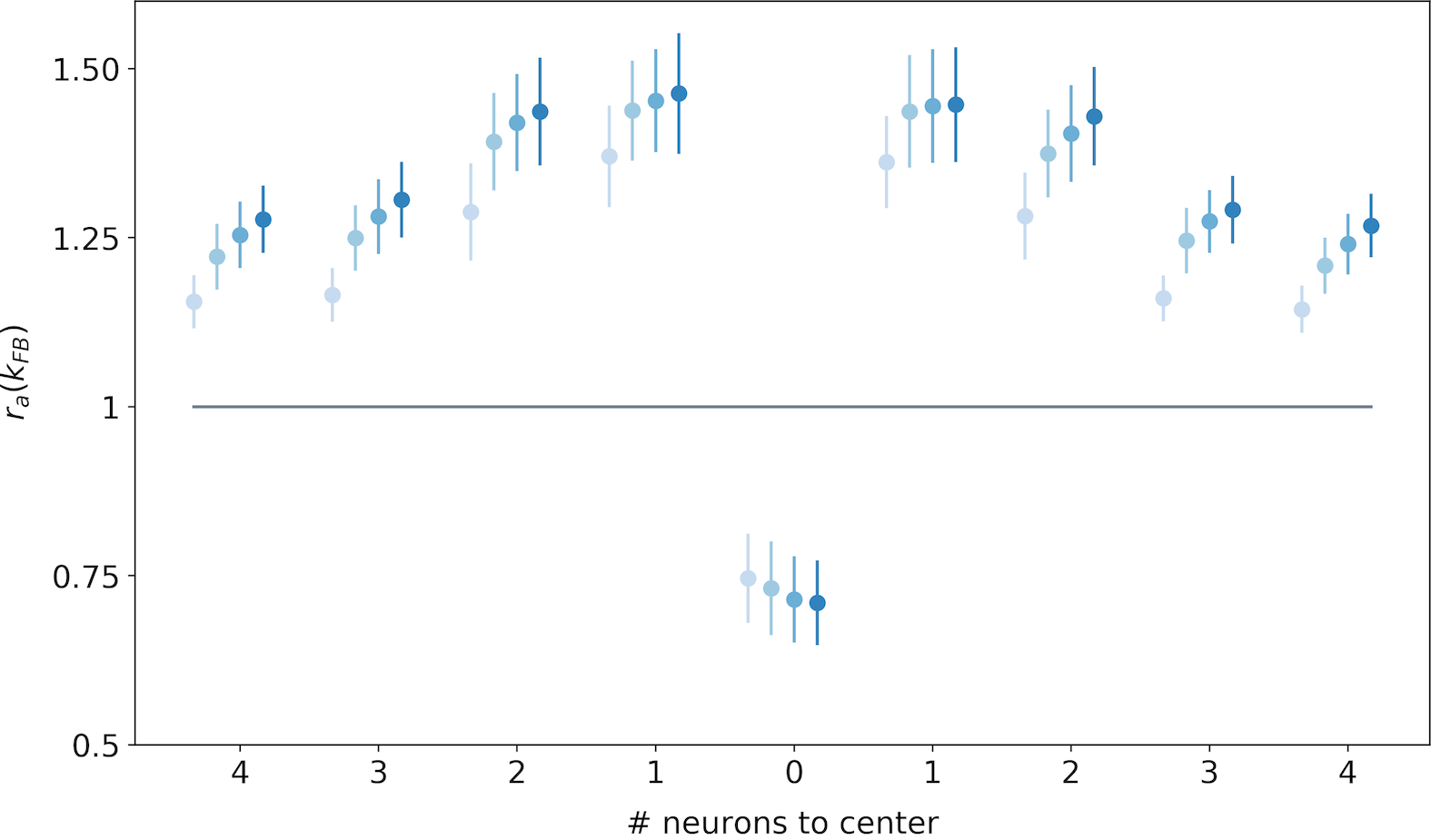}};
	\draw [anchor=north west] (0.715\linewidth, 0.655\linewidth) node {\includegraphics[width=0.08\linewidth]{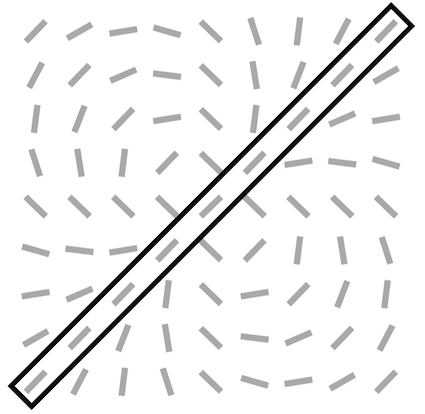} };
	\draw [anchor=north west] (0.\linewidth, 0.50\linewidth) node  {\includegraphics[width=0.8\linewidth]{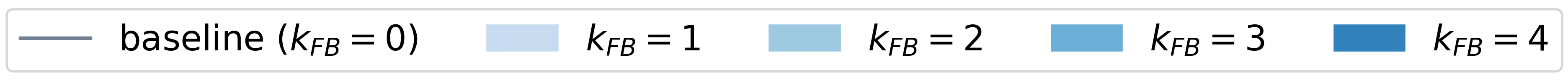}};
\end{tikzpicture}
\caption
{{\bf Relative response w.r.t. no feedback along the axis of the central preferred orientation of V1 interaction map}. Each point represents the median over all the orientations, and error bars are computed as the Median Absolute Deviation. The x-axis represents the distance, in number of neurons, to the center of the interaction map. The computation of the relative response, denoted $\boldsymbol{r_{a}}(k_\mathrm{FB})$, is detailed in Eq.~\ref{eq:eq12}. The baseline represents the relative response without feedback.}
\label{fig:fig8}
\end{figure}

We report the spatial profile of the median activity along the axis of the central preferred orientation (see Fig.~\ref{fig:fig8}).
For all distances from the center, the activity along the central preferred orientation axis of interaction map is significantly higher than the activity without feedback (all pair-wise statistical tests with the baseline: $\mathrm{WT}(N=51, p<0.01)$). The only exception is in the center of the interaction map, where the activity is weaker when feedback is active (see also Fig.~\ref{fig:fig7}-C). This inhibition in the center of the map compared to the baseline is significant as quantified with pair-wise statistical tests ($\mathrm{WT}(N=51, p<0.01)$). Even if activities for $k_\mathrm{FB}\neq0$ along the central preferred orientation are always higher than the activity with $k_\mathrm{FB}=0$, they tend to decrease with distance to the center. Especially, for $k_\mathrm{FB}=4$, the neurons located just near the center exhibit a response $+36\%$ higher than the same neurons without feedback. With the same feedback strength, this increase of activity w.r.t to no feedback is reduced to $15\%$ when the neurons are located 4 neurons away from the center. At a given position different from the center, increasing the feedback strength significantly increases the activity as quantified by all pair-wise statistical test ($\mathrm{WT}(N=51, p<0.01)$). 
Our results exhibit three different kinds of modulations in the interaction map due to feedback signals. First, the activity in the center of the map is reduced with the feedback. Second, the activity in the end-zone, and more specifically along the axis of the central preferred orientation is increased with the feedback. Third, the activity in the side-zone is reduced with the feedback.

\subsection*{Effect of the feedback at the representational level}
After investigating the effect of feedback at the lowest level of neural organization, we now explore its functional and higher-level aspects. In particular, this subsection is dealing with the denoising ability of the feedback signal.

\subsubsection*{SDPC feedback signals improve input denoising}
To evaluate the denoising ability of the feedback connection, we feed the SDPC model with increasingly more noisy images of the STL-10 and CFD databases. Then, we compare the resulting representations ($\boldsymbol{\gamma}_{i}^\mathrm{eff}$) with the original (non-degraded) image. To do this comparison, we conduct two types of experiments: a qualitative experiment that visually displays what has been represented by the model (see Fig.~\ref{fig:fig9}-A on STL-10 and Fig.~\ref{fig:fig10}-A on CFD), and a quantitative experiment measuring the similarity between representations of noisy and original images (see Fig.~\ref{fig:fig9}-B \& C on STL-10 and Fig.~\ref{fig:fig10}-B \& C on CFD). These two experiments are repeated for a noise level ($\sigma$) ranging from $0$ to $5$ and a feedback strength ($k_\mathrm{FB}$) varying from $0$ to $4$. The similarity between images is computed using the median structural similarity index (SSIM)~\cite{wang2004image} over $1200$ and $400$ images for the STL-10 and CFD database, respectively. The SSIM index varies from $0$ to $1$ such that the more similar the images, the closer the SSIM index is to $1$. For comparison, we include a baseline (see the black curves in Fig.~\ref{fig:fig7}) which is computed as the SSIM index between original and noisy images for different levels of noise.
\begin{figure}[h!]
	\centering
\begin{tikzpicture}
\draw [anchor=north west] (0.055\linewidth, 0.91\linewidth) node {\includegraphics[width=0.079\linewidth]{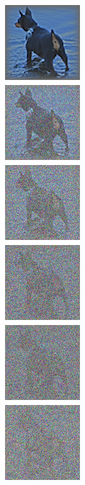}};
\draw [anchor=north west] (0.18\linewidth, 0.91\linewidth) node {\includegraphics[width=0.37\linewidth]{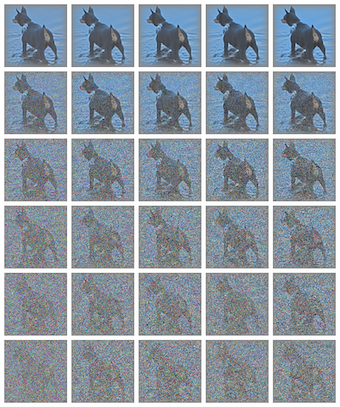}};
\draw [anchor=north west] (0.625\linewidth, 0.91\linewidth) node {\includegraphics[width=0.37\linewidth]{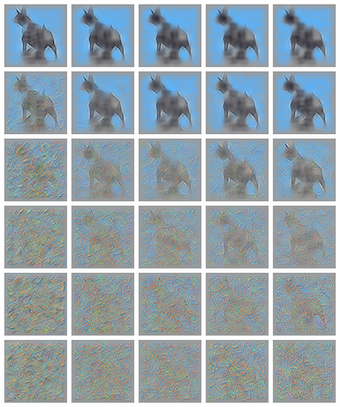}};
\draw[draw=gray, anchor=north west] (0.195\linewidth, 0.965\linewidth) rectangle (0.555\linewidth,0.90\linewidth);
\draw[draw=gray, anchor=north west] (0.64\linewidth, 0.965\linewidth) rectangle (1\linewidth,0.90\linewidth);
\draw[draw=gray, anchor=north west] (0\linewidth, 0.965\linewidth) rectangle (0.055\linewidth,0.46\linewidth);
\draw [anchor=north west] (0.1\linewidth, 0.36\linewidth) node {\includegraphics[width=0.44\linewidth]{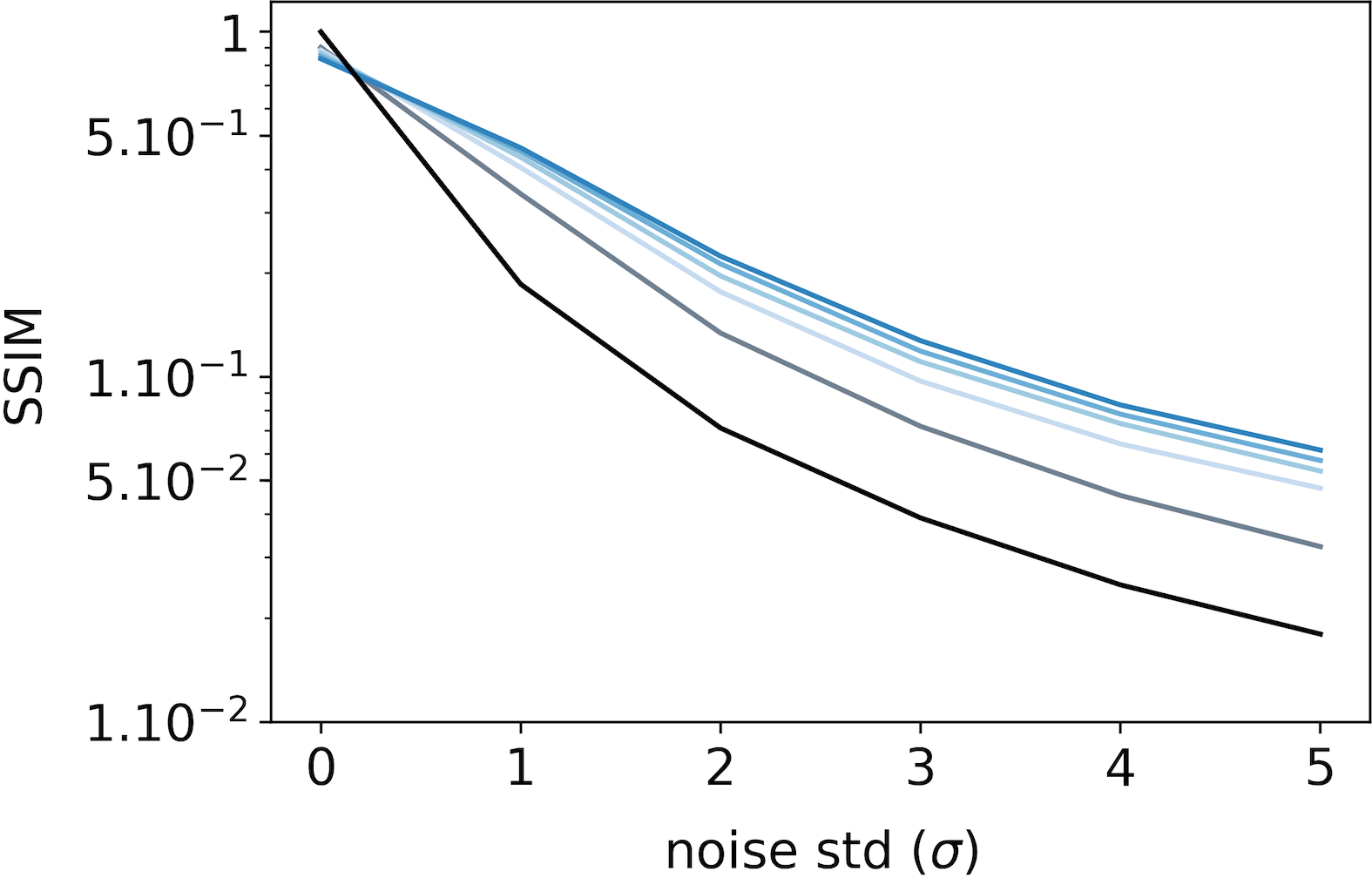}};
\draw [anchor=north west] (0.56\linewidth, 0.36\linewidth) node {\includegraphics[width=0.44\linewidth]{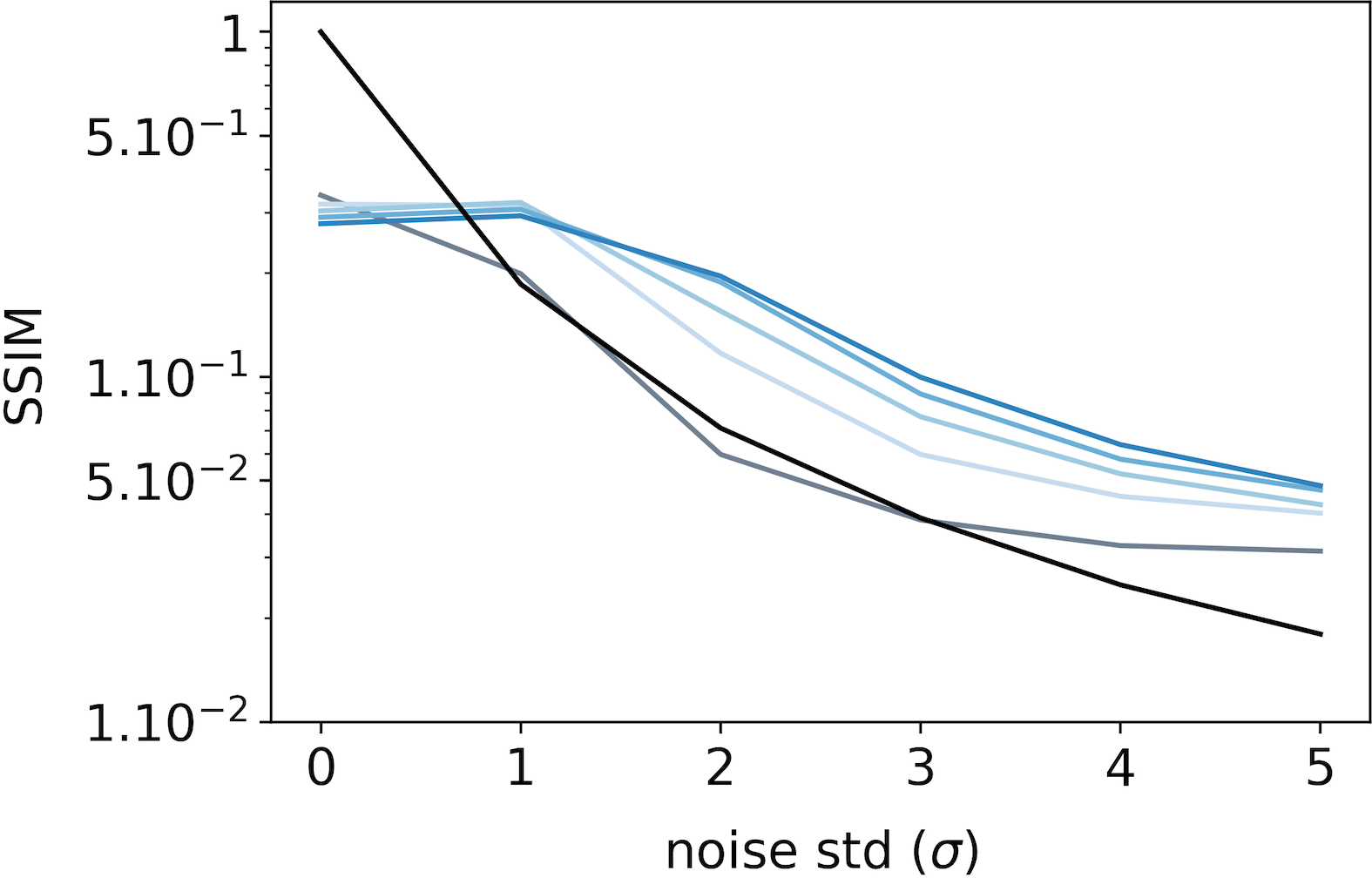}};
\draw [anchor=north west] (0.12\linewidth, 0.05\linewidth) node {\includegraphics[width=0.88\linewidth]{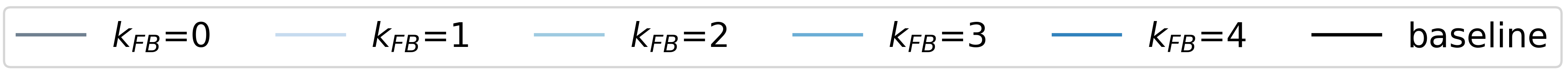}};

\begin{scope}
\draw [anchor=west,fill=white] (0.0\linewidth, 1\linewidth) node {{\bf A}};
\draw [anchor=west,fill=white] (0.22\linewidth, 0.92\linewidth) node {0};
\draw [anchor=west,fill=white] (0.29\linewidth, 0.92\linewidth) node {1};
\draw [anchor=west,fill=white] (0.36\linewidth, 0.92\linewidth) node {2};
\draw [anchor=west,fill=white] (0.43\linewidth, 0.92\linewidth) node {3};
\draw [anchor=west,fill=white] (0.51\linewidth, 0.92\linewidth) node {4};
\draw [anchor=west] (0.23\linewidth, 0.95\linewidth) node {feedback strength ($k_\mathrm{FB}$)};
\draw [anchor=west] (0.67\linewidth, 0.95\linewidth) node  {feedback strength ($k_\mathrm{FB}$)};
\draw [anchor=west,fill=white] (0.\linewidth, 0.95\linewidth) node {std};
\draw [anchor=west,fill=white] (0.\linewidth, 0.92\linewidth) node {($\sigma$)};
\draw [anchor=west,fill=white] (0.665\linewidth, 0.92\linewidth) node {0};
\draw [anchor=west,fill=white] (0.73\linewidth, 0.92\linewidth) node {1};
\draw [anchor=west,fill=white] (0.805\linewidth, 0.92\linewidth) node {2};
\draw [anchor=west,fill=white] (0.88\linewidth, 0.92\linewidth) node {3};
\draw [anchor=west,fill=white] (0.95\linewidth, 0.92\linewidth) node {4};
\draw [anchor=west,fill=white] (0.007\linewidth, 0.875\linewidth) node {0};
\draw [anchor=west,fill=white] (0.007\linewidth, 0.79\linewidth) node {1};
\draw [anchor=west,fill=white] (0.007\linewidth, 0.71\linewidth) node {2};
\draw [anchor=west,fill=white] (0.007\linewidth, 0.645\linewidth) node {3};
\draw [anchor=west,fill=white] (0.007\linewidth, 0.575\linewidth) node {4};
\draw [anchor=west,fill=white] (0.007\linewidth, 0.50\linewidth) node {5};
\draw [anchor=west,fill=white] (0.07\linewidth, 0.44\linewidth) node {Input};
\draw [anchor=west,fill=white] (0.18\linewidth, 0.44\linewidth) node {Reconstructed by layer 1 ($\boldsymbol{\gamma}_{1}^\mathrm{eff}$)};
\draw [anchor=west,fill=white] (0.625\linewidth, 0.44\linewidth) node {Reconstructed by layer 2 ($\boldsymbol{\gamma}_{2}^\mathrm{eff}$)};
\draw [anchor=west,fill=white] (0.15\linewidth, 0.38\linewidth) node {{\bf B}};
\draw [anchor=west,fill=white] (0.60\linewidth, 0.38\linewidth) node {{\bf C}};
\end{scope}
\end{tikzpicture}
\caption
{{\bf Effect of the feedback strength on noisy images from STL-10 database.} {\bf(A)} In the left block, one image is corrupted by Gaussian noise of mean $0$ and standard deviation ($\sigma$) varying from $0$ to $5$. The central block exhibits the representations made by the first layer ($\boldsymbol{\gamma}_{1}^\mathrm{eff}$), and the right-hand block the representations made by the second layer ($\boldsymbol{\gamma}_{2}^\mathrm{eff}$). Within each of these blocks, the feedback strength ($k_\mathrm{FB}$) is ranging from $0$ to $4$ in columns {\bf(B)} We plot the SSIM index (higher is better) between original images and their representation by the first layer of the SDPC. {\bf(C)} We plot the SSIM index between original images and their representation by the second layer of the SDPC. All curves represent the median SSIM over $1200$ samples of the testing set and present a logarithmic scale on the y-axis. The color code corresponds to the feedback strength, from grey for $k_\mathrm{FB}=0$ to darker blue for higher feedback strength. The black line is the baseline, it is the SSIM between the noisy and original input images.}
\label{fig:fig9}
\end{figure}
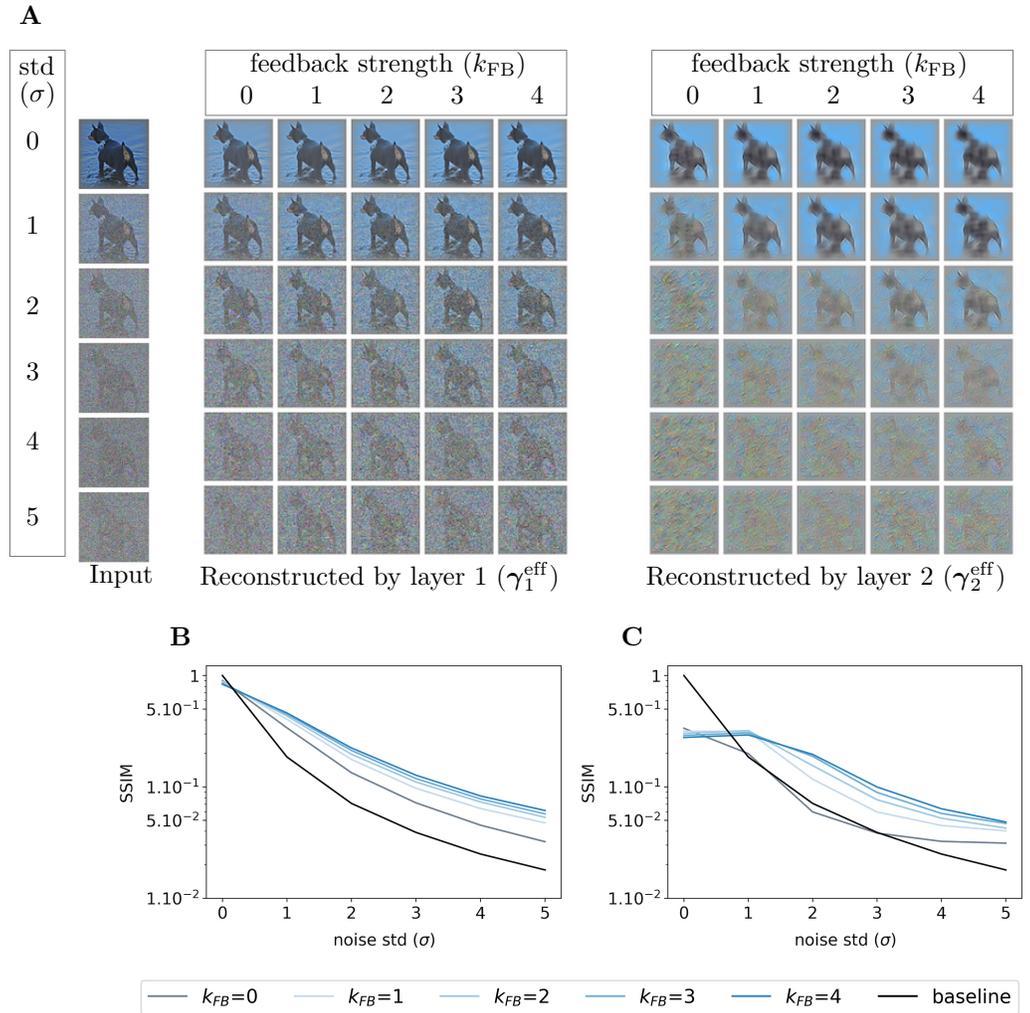
\begin{figure}[h!]
	\centering
\begin{tikzpicture}
\draw [anchor=north west] (0.055\linewidth, 0.91\linewidth) node {\includegraphics[width=0.085\linewidth]{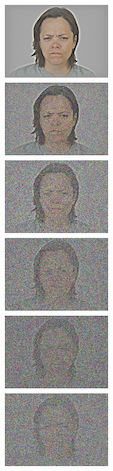}};
\draw [anchor=north west] (0.16\linewidth, 0.91\linewidth) node {\includegraphics[width=0.41\linewidth]{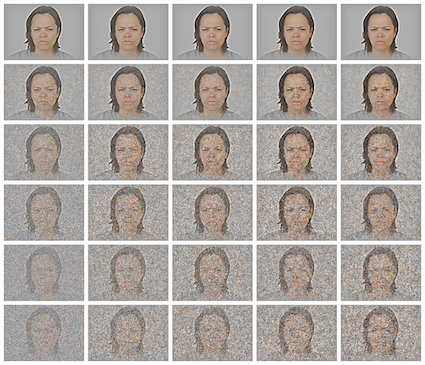}};
\draw [anchor=north west] (0.58\linewidth, 0.91\linewidth) node {\includegraphics[width=0.41\linewidth]{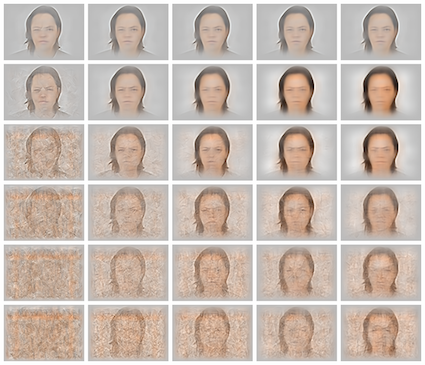}};
\draw[draw=gray, anchor=north west] (0.175\linewidth, 0.965\linewidth) rectangle (0.575\linewidth,0.90\linewidth);
\draw[draw=gray, anchor=north west] (0.595\linewidth, 0.965\linewidth) rectangle (0.995\linewidth,0.90\linewidth);
\draw[draw=gray, anchor=north west] (0\linewidth, 0.965\linewidth) rectangle (0.055\linewidth,0.55\linewidth);
\draw [anchor=north west] (0.1\linewidth, 0.44\linewidth) node {\includegraphics[width=0.44\linewidth]{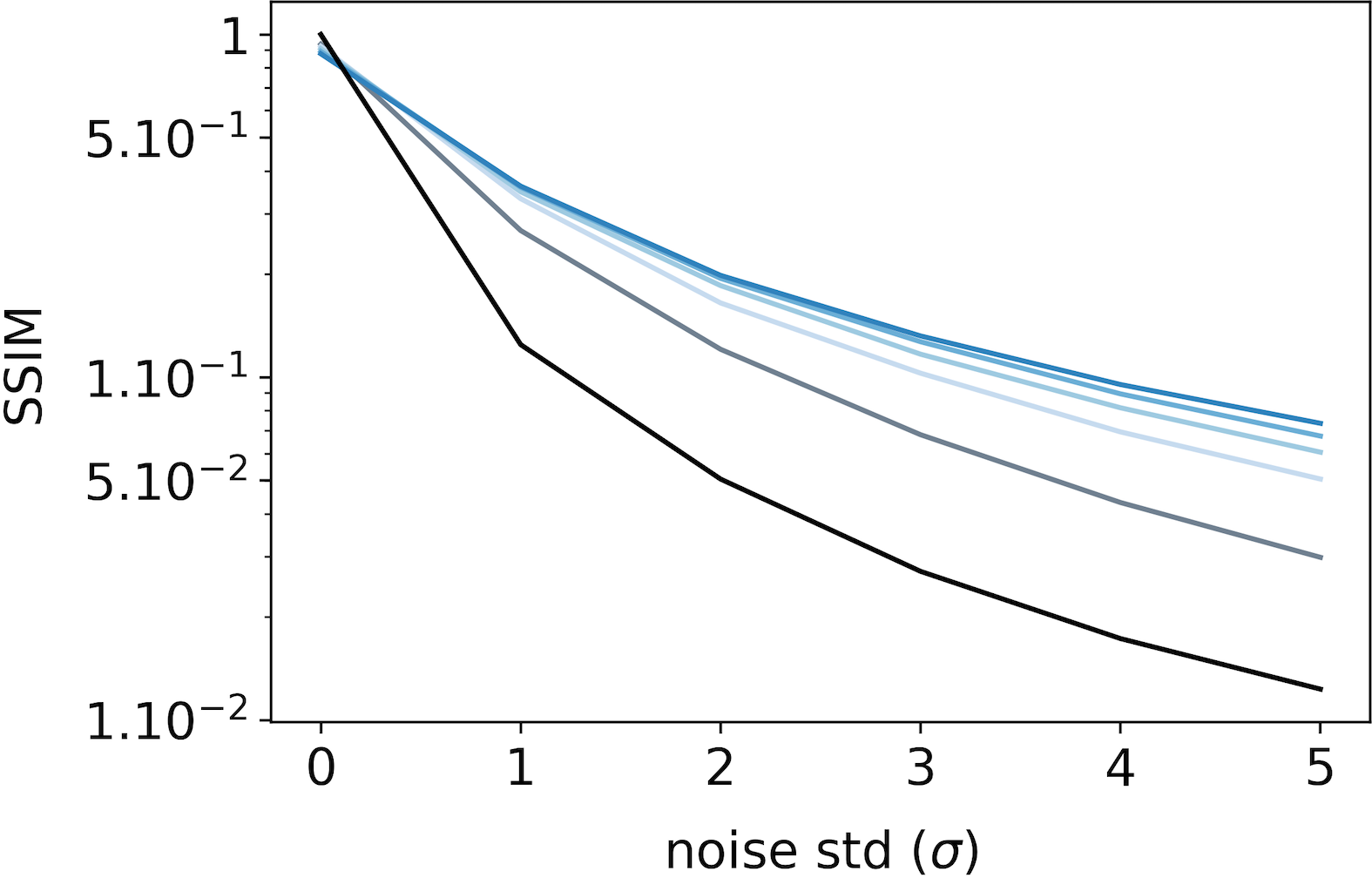}};
\draw [anchor=north west] (0.56\linewidth, 0.44\linewidth) node {\includegraphics[width=0.44\linewidth]{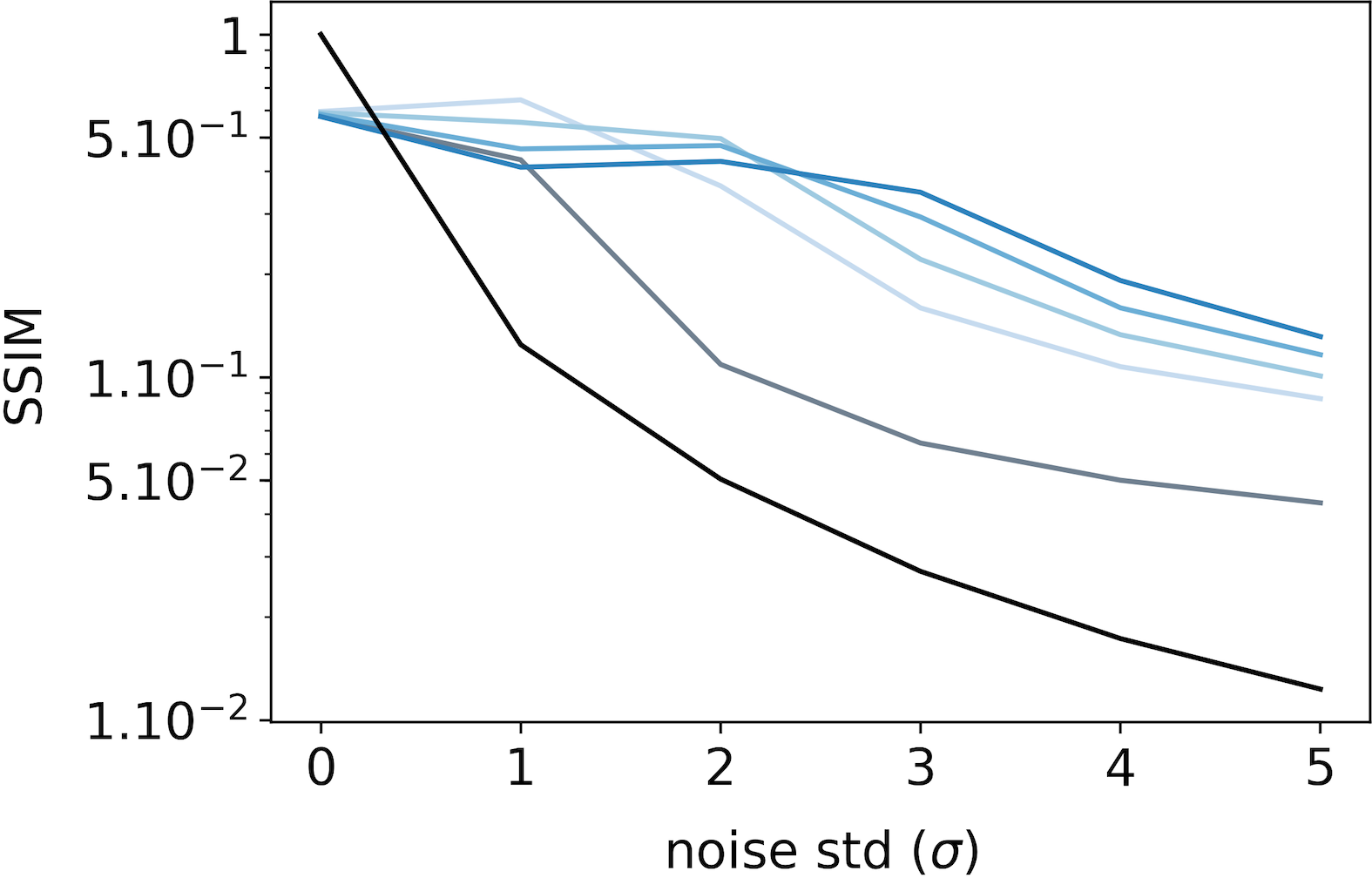}};
\draw [anchor=north west] (0.12\linewidth, 0.15\linewidth) node {\includegraphics[width=0.88\linewidth]{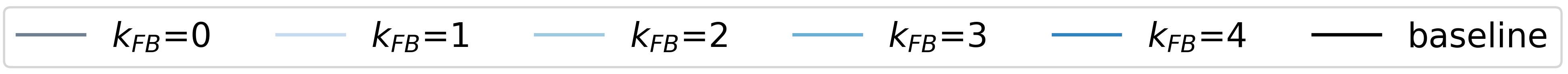}};

\begin{scope}
\draw [anchor=west,fill=white] (0.0\linewidth, 1\linewidth) node {{\bf A}};
\draw [anchor=west,fill=white] (0.20\linewidth, 0.92\linewidth) node {0};
\draw [anchor=west,fill=white] (0.28\linewidth, 0.92\linewidth) node {1};
\draw [anchor=west,fill=white] (0.36\linewidth, 0.92\linewidth) node {2};
\draw [anchor=west,fill=white] (0.44\linewidth, 0.92\linewidth) node {3};
\draw [anchor=west,fill=white] (0.52\linewidth, 0.92\linewidth) node {4};
\draw [anchor=west] (0.23\linewidth, 0.95\linewidth) node {feedback strength ($k_\mathrm{FB}$)};
\draw [anchor=west] (0.66\linewidth, 0.95\linewidth) node  {feedback strength ($k_\mathrm{FB}$)};
\draw [anchor=west,fill=white] (0.\linewidth, 0.95\linewidth) node {std};
\draw [anchor=west,fill=white] (0.\linewidth, 0.92\linewidth) node {($\sigma$)};
\draw [anchor=west,fill=white] (0.62\linewidth, 0.92\linewidth) node {0};
\draw [anchor=west,fill=white] (0.70\linewidth, 0.92\linewidth) node {1};
\draw [anchor=west,fill=white] (0.78\linewidth, 0.92\linewidth) node {2};
\draw [anchor=west,fill=white] (0.86\linewidth, 0.92\linewidth) node {3};
\draw [anchor=west,fill=white] (0.94\linewidth, 0.92\linewidth) node {4};
\draw [anchor=west,fill=white] (0.007\linewidth, 0.87\linewidth) node {0};
\draw [anchor=west,fill=white] (0.007\linewidth, 0.815\linewidth) node {1};
\draw [anchor=west,fill=white] (0.007\linewidth, 0.76\linewidth) node {2};
\draw [anchor=west,fill=white] (0.007\linewidth, 0.705\linewidth) node {3};
\draw [anchor=west,fill=white] (0.007\linewidth, 0.64\linewidth) node {4};
\draw [anchor=west,fill=white] (0.007\linewidth, 0.58\linewidth) node {5};
\draw [anchor=west,fill=white] (0.07\linewidth, 0.53\linewidth) node {Input};
\draw [anchor=west,fill=white] (0.18\linewidth, 0.53\linewidth) node {Reconstructed by layer 1 ($\boldsymbol{\gamma}_{1}^\mathrm{eff}$)};
\draw [anchor=west,fill=white] (0.60\linewidth, 0.53\linewidth) node {Reconstructed by layer 2 ($\boldsymbol{\gamma}_{2}^\mathrm{eff}$)};
\draw [anchor=west,fill=white] (0.15\linewidth, 0.46\linewidth) node {{\bf B}};
\draw [anchor=west,fill=white] (0.60\linewidth, 0.46\linewidth) node {{\bf C}};
\end{scope}
\end{tikzpicture}
\caption
{{\bf Effect of the feedback strength on noisy images from CFD database.} This figure description is similar to the description of the Fig.~\ref{fig:fig9}. For the CFD database, all presented curves represent the median SSIM over $400$ samples of the testing set. }
\label{fig:fig10}
\end{figure}

We first observe that whatever the feedback strength, the first layer representations of the original image (first row, column 2 to 6 in Fig.~\ref{fig:fig9}-A for STL-10 and Fig.~\ref{fig:fig10}-A for CFD) are relatively similar to the input image itself. This observation is supported by a SSIM index close to $0.9$ for all feedback strengths ($\sigma = 0$ in Fig.~\ref{fig:fig9}-B for STL-10 and  Fig.~\ref{fig:fig10}-B for CFD). On the contrary, second layer representations look more sketchy and exhibited fewer details than the image they represent (first row, column 7 to 11 in Fig.~\ref{fig:fig9}-A for STL-10 and Fig.~\ref{fig:fig10}-A for CFD). This is also quantitatively backed by a SSIM index fluctuating around $0.4$ for the STL-10 database ($\sigma = 0$ in Fig.~\ref{fig:fig9}-C) and $0.6$ for the CFD database ($\sigma = 0$ in Fig.~\ref{fig:fig10}-C). Interestingly, when input images are corrupted with noise (i.e. when $\sigma \geq 1$), and whatever the feedback strength, first layer representations systematically exhibit higher SSIM index than the baseline (Fig.~\ref{fig:fig9}-B for STL-10 and Fig.~\ref{fig:fig10}-B for CFD). This denoising ability of the SDPC, even without feedback is significant as reported by the pair-wise statistical tests with the baseline for both databases ($\mathrm{WT}(N=1200, p<0.01)$ for STL-10 database, and $\mathrm{WT}(N=400, p<0.01)$ for the CFD database). More importantly, the higher the feedback strength, the higher the SSIM index. In particular, on the STL-10 database, when the input is highly degraded by noise ($\sigma = 5$), the SSIM is $0.02$ for the baseline, $0.03$ for $k_\mathrm{FB}=0$, $0.05$ for $k_\mathrm{FB}=1$ and $0.06$ for $k_\mathrm{FB}=4$ (see Fig.~\ref{fig:fig9}-B). This improvement of the denoising ability with higher feedback strength when $\sigma=5$ is also significant as quantified by the pair-wise statistical tests between all feedback strength ($\mathrm{WT}(N=1200, p<0.01)$).  The inter-image variability of the SSIM of first layer representation as quantified by the MAD is low compared to the median SSIM on the STL-10 database (see Fig.~\ref{fig:figSD5}-C). In the CFD database, for a highly degraded input ($\sigma = 5$), the SSIM is $0.01$ for the baseline, $0.03$ for $k_\mathrm{FB}=0$, $0.05$ for $k_\mathrm{FB}=1$ and $0.07$ for $k_\mathrm{FB}=4$.  On the CFD database, the increases of the first layer SSIM with the feedback strength when inputs are highly degraded ($\sigma = 5$) are significative as measured by all the pair-wise statistical tests between all feedback strength ($\mathrm{WT}(N=400, p<0.01)$). Inter-image variability of the SSIM of the first layer representation is also lower than the corresponding median on the STL-10 database~(see Fig.~\ref{fig:figSD6}-C).
It is interesting to observe that even if the second layer representations are less detailed and more sketchy than the first layer reconstructions, they offer a piece of valuable information, in the form of feedback signals, that allow these first layer to better denoise the input.
%
To conclude this subsection, we conducted a qualitative and a quantitative analysis of the denoising ability of the feedback connection. Our results suggest that feedback improves the denoising ability of the first layer. Especially, as feedback gets stronger, then the first layer is more able to properly recover from a degraded input.





\section*{Discussion}
Herein, we have conducted computational experiments on a 2-layered Sparse Deep Predictive Coding (SDPC) model. The SDPC leverages feedforward and feedback connections into a model combining sparse coding and predictive coding. As such, the SDPC learns the causes (i.e. the features) and infers the hidden states (i.e. the activity maps) that best describe the hierarchical generative model giving rise to the visual stimulus (see. Fig.~\ref{fig:fig8} for an illustration of this hierarchical model and Eq.~\ref{eq:eq0} for its mathematical description).

\ifnum \plotfig=1
\begin{figure}[h]
	\centering
\begin{tikzpicture}
\draw [anchor=north west] (0.05\linewidth, 1\linewidth) node {\includegraphics[width=1\linewidth]{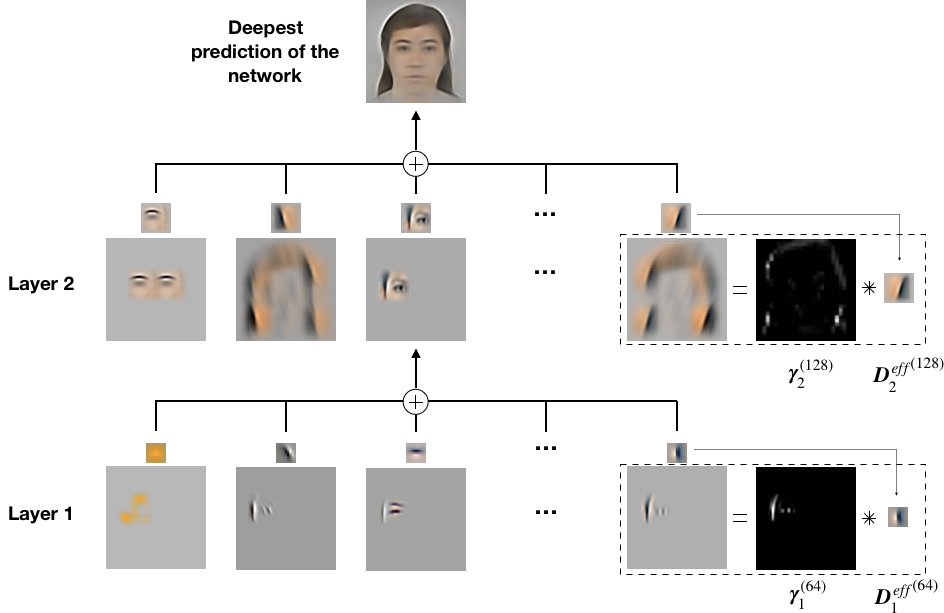}};
\end{tikzpicture}
\caption
{Illustration of the hierarchical generative model learned by the SDPC model on the AT\&T database. The deepest prediction (first row) is viewed as the sum of the features prediction (the second row). These feature predictions are computed as the convolution between one channel of $\boldsymbol{\gamma}_{2}$  and the corresponding features in $\boldsymbol{D}_{2}$. Similarly, the eyes can be decomposed using  $\boldsymbol{\gamma}_{1}$  and $\boldsymbol{D}_{1}$(the third row)}
\label{fig:fig11}
\end{figure}
\fi

We use this model of the early visual cortex to assess the effect of the early feedback connection (i.e feedback from V2 to V1) through different levels of analysis. At the neural level, we have shown that feedback connections tend to recruit more neurons in the first layer of the SDPC. We have introduced the concept of interaction map to describe the neural organization in our V1-model. Interestingly, the interaction maps generated when natural images are presented to the model are very similar to association fields. In addition, interaction maps allow us to describe the neural reorganization due to feedback signals. In particular, we have observed that feedback signals align neurons in the side-zone of the interaction map co-linearly to the central preferred orientation. At the activity level, we observed $3$ main kind of feedback modulatory effects. First, the activity in the center of the interaction map is decreased. Second, the activity in the end-zone and more specifically along the axis of the central preferred orientation is increased. Third, the activity in the side-zone is reduced. At the representational level, we have investigated the role of feedback signals when input images are degraded using Gaussian noise. We have demonstrated that higher feedback strengths allow better denoising ability. In this section, we interpret these computational findings in light of current neuroscientific knowledge. 

\subsection*{SDPC learns cortex-like RFs while performing neuro-plausible computation}
The SDPC model satisfies some of the computational constraints that are thought to occur in the brain: notably local computation and plasticity~\cite{whittington2017approximation}. The locality of the computation is ensured by Eq.~\ref{eq:update_inference}: the new state of a neural population (whose activity is represented by $\boldsymbol{\gamma}_{i}^{t+1}$) only depends on its previous state ($\boldsymbol{\gamma}_{i}^{t}$), the state of adjacent layers ($\boldsymbol{\gamma}_{i-1}^{t}$ and $\boldsymbol{\gamma}_{i+1}^{t}$) and the associated synaptic weights ($\boldsymbol{D}_i$ and $\boldsymbol{D}_{i+1}$). Similarly, the update of the synaptic weights ($\boldsymbol{D}_i$), described by Eq.~\ref{eq:eq_learning}, is exclusively based on the pre-synaptic and post-synaptic activity (respectively $\boldsymbol{\gamma}_{i-1}$ and $\boldsymbol{\gamma}_i$). The learning process could then be assimilated to an Hebbian learning. Not only the processing but also the result of the training exhibits tight connections with neuroscience. The first-layer RFs (Fig.~\ref{fig:fig2}-B for the STL-10 database and Fig.~\ref{fig:fig2}-G for the CFD database) are similar to the V1 simple-cells RFs, which are oriented Gabor-like filters~\cite{jones1987evaluation, jones1987two}. Olshausen \& Field have already demonstrated, in a shallow network, that oriented Gabor-like filters emerge from sparse coding strategies~\cite{olshausen1997sparse}, but to the best of our knowledge, this is the first time that such filters are exhibited in a 2-layers network combining Predictive Coding and Sparse Coding. This architecture allows us to observe an increase in specificity of the neuron's RFs with the depth of the network. This observation is even more striking when the SDPC is trained on the CFD database, which presents less variability compared to the STL-10 database. On CFD, second layer RFs exhibit features that are highly specific to faces (eyes, mouth, eyebrows, contours of the face). Interestingly, it was demonstrated with neurophysiological experiments that neurons located in deeper regions of the central visual stream are also sensitive to that particular face features~\cite{henriksson2015faciotopy, leder2004part}.

\subsection*{Functional interpretation of the observed V1 interaction maps}
At the electrophysiological level, it has been demonstrated that as early as in the V1 area, feedback connections from V2 could either facilitate~\cite{gilbert2013top} or suppress~\cite{nurminen2018top} lateral interactions. In particular, these modulations help V1 neurons to integrate contextual information from a larger part of the visual field and play a crucial role in contour integration~\cite{gilbert2000interactions}. It was assumed that association fields were represented in V1 to perform such a contour integration~\cite{gilbert2012adult}. Interestingly, the SDPC first-layer interaction maps exhibited a co-linear and co-circular neural organization very similar to association fields even without feedback (see Fig.~\ref{fig:fig4}). We formulate the hypothesis that this specific organization is mainly related to the statistics of edge co-occurrence in natural images~\cite{geisler2001edge}. Nevertheless, the modulation of neural activity within the interaction map mediated by feedback goes towards a better contour integration. Indeed, the increase of activity in the end-zone and the decrease of activity in the side-zone seem to be optimal to integrate smooth and close contours~\cite{kovacs1994perceptual} (see Fig.~\ref{fig:fig6}, Fig.~\ref{fig:fig7} and Fig.~\ref{fig:fig8}). In addition, the organizational feedback modulation in the interaction map revealed that feedback signals tend to reorganize the side-zone to promote orientations that are co-linear to the central preferred orientation (see Fig.~\ref{fig:fig5}). This organization may provide an optimal substrate to integrate dynamic stimuli along two axes: a parallel one with apparent motion-like sequence of oriented stimuli moving along the end-zone direction~\cite{pham2001mean,chavane2000visual,georges2002orientation}, and a perpendicular one for oriented stimuli moving perpendicular to their orientation. Interestingly, oriented Gabors moving in an apparent sequence along the parallel axis are perceived as faster as Gabors moving along the orthogonal axis~\cite{georges2002orientation}. Furthermore, aligning the side-zone region to the preferred orientation could also contribute to the aperture problem~\cite{wuerger1996visually} and the observed bias for perceiving oriented bars as moving in direction perpendicular to their orientation~\cite{lorenceau1993different, montagnini2007bayesian}.

\subsection*{Do lateral interactions increase the sparseness of neural activity in V1?}
In this paper, we have assumed that recurrent internal processing could be modeled using sparse coding. Is it a realistic hypothesis? One of the main roles of sparse coding is to enforce competition among neurons: it suppresses weakly activated neurons to promote strongly activated ones. In other words, sparse coding performs 'explaining away'. Interestingly, when $2$ stimuli (blobs) were presented at different locations and timings, it has been observed in monkeys' area V1 that a suppressive wave tends to spatially disambiguate the positions of the $2$ stimuli~\cite{chemla2019suppressive}. This effect was attributed to lateral interactions (due to the spatio-temporal properties of the effect) and can be thought as such an explaining away mechanism. Other studies have also demonstrated that lateral interactions exacerbate competition in cortical columns with different orientations or ocular dominance~\cite{crook1998evidence, crook2002gaba, piepenbrock1999role}. Therefore, our sparse coding model accounts for one possible function of lateral interaction. Nevertheless, the sparse coding algorithm we use  (i.e. FISTA, see section~'\nameref{Sec:Methods}') doesn't allow to explicitly learn a lateral connectivity matrix. Consequently, one might consider other sparse coding algorithms including lateral connection weights to provide a more accurate model of cortical columns~\cite{gregor2011structured}.

\subsection*{SDPC feedback accounts for object processing in V1 with degraded images}
Psychophysical experiments using backward masking demonstrated that categorization performances were substantially impaired when a mask followed a highly degraded stimulus (by occlusion or contrast reduction)~\cite{wyatte2012limits, wyatte2014early}. This suggests that feedback is crucial to recognize degraded image. In this paper, we demonstrate a similar result by assessing the representations of the first layer of the SDPC when the model is fed with increasingly more noisy images and with different feedback strengths. In particular, we show that feedback connections from V2 to V1 have the ability to denoise corrupted images (see Fig.~\ref{fig:fig10} and Fig.~\ref{fig:fig11}). In addition, the previously mentioned psychophysical studies suggest that feedback connections are not bringing any change in recognition accuracy when non-degraded images are presented to the subjects~\cite{wyatte2012limits, wyatte2014early}. In contrast, the SSIM between the original image and the first layer representation when the SDPC is fed with non-degraded image exhibits a slight decrease, but significant enough, when the feedback strength is increased. We formulate the hypothesis that this discrepancy is mainly coming from the fixed value we give to the feedback strength parameter.  Feedback is strongly subject to attentional modulation, and a recent study has suggested that attention can be understood as a mechanism weighting feedback connections using the level of uncertainty~\cite{feldman2010attention}. Therefore, one might consider replacing the scalar $k_\mathrm{FB}$ with a precision matrix proportional to the prediction error at each layer. Said differently, if the first layer prediction error is very low, feedback signals should be weak as no higher-layer information are needed to faithfully represent the sensory input. On the contrary, if the first layer prediction error is high, the feedback connection should be strong enough to bring additional information from higher-layer to compensate for the high uncertainty in the first layer. Extending the SDPC framework with a finer-grained feedback weighting should greatly improve its performance and tighten that particular link with neuroscience.

\subsection*{Concluding remarks}
In this study, we have shown that the first layer of the SDPC model represents the visual input similarly to V1. We have also demonstrated that feedback from V2 may modulate the interaction map in such a way to promote contour integration. This improvement in contour integration with feedback strength resulted in a better representation when noisy images were presented to the SDPC. Note that the proposed SDPC is a simplified version of perceptual inference models based on free-energy optimization~\cite{friston2010free, friston2008hierarchical}. While free-energy estimates the entire distribution of error and prediction signals, our SDPC only assesses their most likely values. One interesting perspective would be to extend the SDPC to make it fit the precision-weighted message passing implemented in the free-energy framework. In general, we foresee great perspectives to such a description of the brain both in computational neuroscience to understand perceptual mechanisms and in artificial intelligence for tasks like denoising, classification or inpainting.

\section*{Model and Methods}\label{Sec:Methods}
In this section, we detail the Sparse Deep Predictive Coding (SDPC) model. We first explain how the SDPC is directly related to the Predictive Coding (PC) theory. Next, we describe the mathematics behind the inference and the learning process. We then explicit the back-projection mechanism used to interpret and visualize inference and learning results.  We also describe the databases and the network parameters we adopted to train the SDPC. Finally, we detailed all the calculations needed to generate interaction maps.

\subsection*{SDPC Model}
\subsubsection*{From Predictive Coding to Sparse Deep Predictive Coding}
Fig.~\ref{fig:fig_SDPC} shows the architecture of a 2-layered SDPC model that takes an image $\boldsymbol{x}$ as an input. As the SDPC is relying on the Predictive Coding (PC) theory~\cite{rao1999predictive}, it is continuously generating top-down predictions such that the neural population at one level ($\boldsymbol{\gamma}_{i}$) predicts the neural activity at the lower level ($\boldsymbol{\gamma}_{i-1}$). The prediction from a higher level is sent through a feedback connection to be compared to the actual neural activity. This elicits a prediction error, $\boldsymbol{\epsilon}_{i}$, that is forwarded to the following layer to update the population activity towards improved prediction. This dynamical process repeats throughout the hierarchy until the bottom-up process no longer conveys any new information. We force the weights of the feedforward connection ($\mathbf{D}_{i}$) to be reciprocal to the weights of the feedback connection ($\mathbf{D}_{i}^{T}$)~\cite{rao1999predictive, spratling2017hierarchical}. We also impose a convolutional structure to $\mathbf{D}_{i}$ to strengthen the proximity with the overlapping RFs observed in the visual cortex. For the sake of clarity, in the mathematical description of our model, we replaced the convolution by a matrix product. The rigorous equivalence between discrete convolution and matrix product could be easily demonstrated by transforming $\mathbf{D}_{i}$ into a Toeplitz-structured matrix. Mathematically, the SDPC solves the hierarchical inverse problem formulated in Eq.~\ref{eq:eq0} by minimizing the loss function $\mathcal{L}$ defined in Eq.~\ref{eq:eq1}. This optimization process is separated in two different but related steps: inference and dictionary learning. The inference process consists in finding a sparse activity map of the input considering the synaptic weights are fixed. Once the activity map has been estimated the next step is to update the synaptic weights to better fit the dataset. We iterate these two processes until the convergence is reached.

\subsubsection*{Inference}
To obtain a convex cost, we relax the $\ell_{0}$ constraint in Eq.~\ref{eq:eq0} into a $\ell_{1}$-penalty. It defines, therefore, a loss function that could be minimized using first-order methods like Iterative Shrinkage Thresholding Algorithms (ISTA)~\cite{beck2009fast}. ISTA is proven to be computationally cheap and offers fast convergence rate. In practice, we use an accelerated version of the ISTA algorithm called FISTA. One inference step used to update $\boldsymbol{\gamma}_i$ is shown in Eq.~\ref{eq:update_inference}

\begin{equation}
    \label{eq:update_inference}
    \begin{array}{ll}
    \boldsymbol{\gamma}_{i}^{t+1} & =  \mathcal{T}_{\eta_{c_{i}} \lambda_i}\big(\boldsymbol{\gamma}_{i}^{t} - \eta_{c_{i}} \displaystyle \frac{\partial  \mathcal{L}}{\partial \boldsymbol{\gamma}_{i}^{t}}\big)\\
    & = \mathcal{T}_{\eta_{c_{i}} \lambda_i}\big(\boldsymbol{\gamma}_{i}^{t}+\eta_{c_{i}} \mathbf{D}_{i}(\boldsymbol{\gamma}_{i-1}^{t} - \mathbf{D}_{i}^{T}\boldsymbol{\gamma}_{i}^{t}) - k_\mathrm{FB} \cdot \eta_{c_{i}}(\boldsymbol{\gamma}_{i}^{t} - \mathbf{D}_{i+1}^{T}\boldsymbol{\gamma}_{i+1}^{t})\big)
    \end{array}
\end{equation}

In Eq.~\ref{eq:update_inference}, $\mathcal{T}_{\alpha}$ denotes a non-negative soft thresholding operator, as defined in Eq.~\ref{eq:soft_thresholding}. $\eta_{c_{i}}$ is the learning rate of the inference process, it is computed as the inverse of the largest eigenvalue of $\mathbf{D}^{T}_{i} \mathbf{D}_{i}$~\cite{beck2009fast}. $\boldsymbol{\gamma}_{i}^{t}$ is the state of the neural population at layer $i$ and time $t$. $\lambda_i$ is a scalar tuning the sparsity of the activity map $\boldsymbol{\gamma}_{i}$.
\begin{equation}
    \label{eq:soft_thresholding}
    \mathcal{T}_{\alpha} (x) = \left\{
    \begin{array}{ll}
    x -  \alpha & \textnormal{if} \mQuad x \geq \alpha \\
    0 &  \textnormal{if} \mQuad x \leq \alpha\\
    \end{array}
    \right.
\end{equation}

Fig.~\ref{fig:fig_SDPC} shows how we can interpret the update scheme described in Eq.~\ref{eq:update_inference} as one loop of the inference process of a recurrent layer. This recurrent layer forms the building block of the Sparse Deep Predictive Coding (SDPC) network (see Algo.S~\ref{algo:pseudo_code} for the complete pseudo-code of the SDPC inference process). We initialize all the activity maps $\boldsymbol{\gamma}_{i}^{t}$ to zero at the beginning of the inference process. We consider the inference process is finalized once all the activity maps have reached a stability criterion. Our stability criterion consists in a threshold ($T_{stab}$) on the relative variation of each activity map (Eq.~\ref{eq:stability_criterion}).
\begin{equation}
    \label{eq:stability_criterion}
    \boldsymbol{\gamma}_{i}^t \mQuad \textnormal{is stable if}\mQuad \frac{\Vert \boldsymbol{\gamma}_{i}^t - \boldsymbol{\gamma}_{i}^{t-1}\Vert_{2} }{\Vert \boldsymbol{ \gamma}_{i}^t \Vert_{2}} < T_{stab}
\end{equation}

\subsubsection*{Dictionary learning}
 The SDPC learns the synaptic weights using a Stochastic Gradient Descent (SGD) on $\mathcal{L}$. Eq.~\ref{eq:eq_learning} describes on step of the dictionary learning process.
 \begin{equation}
    \label{eq:eq_learning}
   \begin{array}{ll}
  \mathbf{D}_{i}^{t+1} & = \mathbf{D}_{i}^{t} - \eta_{L_{i}} \frac{\displaystyle \partial \mathcal{L}}{\displaystyle \partial \mathbf{D}_{i}} \\
   & = \mathbf{D}_{i}^{t} + \eta_{L_{i}} \boldsymbol{\gamma}_{i}^{T}({\boldsymbol{\gamma}_{i-1}-\mathbf{D}_{i}^{t}}^{T}\boldsymbol{\gamma}_{i})
   \end{array}
\end{equation}
 In Eq.~\ref{eq:eq_learning}, $\mathbf{D}_{i}^{t}$ is the set of synaptic weights at time step $t$ and $\eta_{L_{i}}$ is its learning rate. 
At the beginning of the learning, all weights are initialized using the standard normal distribution (mean $0$ and variance $1$). The learning step takes place after the convergence of the inference process is achieved (see Algo.~\ref{AlgoAlternation}). It was demonstrated that this alternation of inference and learning offers reasonable convergence guarantee~\cite{Sulam2017}. After every dictionary learning step we $\ell_{2}$-normalize each weight to avoid any redundant solution.

 \begin{algorithm}[!h]
\caption{Alternation of inference and learning} \label{AlgoAlternation}
\SetKwProg{Fn}{}{\string:}{}
\DontPrintSemicolon
\While{convergence not reached }  {
	\For{$i= 1$ \KwTo $L$}{
$\boldsymbol{\gamma}_{i}^{t+1} =  \mathcal{T}_{\eta_{c_{i}} \lambda_i}\big(\boldsymbol{\gamma}_{i}^{t} - \eta_{c_{i}} \nabla_ {\boldsymbol{\gamma}_{i}^{t} }\mathcal{L}\big)$ \textcolor{gray}{ \mQuad \mQuad \# inference }
	}
	}
\For{$i= 1$ \KwTo $L$}{
$\mathbf{D}_{i}^{t+1} =  \mathbf{D}_{i}^{t} - \eta_{i} \nabla_ {\boldsymbol{D}_{i}^{t} }\mathcal{L}$ \textcolor{gray}{\mQuad \mQuad \mQuad \mQuad \mQuad \mQuad\ \mQuad \mQuad \mQuad \# learning }
}
\end{algorithm}

 \subsubsection*{Back-projection mechanism}
 Interestingly, the dictionaries could be used to project (or back-project) the activity of a neural population and their associated synaptic weights into the next (or previous) level. Due to their high dimensionality, the weights $\mathbf{D}_{i}$ or difficult to interpret and visualize for $i>1$ as they represent a structure into an intermediate feature space at layer $i-1$. To overcome this limitation, we back-project the weights $\mathbf{D}_{i}$ into the input space, which is the visual space~\cite{Sulam2017}. This back-projection, called effective dictionary and denoted by $\mathbf{D}_{i}^\mathrm{eff}$, could be interpreted as the set of Receptive Fields (RFs) of the neurons located in layer i.  Mathematically, the effective dictionaries are described in Eq.~\ref{eq:Effective_Dictionaries}, and illustrated by Fig.~\ref{fig:fig_backprojection}.
\begin{equation}
 \label{eq:Effective_Dictionaries}
 \mathbf{D}^{\mathrm{eff}^{T}}_{i} = \mathbf{D}^{T}_{0}..\mathbf{D}^{T}_{i-1}\mathbf{D}^{T}_{i}
\end{equation}
Similarly, we defined $\boldsymbol{\gamma}_{i}^\mathrm{eff}$ as the back-projection into the visual space of the hidden states variable $\boldsymbol{\gamma}_{i}$ (Eq.~\ref{eq:Reconstruction}). This mechanism is used to reconstruct the input image from one intermediate layer.
\begin{equation}
 \label{eq:Reconstruction}
 \boldsymbol{\gamma}_{i}^\mathrm{eff} = \mathbf{D}^{\mathrm{eff}^{T}}_{i} \boldsymbol{\gamma}_{i}
\end{equation}

\subsection*{Databases}
We train our SDPC model on two different databases: The Chicago Face Database (CFD)~\cite{ma2015chicago} and STL-10~\cite{coates2011analysis}.

{\bf Chicago Face Database (CFD).} CFD consists of $1,804$ high-resolution ($2,444\times1,718$ px), color, standardized photographs of Black and White males and females between the ages of $18$ and $40$ years. We re-sized the pictures to $170\times 120$ px to keep reasonable computational time. The CFD database is partitioned into batches of 10 images. This dataset is split into a training set composed of $721$ images and a testing set of $400$ images.

{\bf STL-10 database.} The STL-10 dataset is an image recognition dataset developed for unsupervised feature learning and composed of color photographs with a resolution of $96\times96$ px representing animals (bird, cat, deer, dog, horse, monkey) and non-animals (airplane, car, ship, truck). The images are highly diverse (different view-point, background...) and could be considered as natural images. The set is partitioned into a training set of $5000$ images and a testing test of $1200$ images.

All the curves, images and histograms presented in this paper are generated using the testing set. The training set is used only to learn the synaptic weights. All these databases are pre-processed using Local Contrast Normalization (LCN) and whitening. LCN is inspired by neuroscience and consists in a local subtractive and divisive normalization~\cite{jarrett2009best}. In addition, we use whitening to reduce dependency between pixels.

\subsection*{Network parameters}
Networks and training parameters of the SDPC model are summarized in the table Table.~\ref{table:Parameters} for CFD and STL-10 databases.
\begin{table}[h!]
  \caption{{\bf SPDC Network and training parameters on CFD and STL-10 databases}. The size of the convolutional kernels for each layer are shown in the format: [number of features, number of channels, width, height] (value of the convolutional stride).}
  \label{table:Parameters}
  \centering
  \setlength\tabcolsep{3pt}
  \begin{tabular}{?C{18mm}|C{18mm}?C{40mm}?C{40mm}?}
    \cline{3-4}
    \multicolumn{2}{c?}{} & \multicolumn{2}{c?}{DataBase} \\
    \cline{3-4}
    \multicolumn{2}{c?}{}& CFD & STL-10 \\
    \Xhline{2\arrayrulewidth}
    \multirow{5}{1.2cm}{\centering network param.} & $\boldsymbol{D}_{1}$ size & [64, 1, 9, 9] (3) & [64, 3, 8, 8] (2) \\
    & $\boldsymbol{D}_{2}$ size & [128, 64, 9, 9] (1) & [128, 64, 8, 8] (1)\\
    \cline{2-4}
    & $\lambda_1$ & $0.3$ & 0.4\\
    & $\lambda_2$ & $1.6$ & 1.2 \\
    \cline{2-4}
    &  $T_{stab}$ & 5e-3 & 5e-3 \\
    \Xhline{2\arrayrulewidth}
    \multirow{4}{1.2cm}{\centering training param.} &\# epochs & 250 & 250 \\
    \cline{2-4}
    &$\eta_{L_{1}}$ & 1e-4 & 1e-4\\
    &$\eta_{L_{2}}$ & 5e-3 & 5e-3\\
    &momentum& 0.9 & 0.9  \\
    \Xhline{2\arrayrulewidth}
  \end{tabular}
\end{table}

We used PyTorch 1.0~\cite{paszke2017automatic} to implement, train, and test the SDPC model. The codes of the model and all the simulations that generate the result of this paper are available at 'www.github.com/XXX/XXX'.

\subsection*{Interaction maps analysis}\label{sec:InteractionMap}
Using the intrinsic invariance to translations of CNNs and the fact that learned filters are in majority edge-like filters, we define the notion of interactions map to reduce the neural activity to two state variables at every position on the V1 space: the resulting orientation and activity. Using interaction map allows us to represent in 2D the state of the network surrounding a given central position. We choose the location of the center of the interaction map such that neurons, at this position, are strongly responsive to a given orientation. This orientation is called the central preferred orientation and denoted $\theta_c$. Interaction maps are denoted $\boldsymbol{\bar{a}}$.

To mathematically define the resulting orientations and activities of the interaction map, it is convenient to interpret the V1 activity-map as a 3-dimensional tensor, in which the first dimension corresponds to one specific orientation and the two last dimensions describe the V1 space. Eq.~\ref{eq:eq2} formalizes this interpretation using complex number representation.
\begin{equation}
    \label{eq:eq2}
    \boldsymbol{\gamma}_{1} \equiv \{\boldsymbol{\gamma}_{1}[\, \theta,x,y ]\, e^{j\theta}\}_{\forall \theta,\forall x,\forall y}
\end{equation}
In Eq.~\ref{eq:eq2}, $\theta$ denotes the orientation of the filters. In practice, we estimate $\theta$ by fitting the first layer RFs with Gabor filters~\cite{fischer2007self}. Note that textural and low-frequency filters are poorly fitted and are removed (we remove $13$ out of $64$ filters). %
For each orientation, we adjust the V1 activity-map surrounding the central preferred orientation with a marginal activity. The marginal activity is computed by extracting the mean neighborhood in a spatially shuffled version of the V1 activity-map. Said differently, the marginal activity, denoted $\boldsymbol{\gamma}_{1}[\, \theta,x_{\sim c},y_{\sim c}]\,$, is a spatial average over the activity of neurons that respond to one given orientation $\theta$. The adjusted activity is called $\boldsymbol{a}$ and its computation is defined in Eq.~\ref{eq:eq3}.
\begin{equation}
    \label{eq:eq3}
    \boldsymbol{a}[\, \theta,x_{c},y_{c} ] = \frac{\boldsymbol{\gamma}_{1}[\, \theta,x_{c},y_{c} ]\, - \boldsymbol{\gamma}_{1}[\, \theta,x_{\sim c},y_{\sim c} ]\,}{\boldsymbol{\gamma}_{1}[\, \theta,x_{\sim c},y_{\sim c} ]\,}
   \end{equation}
In Eq.~\ref{eq:eq3}, ($x_{c}$, $y_{c}$) denotes the coordinates  of the V1-space neighboring the central preferred orientation, and ($x_{\sim c}$, $y_{\sim c}$) represents the V1-space outside this neighborhood. At each position of the V1-space surrounding the central preferred orientation, the interaction map is computed as the weighted average over all the orientations of the adjusted activity vector (see Eq.~\ref{eq:eq5}). We denote  $\boldsymbol{\bar{\theta}}$ and $\big \vert \boldsymbol{\bar{a}} \big \vert$ the resulting orientation and activity of the interaction map, respectively (see Eq.~\ref{eq:eq4}).

\begin{align}
    \boldsymbol{\bar{a}}[x_{c},y_{c} ]\,& = \frac{1}{n}\sum_{\theta=\theta_{1}}^{\theta_{n}} \boldsymbol{a}[ \theta, x_{c},y_{c} ] \cdot e^{j\theta} \label{eq:eq5}\\
    & = \big \vert \boldsymbol{\bar{a}}[x_{c},y_{c} ] \big \vert \cdot e^{j\bar{\theta}[x_{c},y_{c}]}  \label{eq:eq4}
   \end{align}

We use a circular weighted average to compute the resulting orientation (see Eq.~\ref{eq:eq6}) and activity (see Eq.~\ref{eq:eq7}) of the interaction map.

\begin{align}
\boldsymbol{\bar{\theta}}[x_{c},y_{c}] & = atan2\Big ( \frac{1}{n}\sum_{\theta=\theta_{1}}^{\theta_{n}} \boldsymbol{a}[ \theta, x_{c},y_{c} ] sin(\theta),\frac{1}{n}\sum_{\theta=\theta_{1}}^{\theta_{n}} \boldsymbol{a}[ \theta, x_{c},y_{c} ] cos(\theta)\Big ) \label{eq:eq6} \\
\big \vert \boldsymbol{\bar{a}}[x_{c},y_{c} ] \big \vert &= \frac{1}{n}\sqrt{\bigg(\sum_{\theta=\theta_{1}}^{\theta_{n}} \boldsymbol{a}[ \theta, x_{c},y_{c} ] cos(\theta)\bigg)^{2} + \bigg(\sum_{\theta=\theta_{1}}^{\theta_{n}} \boldsymbol{a}[ \theta, x_{c},y_{c} ] sin(\theta)\bigg)^{2}} \label{eq:eq7}
  \end{align}

For each image, we average $10$ interaction maps with the same central preferred orientation. The centers of these maps correspond to the $10$ locations, in the V1-space, showing the strongest response to the given central preferred orientation. The size of the interaction map is set to cover the second layer neurons' RFs. In practice, our interaction maps have a $9\times9$ size in the V1 space. The interaction maps obtained for each image are finally averaged over $1200$ images of the STL-10 testing set. We iterate this analysis for different feedback strengths ranging from $0$ to $4$.

We measure the co-linearity deviation of the interaction map with a circular difference between the central preferred orientation ($\theta_c$) and the orientation of the interaction map (see Eq.~\ref{eq:eq8}). The co-circularity deviation is quantified using a circular difference between a map of orientations that are co-circular to the central preferred orientation and the angle of the interaction map (see Eq.~\ref{eq:eq9})~\cite{perrinet2015edge}. We simplify the calculation of the co-circular angle map in Eq.~\ref{eq:eq9} by centering the coordinate $(x_{c}, y_{c})$ in the middle of the interaction map (the co-circular map is shown in the top right corner of the Fig.~\ref{fig:fig5}-A and Fig.~\ref{fig:fig5}-B).

\begin{align}
    \boldsymbol{\theta}_{co-lin}[x_{c},y_{c} ] & = \big \vert \theta_{c} - \boldsymbol{\bar{\theta}}[x_{c},y_{c}] \big \vert\label{eq:eq8} \\
   \boldsymbol{\theta}_{co-cir}[x_{c},y_{c} ]  & = \big \vert \atan\Big(\frac{y_c - y_{co}}{x_c - x_{co}}\Big) + \frac{\pi}{2} - \boldsymbol{\bar{\theta}}[x_{c},y_{c}]) \big \vert \label{eq:eq9} \\
      \textnormal{with } x_{co} &= \frac{\sin(\theta_c)\cdot(x_{c}^2+y_{c}^2)}{2\big(\sin(\theta_c)\cdot x_{c} - \cos(\theta_c)\cdot y_{c}\big)} \nonumber \\
      \textnormal{and } y_{co} &= \tan(\theta_{c} + \frac{\pi}{2}) \cdot x_{co} \nonumber
   \end{align}

For a given feedback strength $k_\mathrm{FB}$, we synthesize our results by introducing two ratios to compare the respective precisions in co-linearity ($r_{co-lin}^{k_\mathrm{FB}}$ in Eq.~\ref{eq:eq10})  and co-circularity ($r_{co-cir}^{k_\mathrm{FB}}$ in Eq.~\ref{eq:eq11}). In those equations, the marginal co-linearity and co-circularity are denoted $\boldsymbol{\theta}_{co-lin}^{k_\mathrm{FB}}[x_{\sim c},y_{\sim c}]$ and $\boldsymbol{\theta}_{co-cir}^{k_\mathrm{FB}}[x_{\sim c},y_{\sim c}]$, respectively. To facilitate the interpretation of these ratios, we make sure they are following the same evolution than a precision measure. For example, if an interaction map exhibits a higher co-linearity with the central preferred orientation, then the corresponding $r_{co-lin}^{k_\mathrm{FB}}$ will be necessarily over $1$.

\begin{align}
\boldsymbol{r}_{co-lin}^{k_\mathrm{FB}} =  \frac{\boldsymbol{\theta}_{co-lin}^{k_\mathrm{FB}}[x_{\sim c},y_{\sim c}]}{\boldsymbol{\theta}_{co-lin}^{k_\mathrm{FB}}[x_c,y_c]} \label{eq:eq10} \\
 \boldsymbol{r}_{co-cir}^{k_\mathrm{FB}} =  \frac{\boldsymbol{\theta}_{co-cir}^{k_\mathrm{FB}}[x_{\sim c},y_{\sim c}]}{\boldsymbol{\theta}_{co-cir}^{k_\mathrm{FB}}[x_c,y_c]}  \label{eq:eq11}
   \end{align}
   
To compare this relative activity with or without feedback, we introduce the ratio $\boldsymbol{r_{a}}(k_\mathrm{FB})$ (see Eq.~\ref{eq:eq12}).
\begin{align}
    \boldsymbol{r_{a}}(k_\mathrm{FB}) & =  \frac{\big \vert \boldsymbol{\bar{a}}(k_\mathrm{FB}) \big \vert}{\big \vert \boldsymbol{\bar{a}}(k_\mathrm{FB}=0) \big \vert} \label{eq:eq12}
   \end{align}


\nolinenumbers
\newpage

\bibliography{biblio}






\appendix
\setcounter{figure}{0}
\renewcommand{\thefigure}{S\arabic{figure}}

\makeatletter
\renewcommand{\fnum@algocf}{Algo S\AlCapFnt\thealgocf}
\makeatother

\newpage

\begin{figure}[h]
 \centering
\begin{tikzpicture}
 \draw [anchor=north west] (0.0\linewidth, 0.99\linewidth) node {\includegraphics[width=1\linewidth]{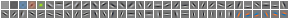}};
\draw [anchor=north west] (0.\linewidth, 0.87\linewidth) node {\includegraphics[width=1\linewidth]{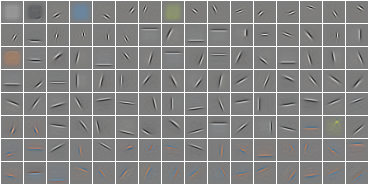} };

\begin{scope}
\draw [anchor=west,fill=white] (0.0\linewidth, 1\linewidth) node {{\bf A}};
\draw [anchor=west,fill=white] (0.0\linewidth, 0.88\linewidth) node {{\bf B}};
\end{scope}
\end{tikzpicture}
\caption{
{\bf RFs when the SDPC is trained on the STL-10 database.} {\bf (A)} $64$ first layer RFs, sorted by activation probability in a descending order. The size of the RFs is 9$\times$9 px. {\bf (B)} $128$ second layer RFs, sorted by activation probability in a descending order. The size of the RFs is 22$\times$22 px. All the visualized RFs are generated using Eq.~\ref{eq:Effective_Dictionaries}.}
\label{fig:figSD1}
\end{figure}

\begin{figure}[h]
\centering
\begin{tikzpicture}
 \draw [anchor=north west] (0.0\linewidth, 0.99\linewidth) node {\includegraphics[width=1\linewidth]{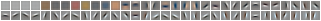}};
\draw [anchor=north west] (0.\linewidth, 0.87\linewidth) node {\includegraphics[width=1\linewidth]{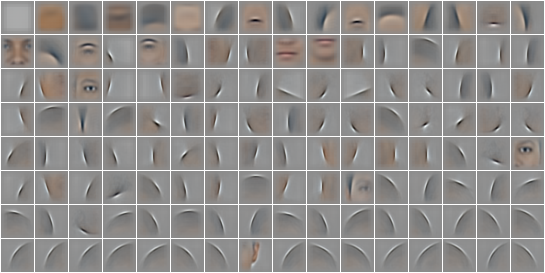} };
 \begin{scope}
\draw [anchor=west,fill=white] (0.0\linewidth, 1\linewidth) node {{\bf A}};
\draw [anchor=west,fill=white] (0.0\linewidth, 0.88\linewidth) node {{\bf B}};
\end{scope}
\end{tikzpicture}
\caption{
{\bf RFs when the SDPC is trained on the CFD database.} {\bf (A)} $64$ first layer RFs, sorted by activation probability in a descending order. The size of the RFs is 9$\times$9 px. {\bf (B)} $128$ second layer RFs, sorted by activation probability in a descending order. The size of the RFs is 33$\times$33 px. All the visualized RFs are generated using Eq.~\ref{eq:Effective_Dictionaries}.}
\label{fig:figSD2}
\end{figure}

\begin{figure}[h]
\begin{tikzpicture}
 \centering
 \draw [anchor=north west] (0.0\linewidth, 0.99\linewidth) node {\includegraphics[width=0.30\linewidth]{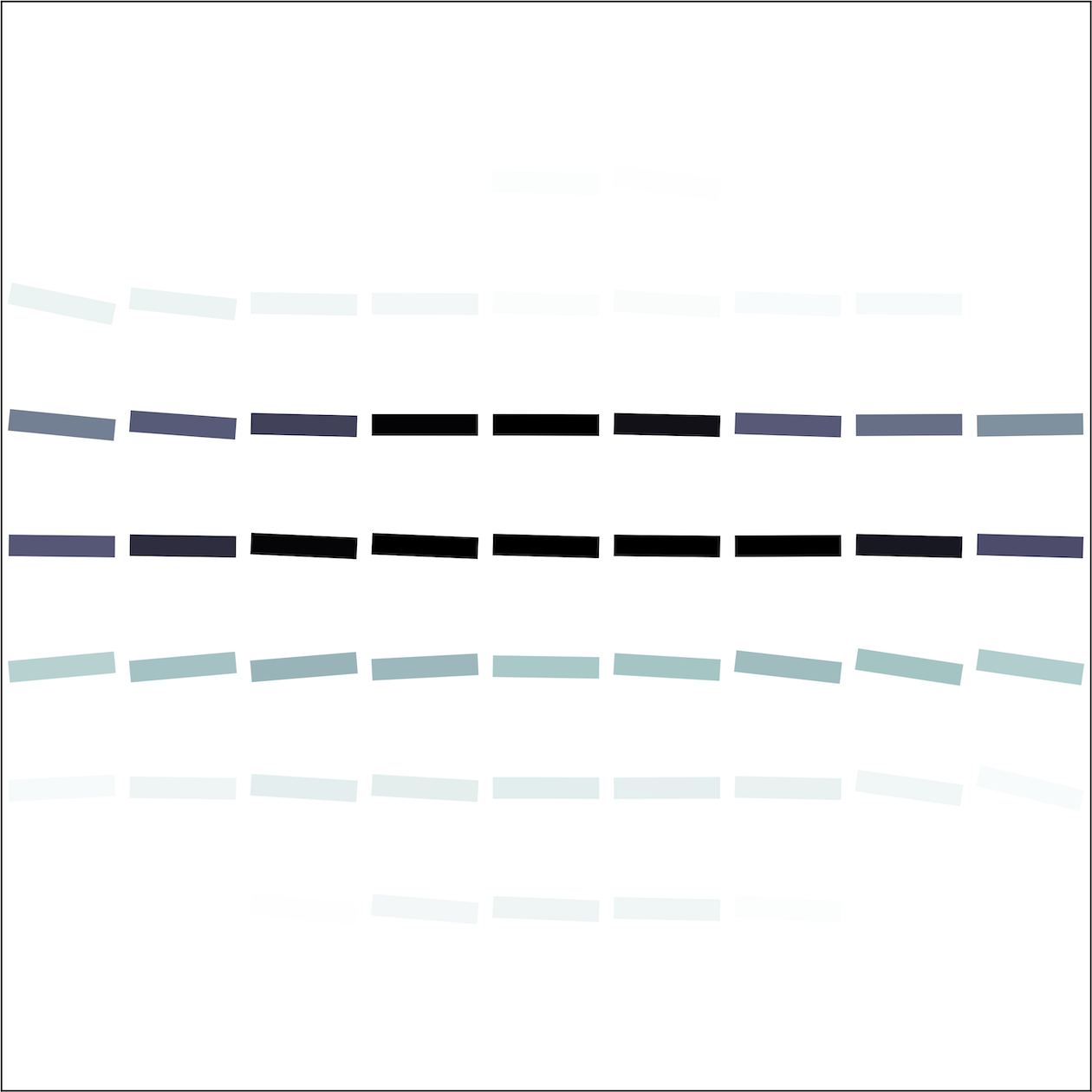}};
\draw [anchor=north west] (0.245\linewidth, 0.985\linewidth) node {\includegraphics[width=0.05\linewidth]{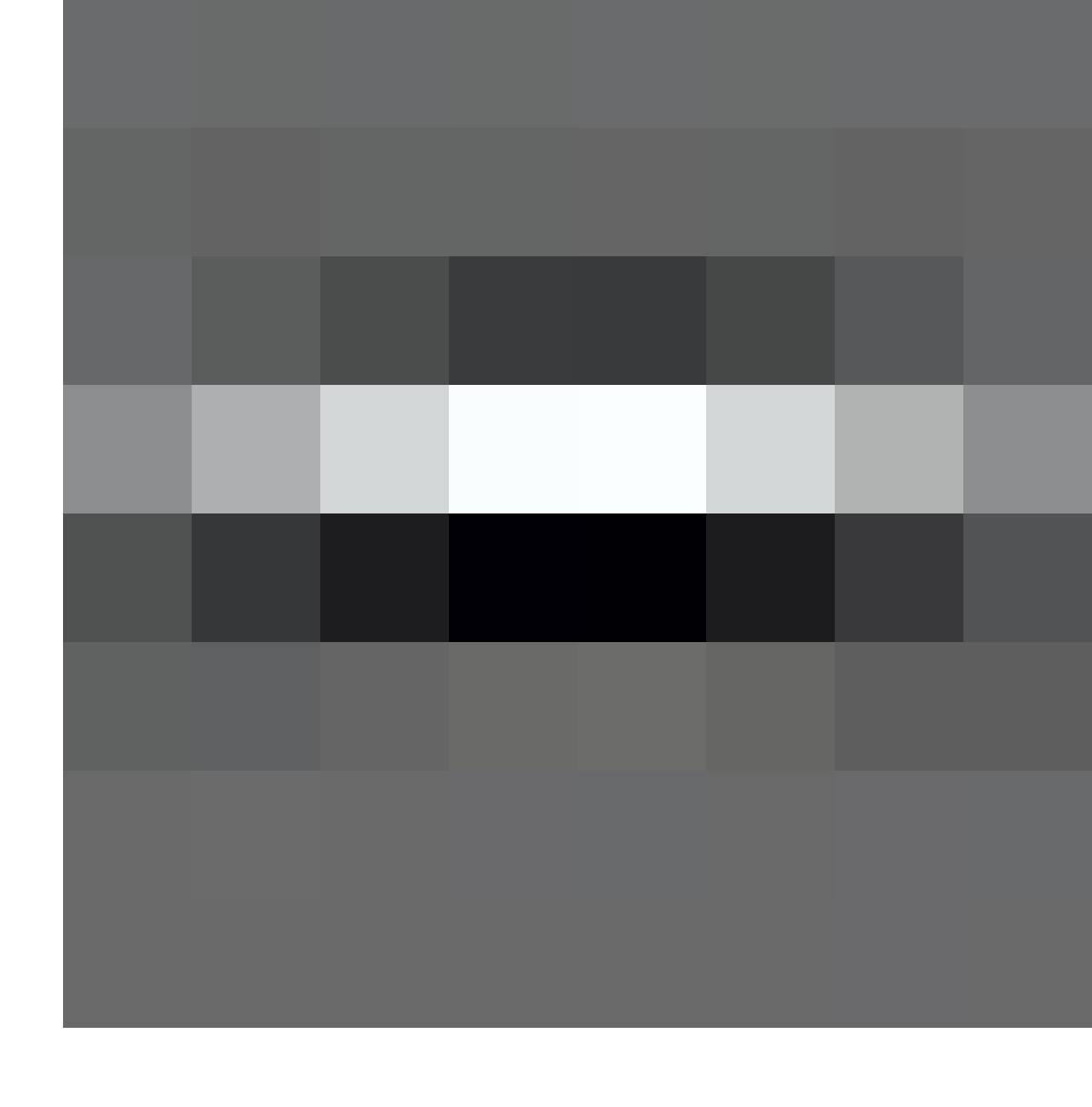} };

 \draw [anchor=north west] (0.33\linewidth, 0.99\linewidth) node {\includegraphics[width=0.30\linewidth]{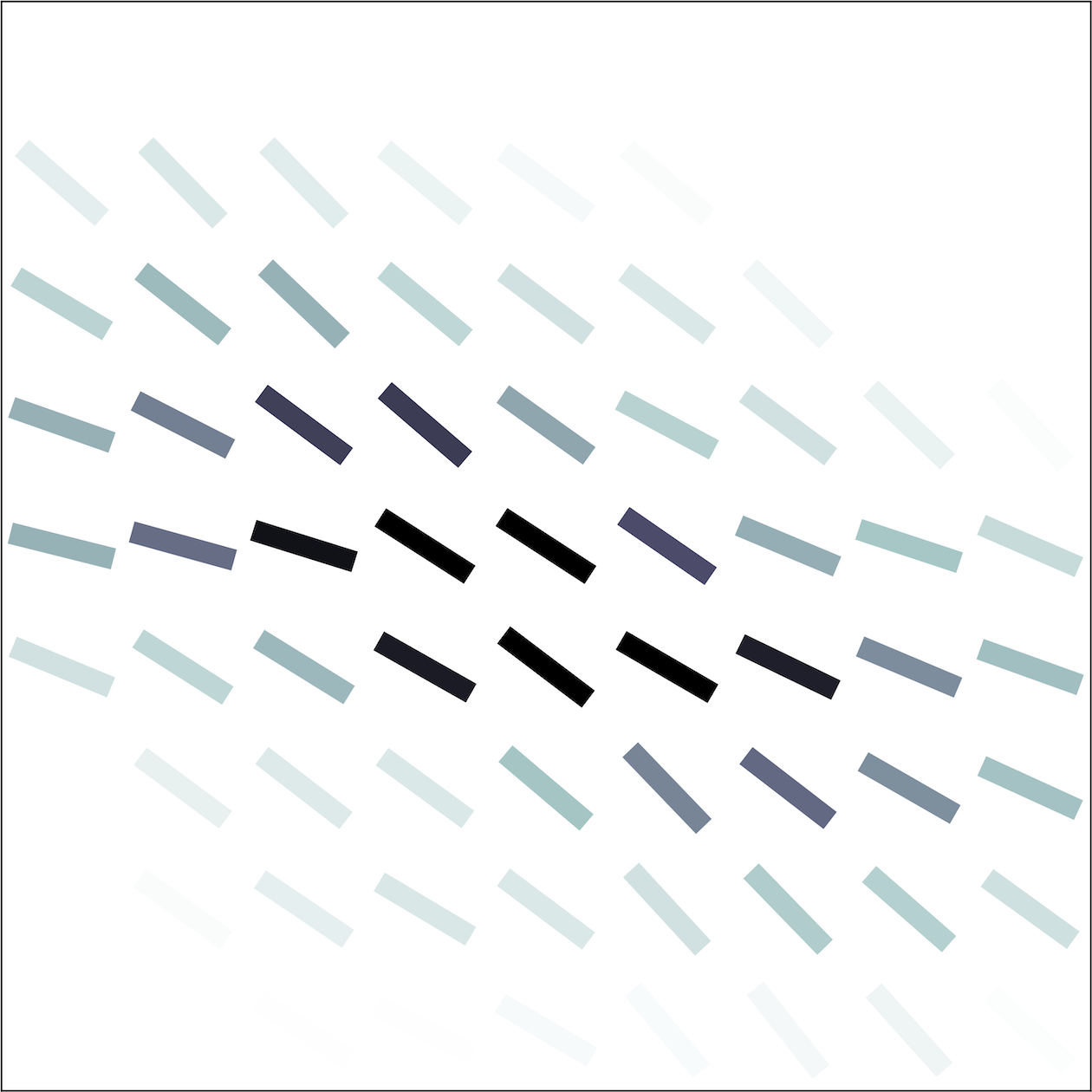}};
\draw [anchor=north west] (0.575\linewidth, 0.985\linewidth) node {\includegraphics[width=0.05\linewidth]{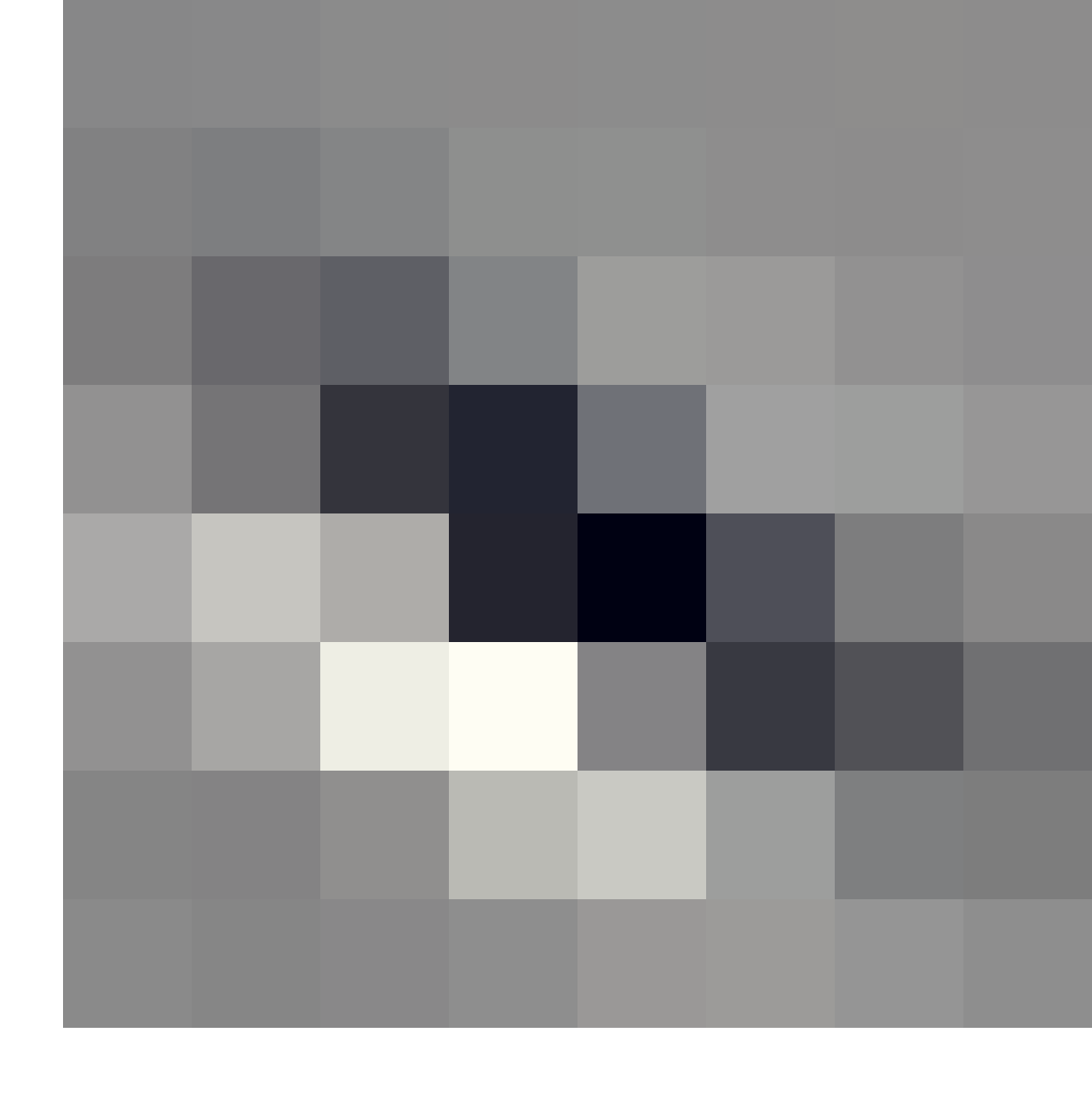} };

\draw [anchor=north west] (0.66\linewidth, 0.99\linewidth) node {\includegraphics[width=0.30\linewidth]{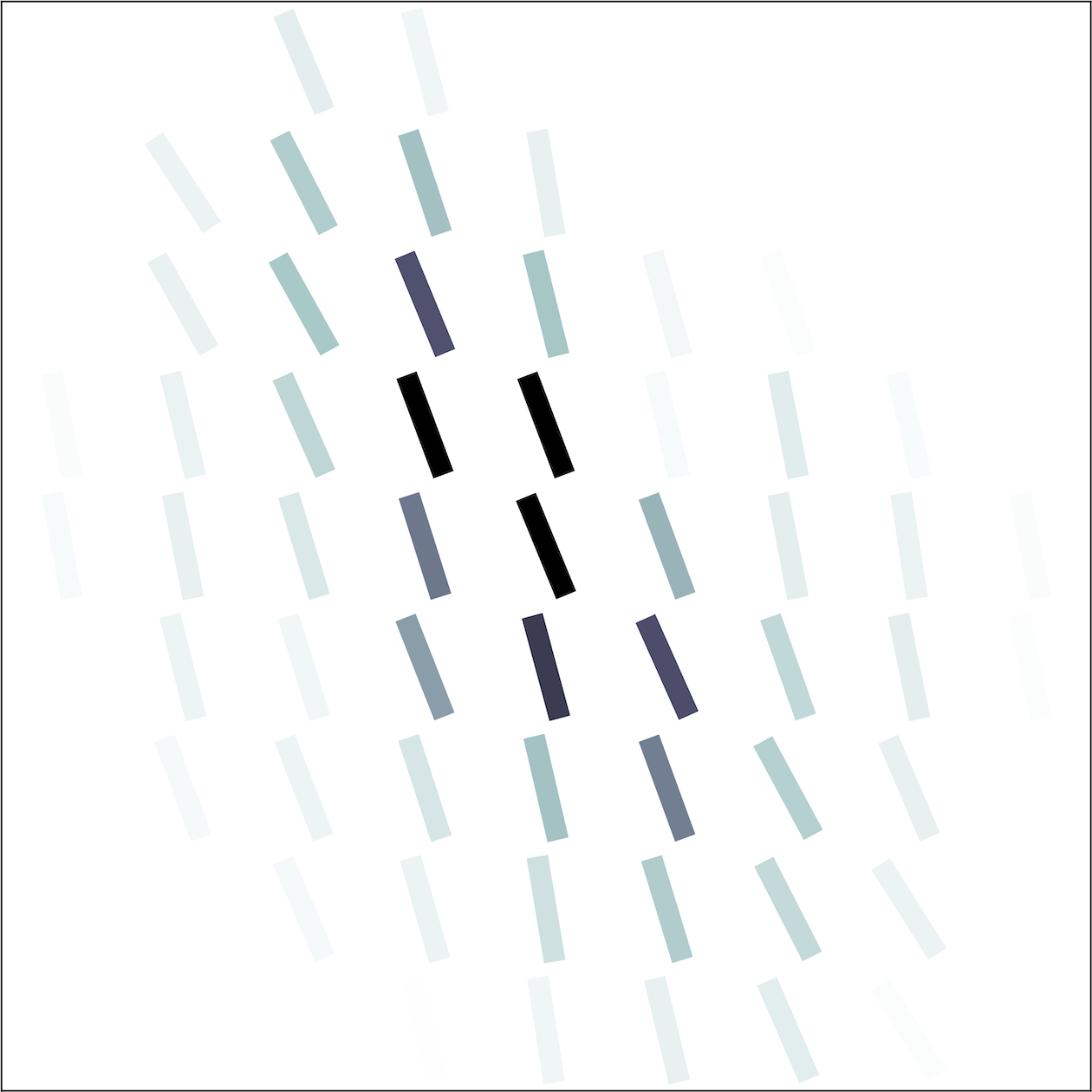}};
\draw [anchor=north west] (0.905\linewidth, 0.985\linewidth) node {\includegraphics[width=0.05\linewidth]{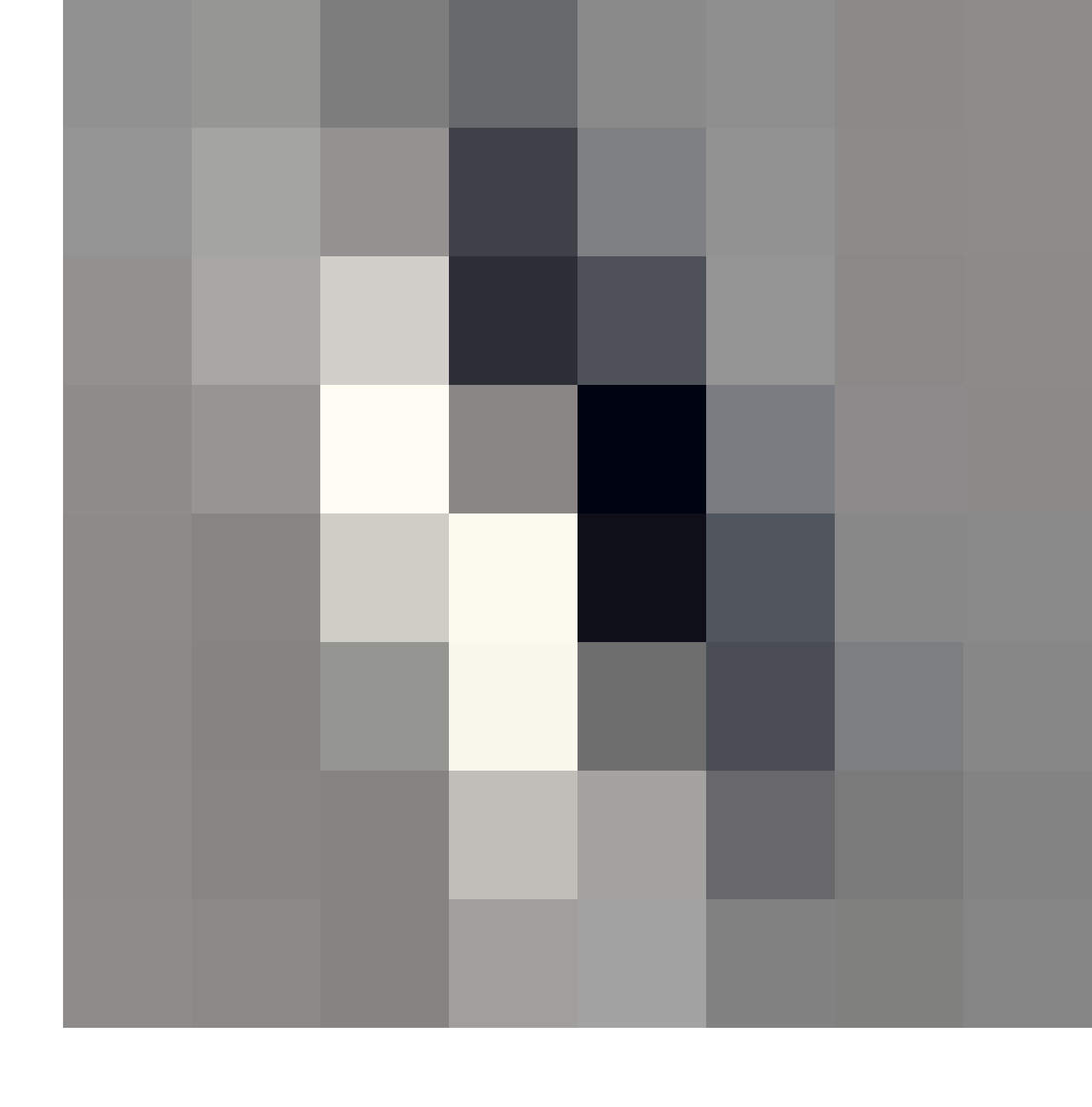} };

\draw [anchor=north west] (0.\linewidth, 0.64\linewidth) node {\includegraphics[width=0.30\linewidth]{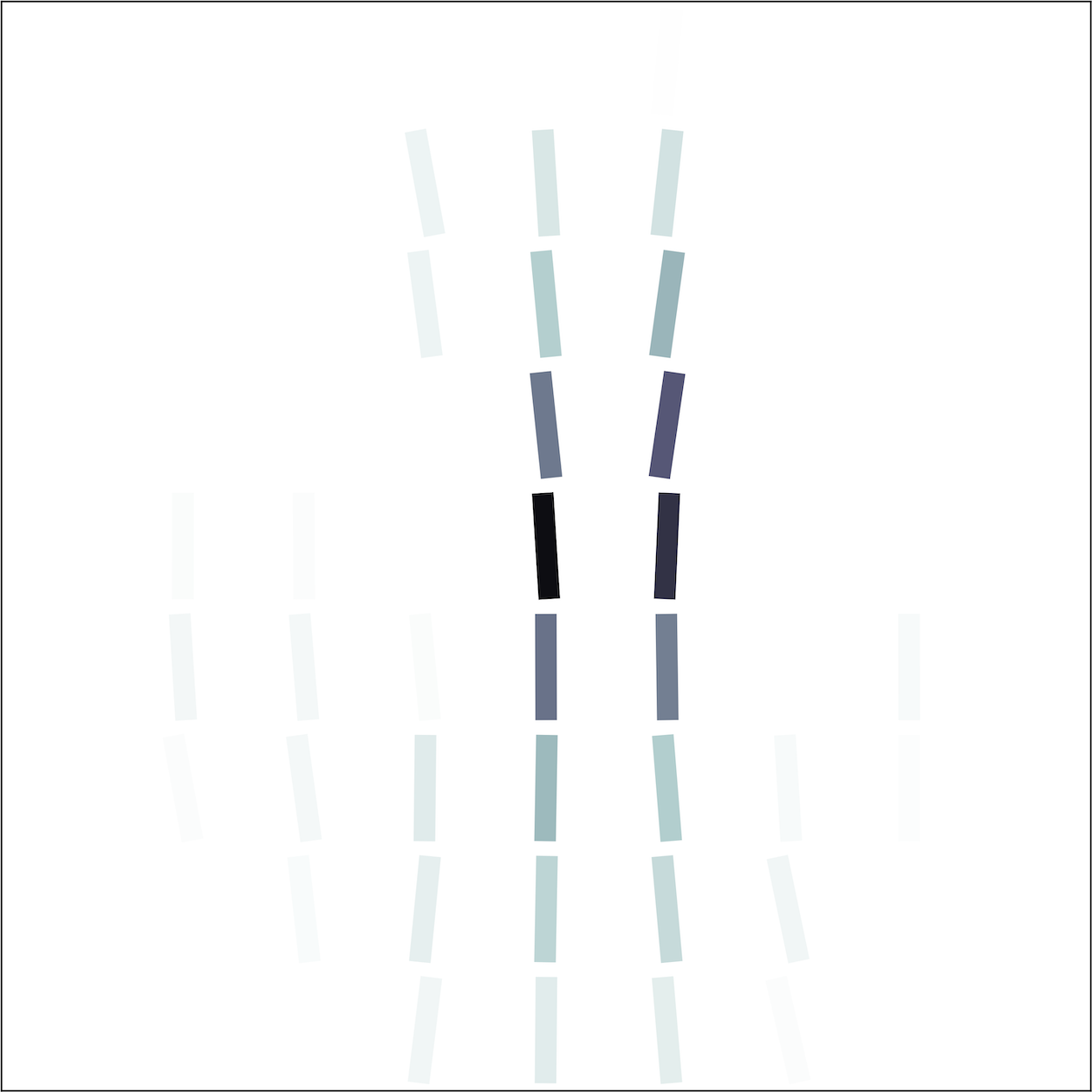}};
\draw [anchor=north west] (0.245\linewidth, 0.635\linewidth) node {\includegraphics[width=0.05\linewidth]{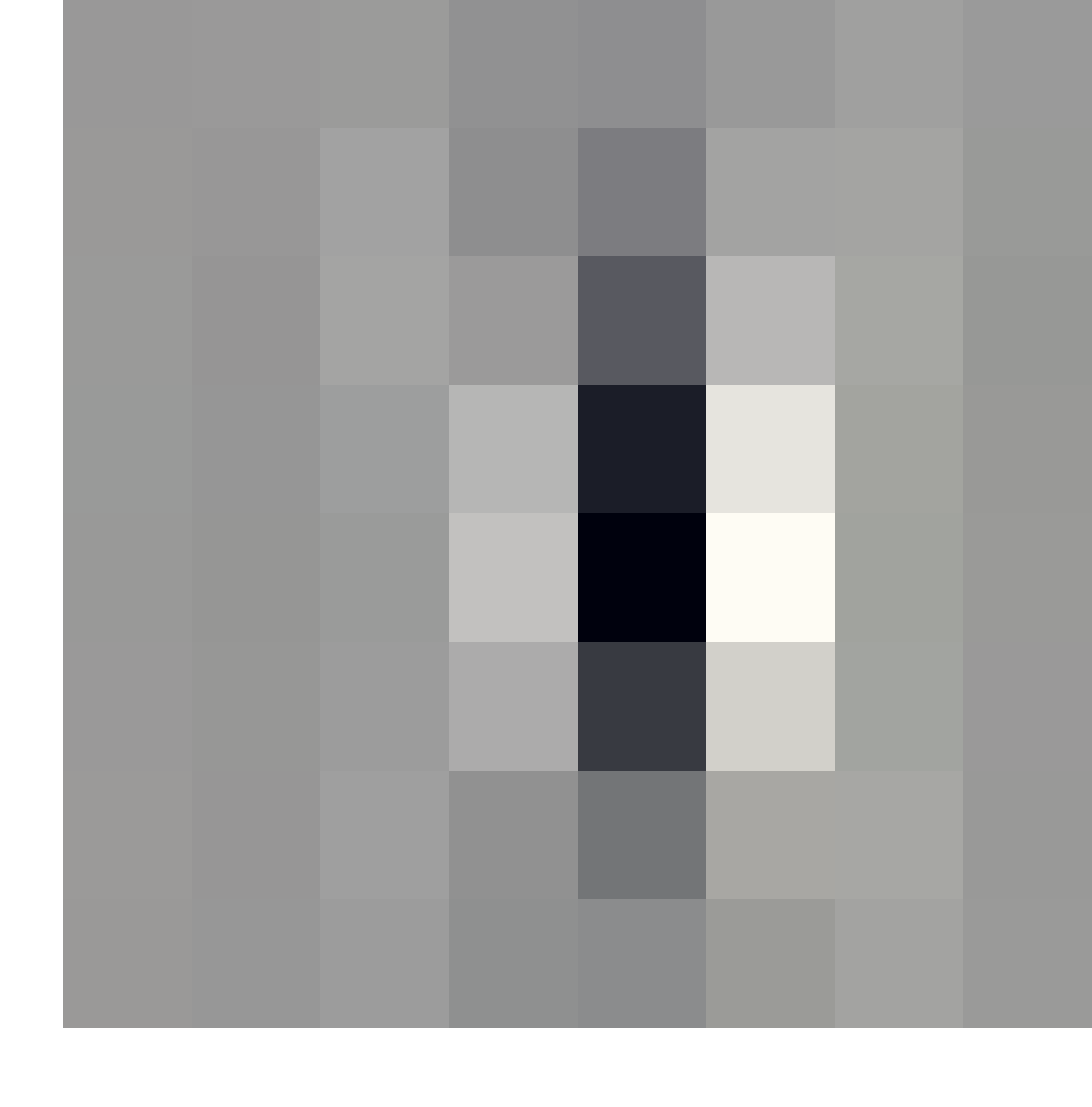} };

\draw [anchor=north west] (0.33\linewidth, 0.64\linewidth) node {\includegraphics[width=0.30\linewidth]{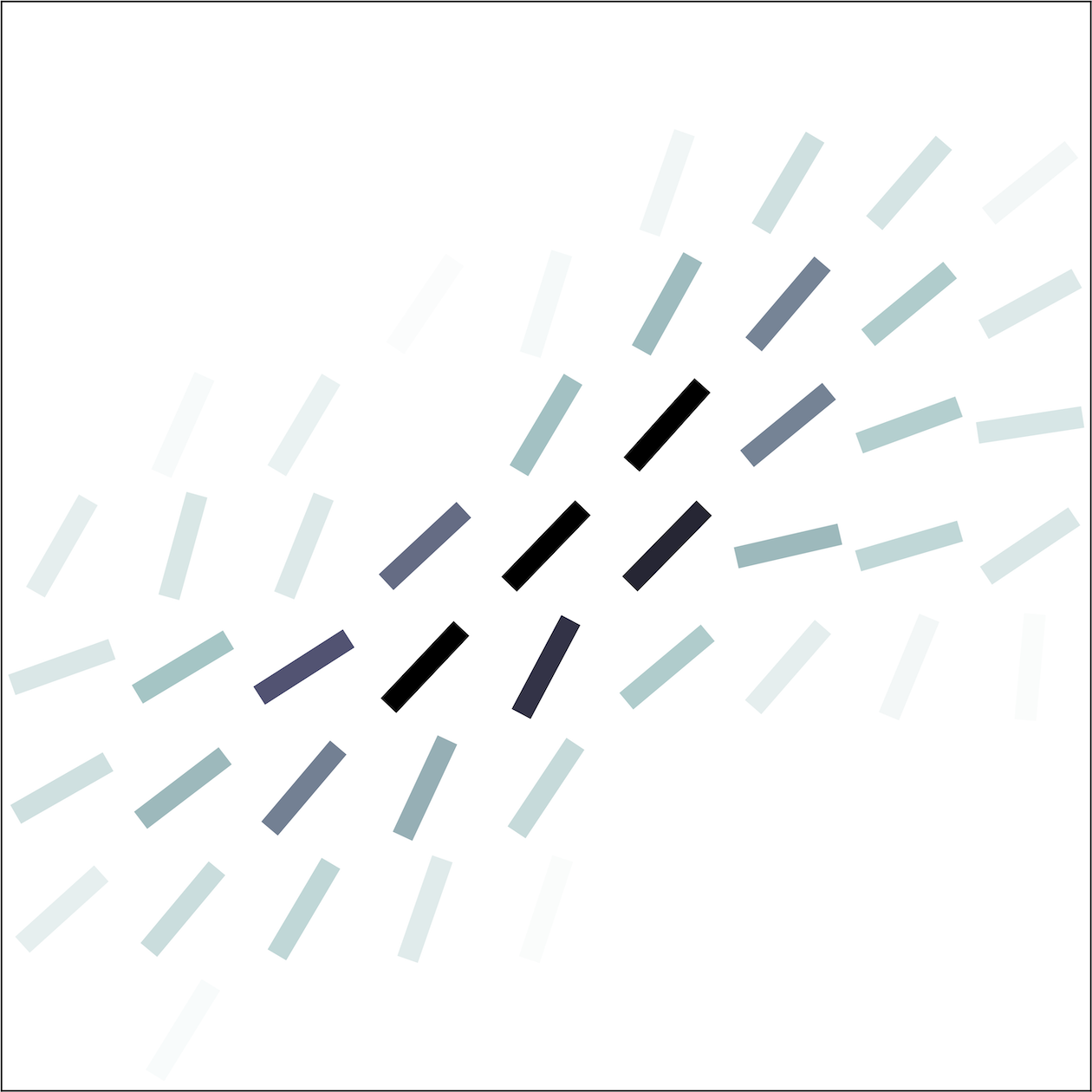}};
\draw [anchor=north west] (0.575\linewidth, 0.635\linewidth) node {\includegraphics[width=0.05\linewidth]{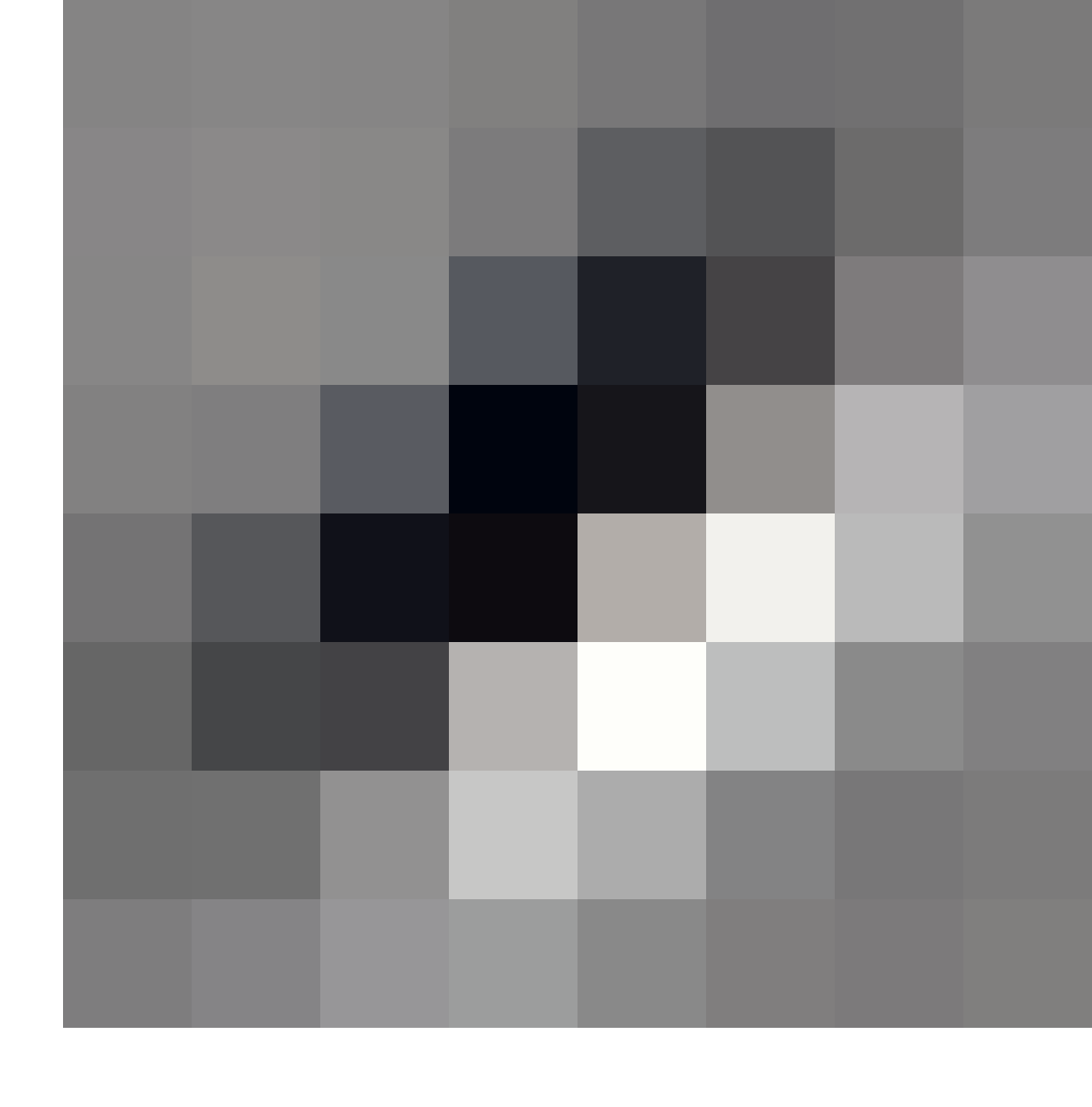} };

\draw [anchor=north west] (0.66\linewidth, 0.64\linewidth) node {\includegraphics[width=0.30\linewidth]{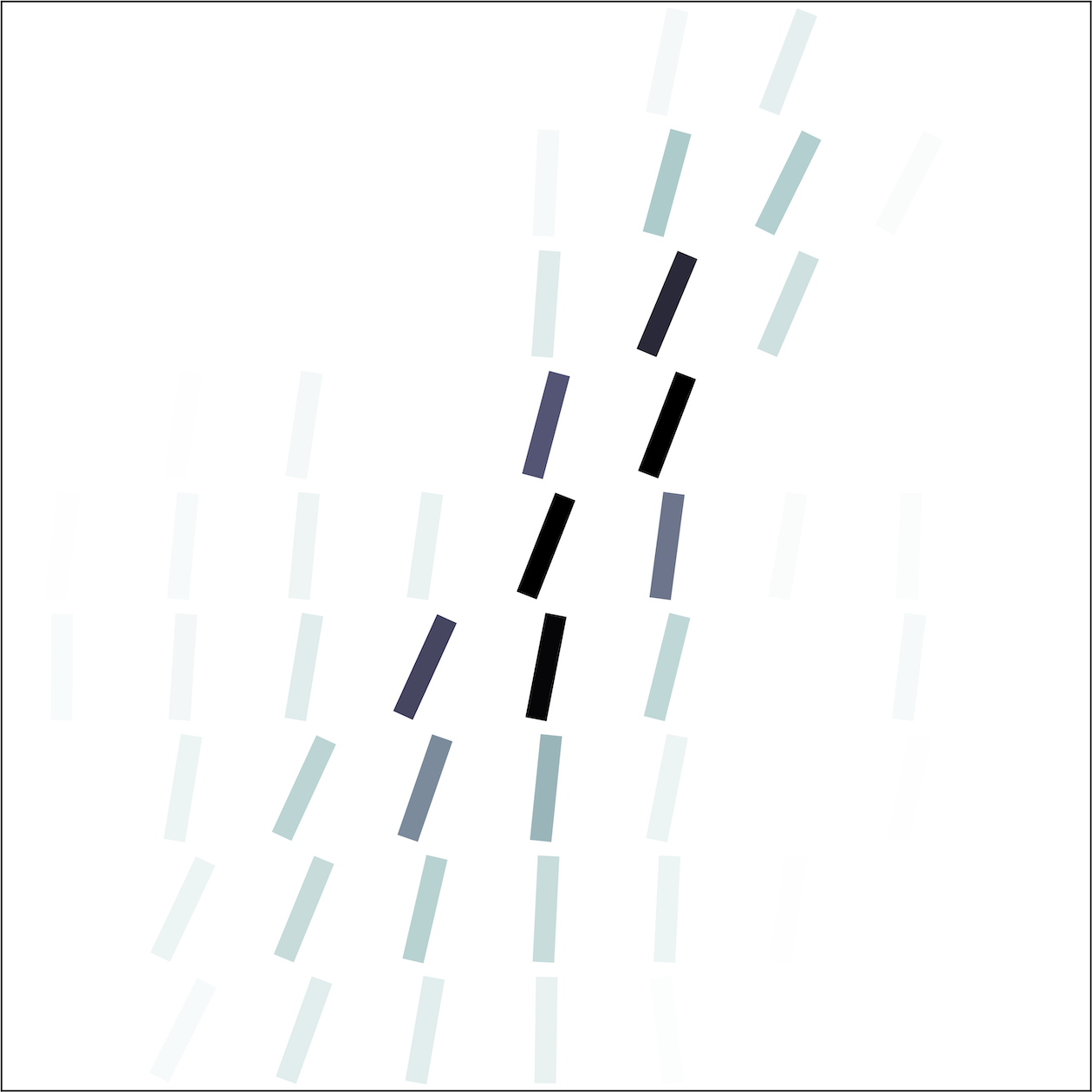}};
\draw [anchor=north west] (0.905\linewidth, 0.635\linewidth) node {\includegraphics[width=0.05\linewidth]{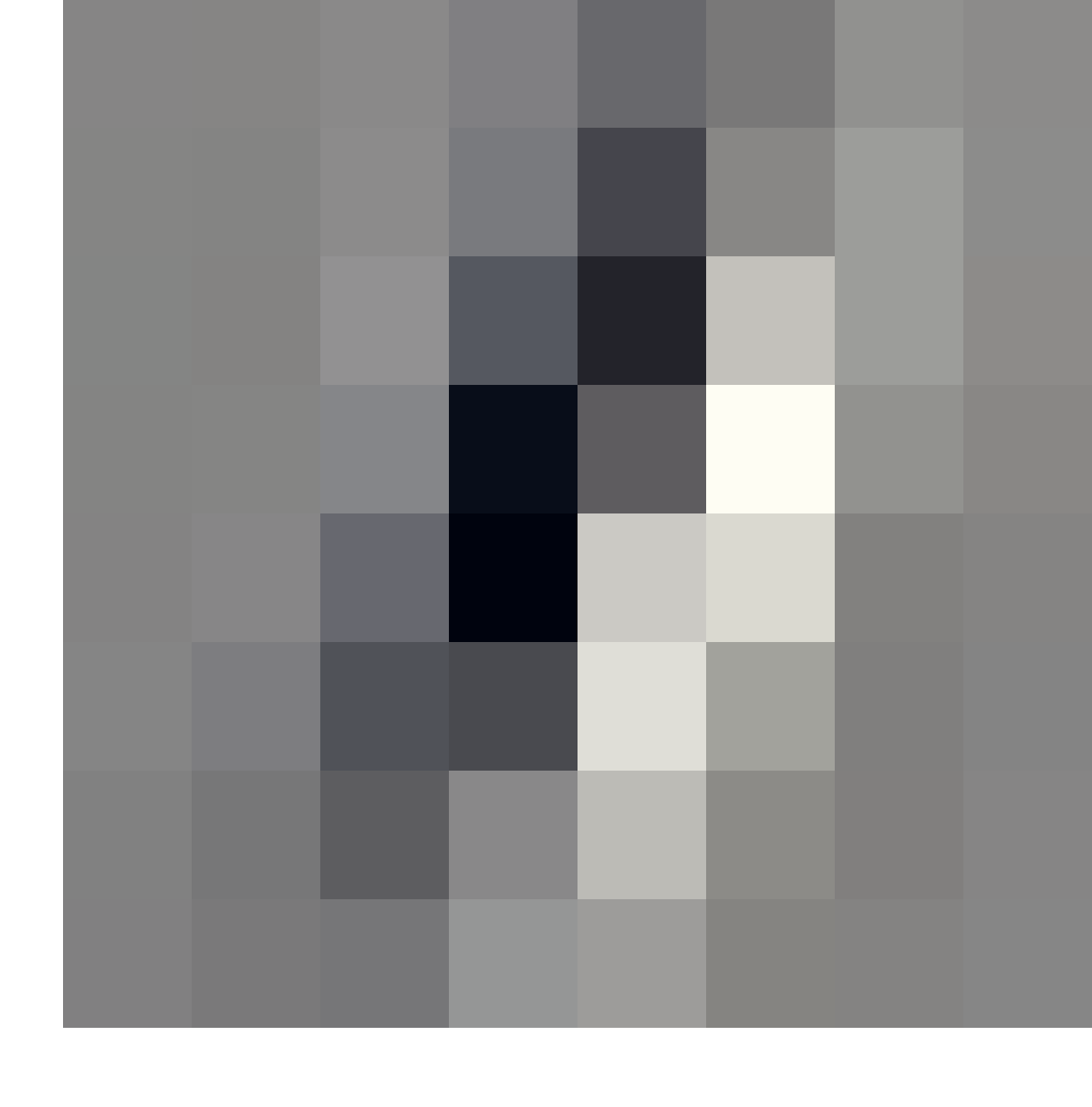} };

\draw [anchor=north west] (0.15\linewidth, 0.30\linewidth) node  {\includegraphics[width=0.7\linewidth]{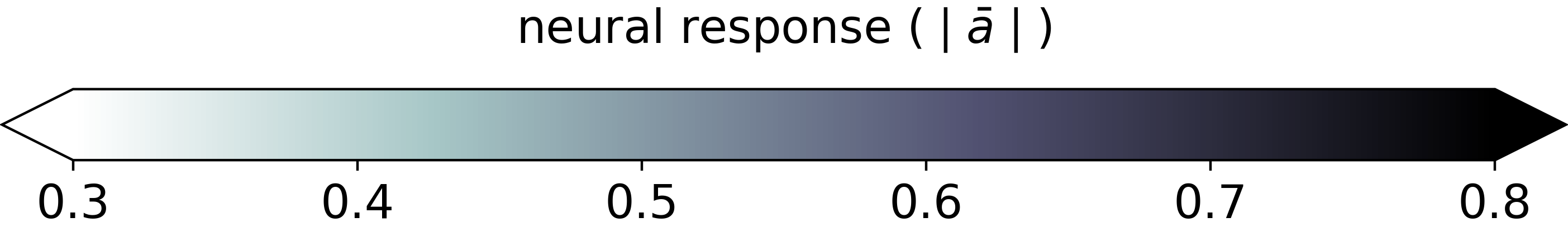}};
\begin{scope}
\draw [anchor=west,fill=white] (0.0\linewidth, 1\linewidth) node {{\bf A}};
\draw [anchor=west,fill=white] (0.33\linewidth, 1\linewidth) node {{\bf B}};
\draw [anchor=west,fill=white] (0.66\linewidth, 1\linewidth) node {{\bf C}};
\draw [anchor=west,fill=white] (0\linewidth, 0.65\linewidth) node {{\bf D}};
\draw [anchor=west,fill=white] (0.33\linewidth, 0.65\linewidth) node {{\bf E}};
\draw [anchor=west,fill=white] (0.66\linewidth, 0.65\linewidth) node {{\bf F}};

\end{scope}
\end{tikzpicture}
\caption
{{\bf Example of 9$\times$9 association field of V1 centered on neurons strongly responding to 6 different contour orientation, when the SDPC is trained on STL-10 database}. From left to right and top to bottom the contour orientations are \ang{0} {\bf (A)}, \ang{-30} {\bf (B)}, \ang{-60} {\bf (C)}, \ang{90} {\bf (D)}, \ang{60} {\bf (E)} and \ang{30} {\bf (F)}. The feedback strength is set to $1$. At each location identified by the coordinates $(x_{c},y_{c})$ the angle is $\boldsymbol{\bar{\theta}}[x_{c},y_{c}]$ (see Eq.~\ref{eq:eq6}) and the color scale is $\big \vert \boldsymbol{\bar{a}}[x_{c},y_{c} ] \big \vert$ (see Eq.~\ref{eq:eq7}). The color scale being saturated toward both maximum and minimum activity, all the activities above $0.8$ or below $0.3$ have the same dark green or white color, respectively.}
\label{fig:figSD3}
\end{figure}

\begin{figure}[h]
\begin{tikzpicture}
 \centering
 \draw [anchor=north west] (0.0\linewidth, 0.99\linewidth) node {\includegraphics[width=0.30\linewidth]{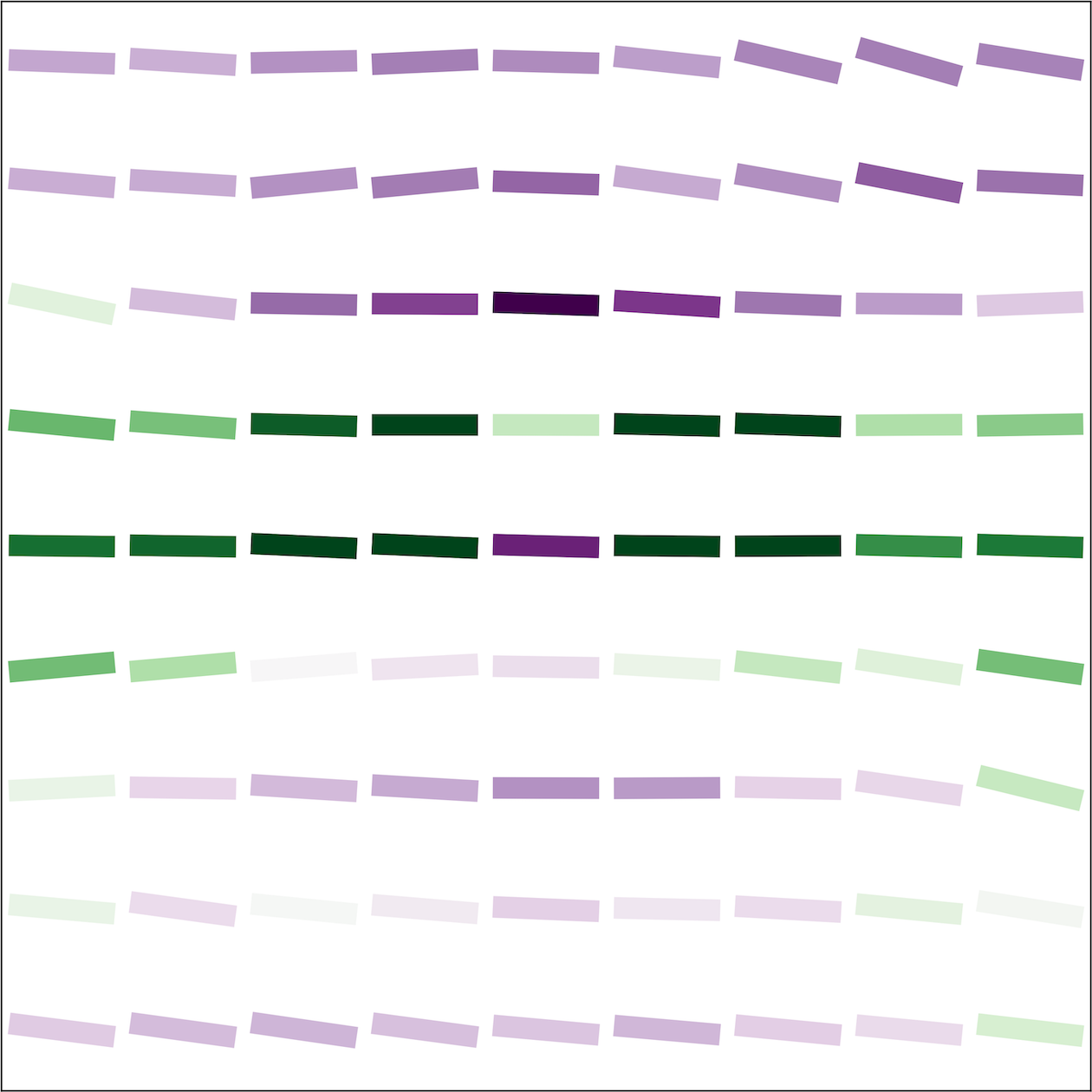}};
\draw [anchor=north west] (0.245\linewidth, 0.985\linewidth) node {\includegraphics[width=0.05\linewidth]{Figure/FigSD3/Filter=-0.png} };

 \draw [anchor=north west] (0.33\linewidth, 0.99\linewidth) node {\includegraphics[width=0.30\linewidth]{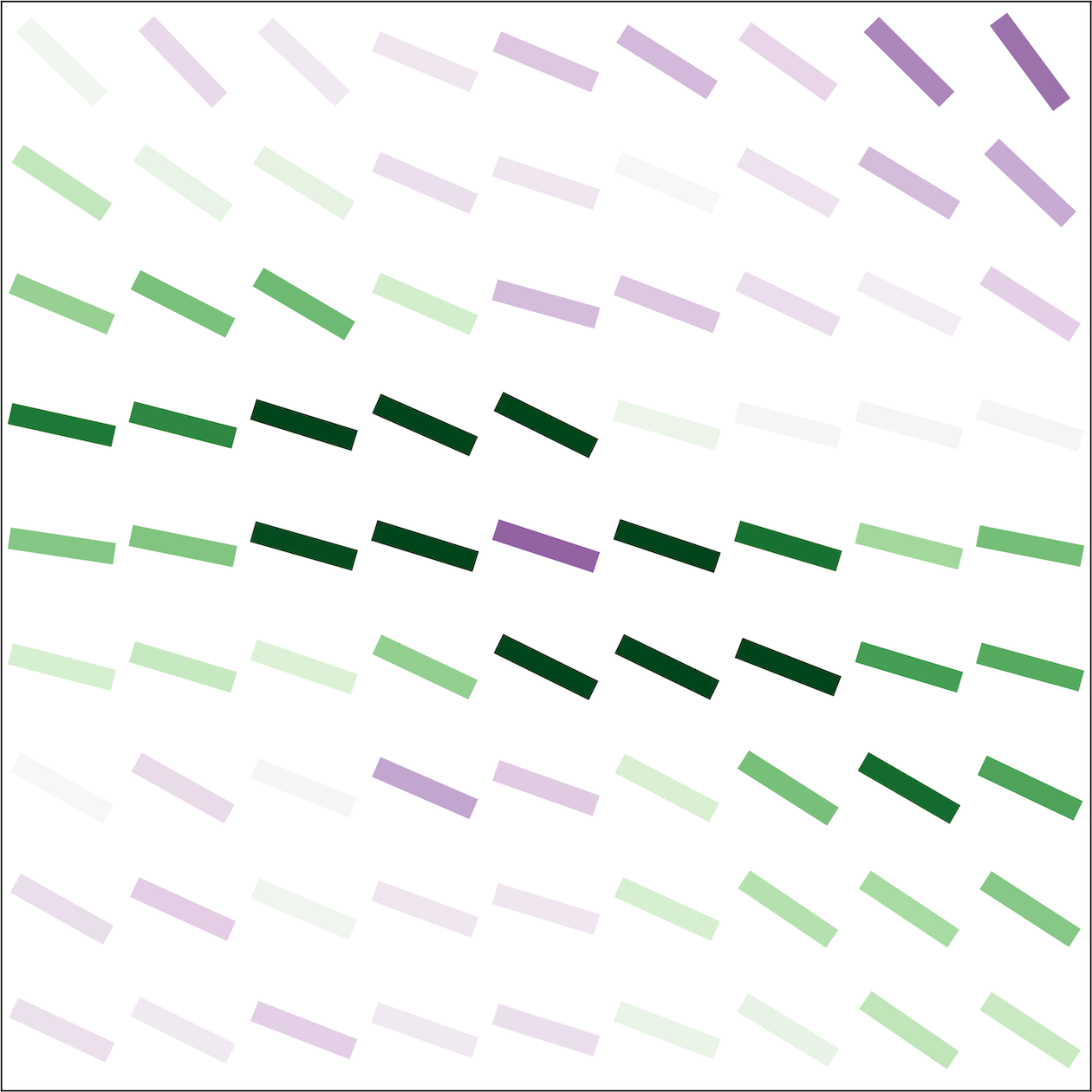}};
\draw [anchor=north west] (0.575\linewidth, 0.985\linewidth) node {\includegraphics[width=0.05\linewidth]{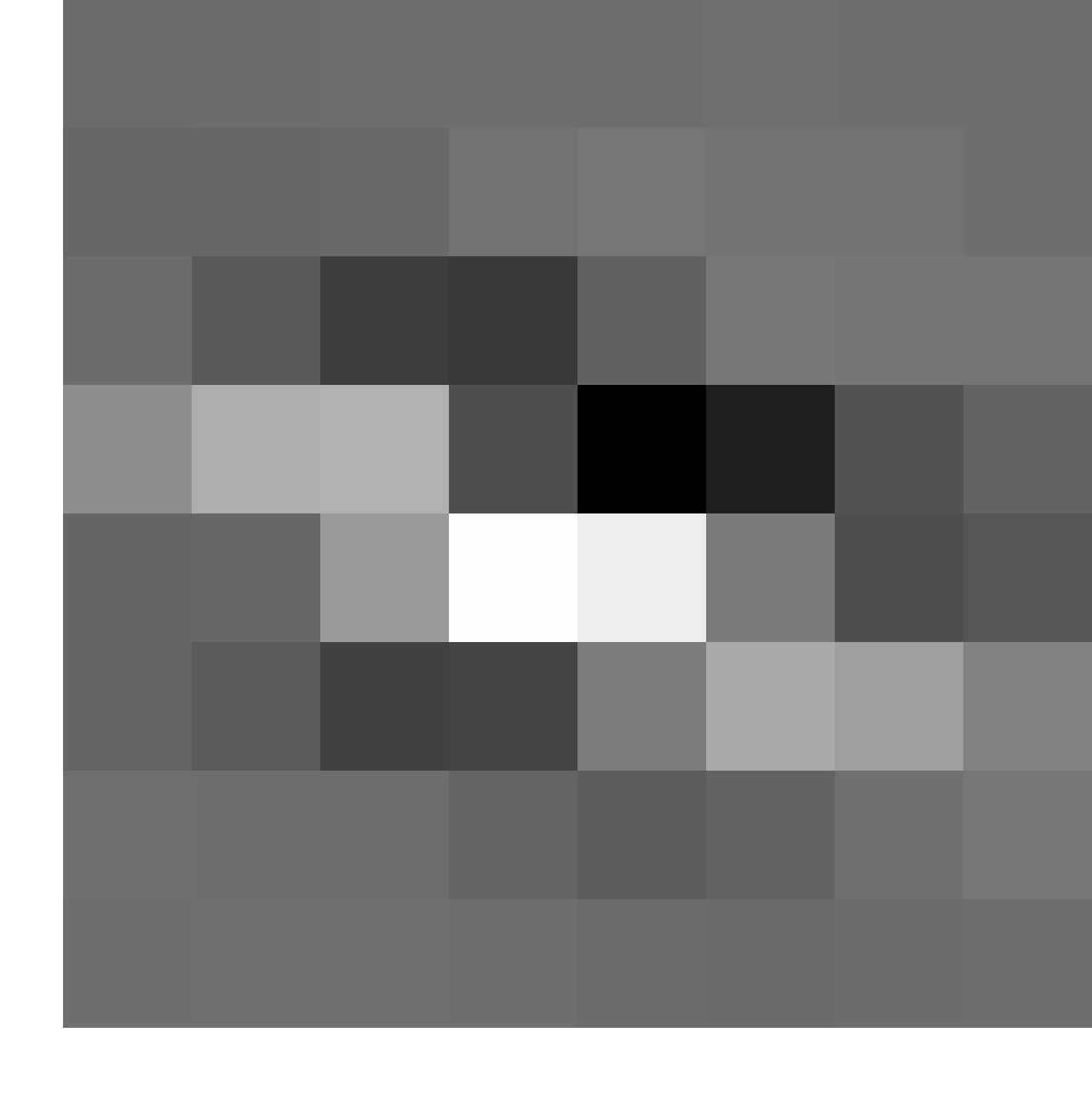} };

\draw [anchor=north west] (0.66\linewidth, 0.99\linewidth) node {\includegraphics[width=0.30\linewidth]{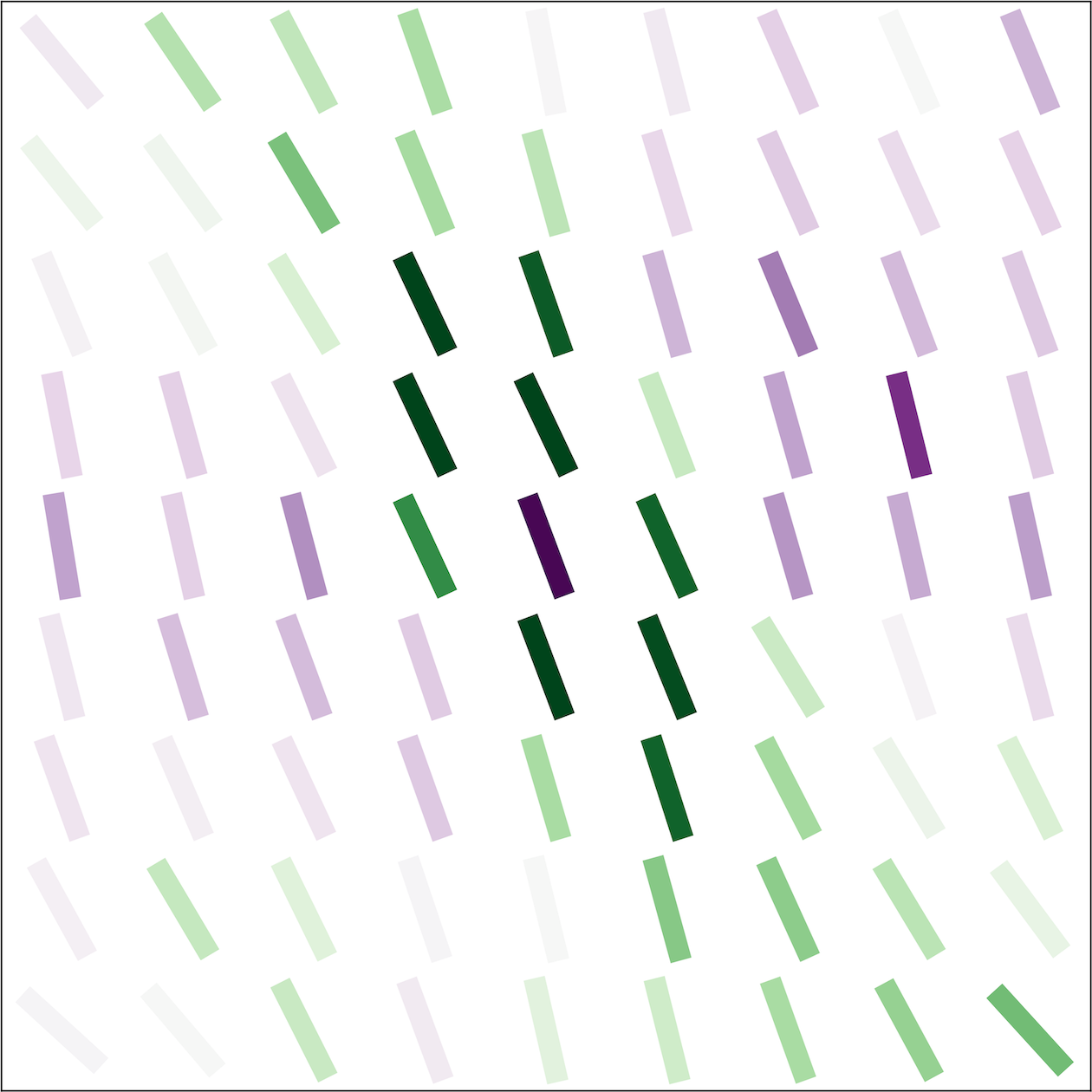}};
\draw [anchor=north west] (0.905\linewidth, 0.985\linewidth) node {\includegraphics[width=0.05\linewidth]{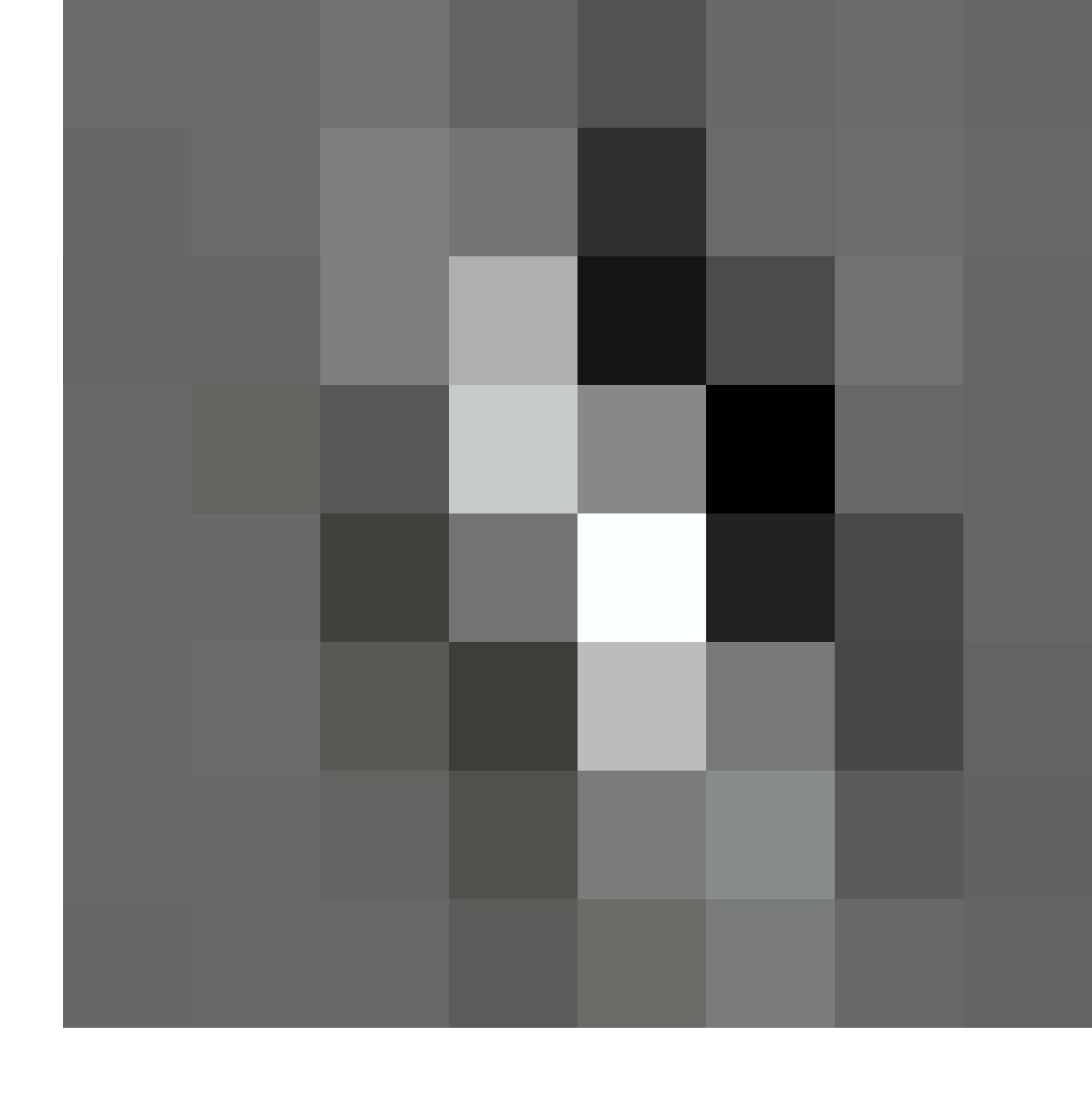} };

\draw [anchor=north west] (0.\linewidth, 0.64\linewidth) node {\includegraphics[width=0.30\linewidth]{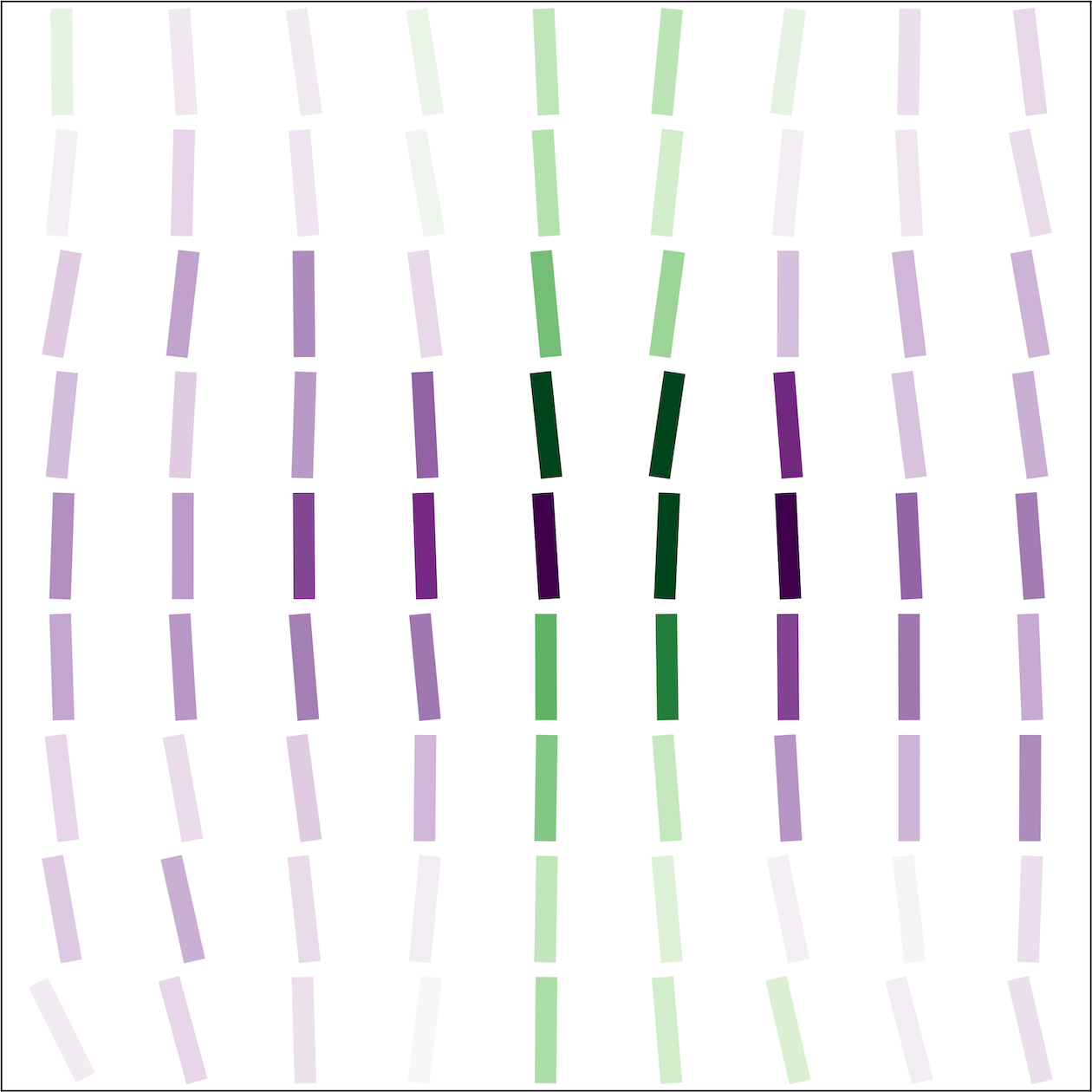}};
\draw [anchor=north west] (0.245\linewidth, 0.635\linewidth) node {\includegraphics[width=0.05\linewidth]{Figure/FigSD3/Filter=-90.png} };

\draw [anchor=north west] (0.33\linewidth, 0.64\linewidth) node {\includegraphics[width=0.30\linewidth]{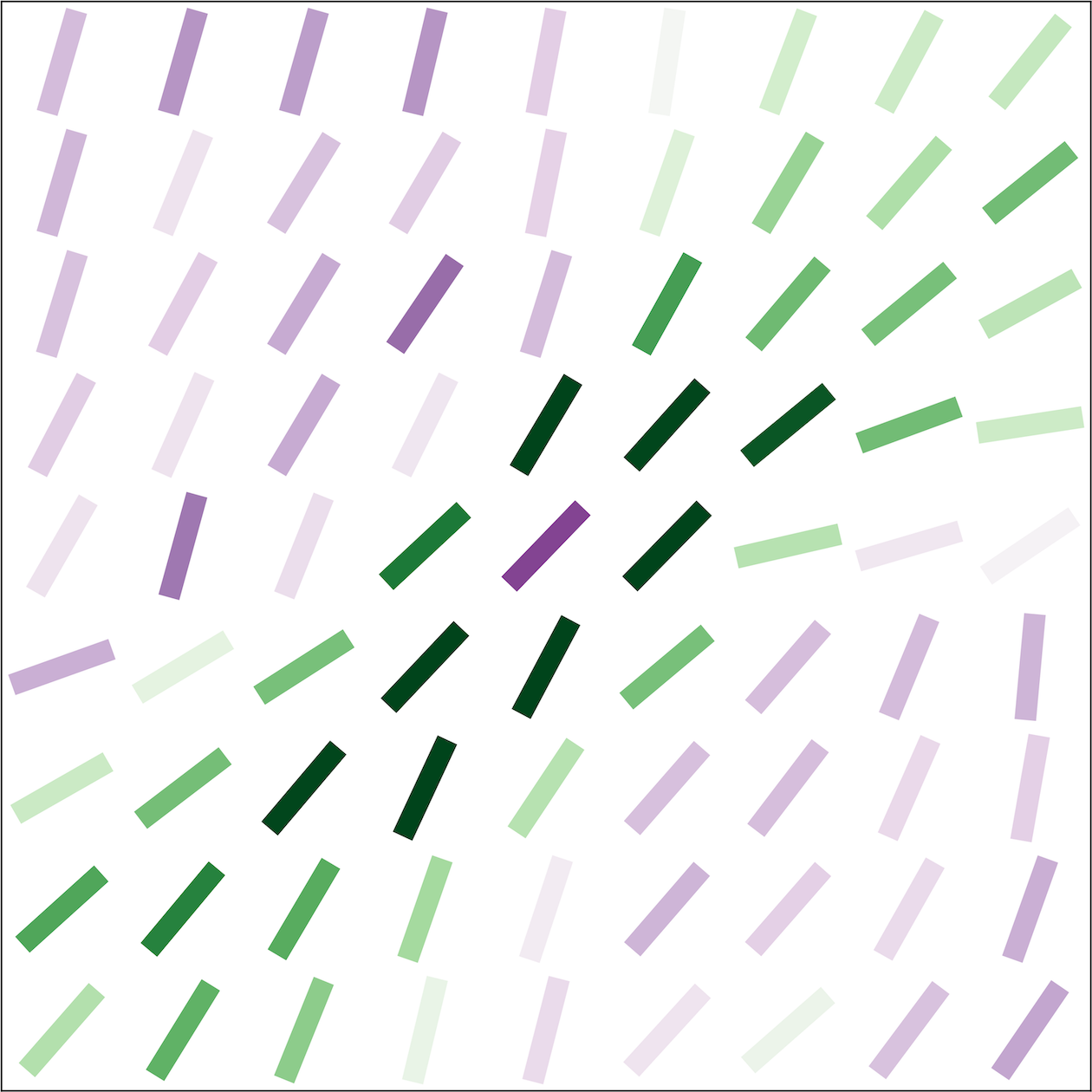}};
\draw [anchor=north west] (0.575\linewidth, 0.635\linewidth) node {\includegraphics[width=0.05\linewidth]{Figure/FigSD3/Filter=43.png} };

\draw [anchor=north west] (0.66\linewidth, 0.64\linewidth) node {\includegraphics[width=0.30\linewidth]{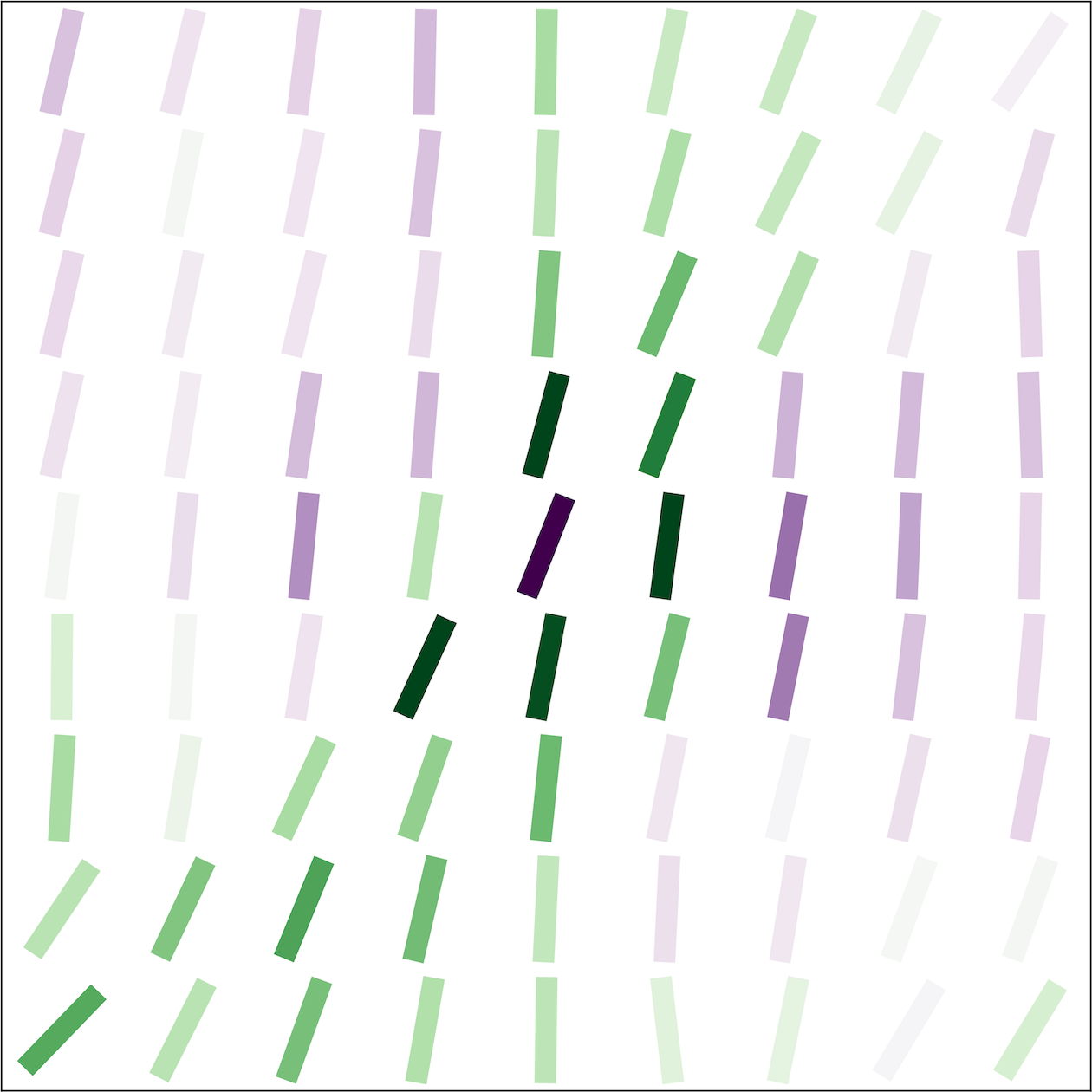}};
\draw [anchor=north west] (0.905\linewidth, 0.635\linewidth) node {\includegraphics[width=0.05\linewidth]{Figure/FigSD3/Filter=67.png} };

\draw [anchor=north west] (0.15\linewidth, 0.30\linewidth) node  {\includegraphics[width=0.7\linewidth]{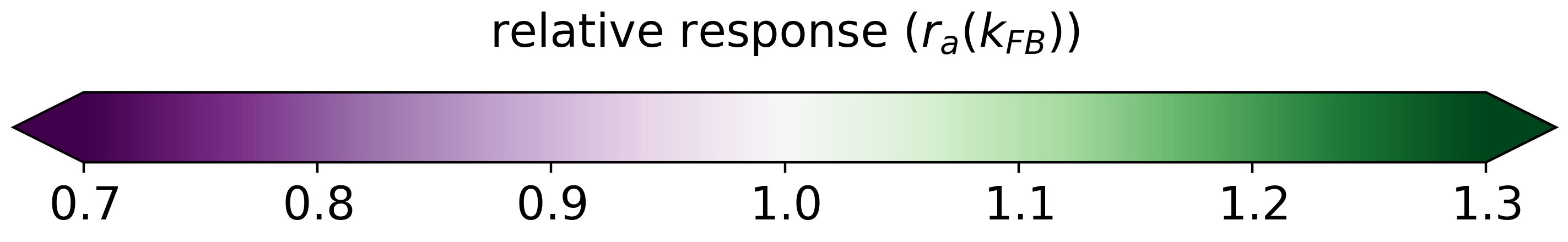}};
\begin{scope}
\draw [anchor=west,fill=white] (0.0\linewidth, 1\linewidth) node {{\bf A}};
\draw [anchor=west,fill=white] (0.33\linewidth, 1\linewidth) node {{\bf B}};
\draw [anchor=west,fill=white] (0.66\linewidth, 1\linewidth) node {{\bf C}};
\draw [anchor=west,fill=white] (0\linewidth, 0.65\linewidth) node {{\bf D}};
\draw [anchor=west,fill=white] (0.33\linewidth, 0.65\linewidth) node {{\bf E}};
\draw [anchor=west,fill=white] (0.66\linewidth, 0.65\linewidth) node {{\bf F}};

\end{scope}
\end{tikzpicture}
\caption
{{\bf Example of a 9$\times$9 association field in V1 colored with relative response w.r.t no feedback, centered on neurons strongly responding to $6$ different contour orientations, when the SDPC is trained on STL-10 database}. The feedback strength is set to $1$. From left to right and top to bottom the contour orientations are \ang{0} {\bf (A)}, \ang{-30} {\bf (B)}, \ang{-60} {\bf (C)}, \ang{90} {\bf (D)}, \ang{60} {\bf (E)} and \ang{30} {\bf (F)}.  At each location identified by the coordinates $(x_{c},y_{c})$ the angle is $\boldsymbol{\bar{\theta}}[x_{c},y_{c}]$ (see Eq.~\ref{eq:eq6}) and the color scale is proportional to $\boldsymbol{r_{a}}(k_\mathrm{FB})$ (see Eq.~\ref{eq:eq12}). The color scale being saturated toward both maximum and minimum activity, all the activities above $1.3$ or below $0.5$ have the same dark green or dark purple color, respectively.}
\label{fig:figSD4}
\end{figure}

\begin{figure}[h]
\begin{tikzpicture}
 \centering
 \draw [anchor=north west] (0.0\linewidth, 0.99\linewidth) node {\includegraphics[width=0.45\linewidth]{Figure/Fig9/STL_SSIM_L1.png}};
\draw [anchor=north west] (0.50\linewidth, 0.99\linewidth) node {\includegraphics[width=0.45\linewidth]{Figure/Fig9/STL_SSIM_L2.png} };
 \draw [anchor=north west] (0.02\linewidth, 0.65\linewidth) node {\includegraphics[width=0.45\linewidth]{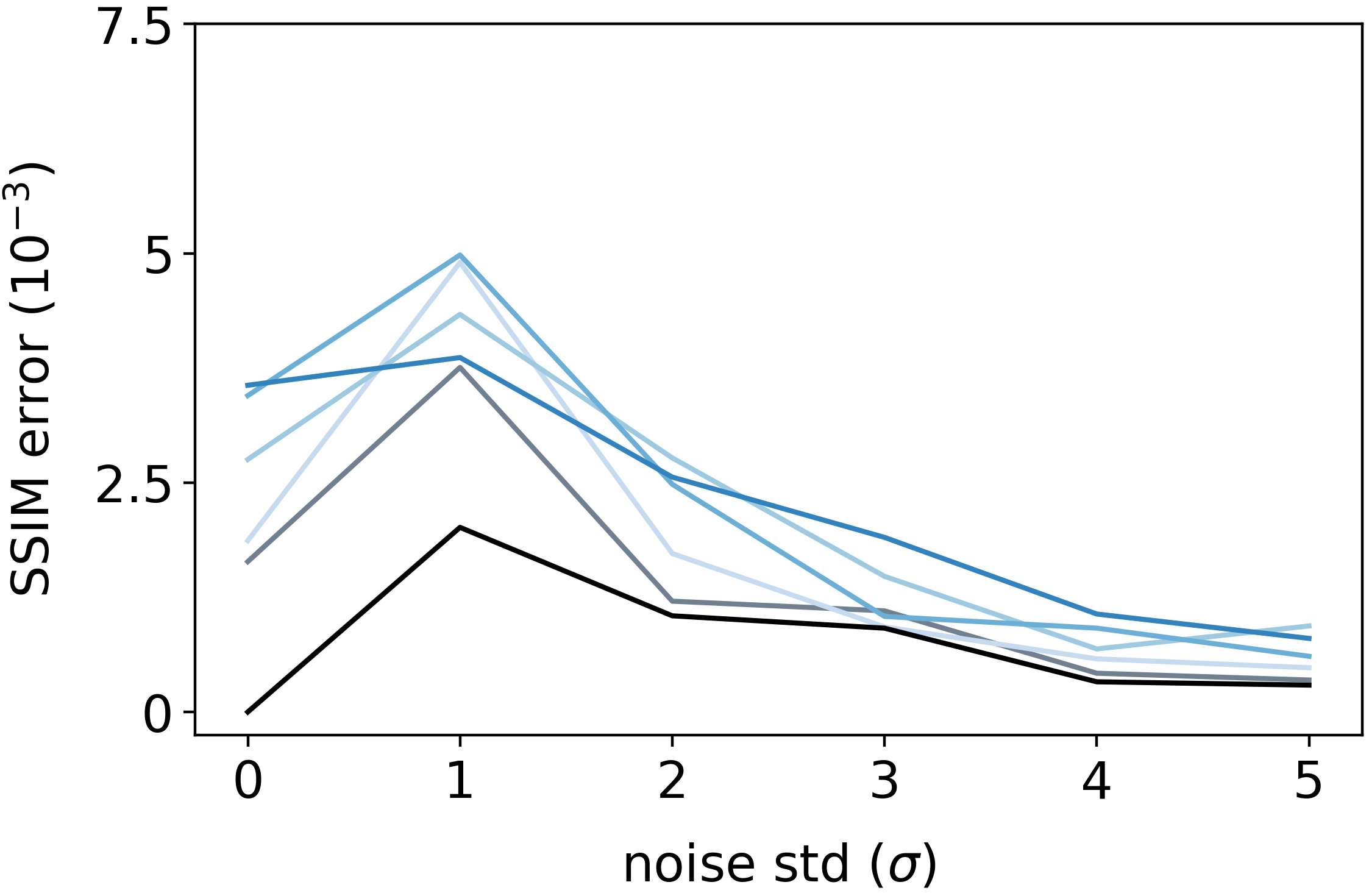}};
\draw [anchor=north west] (0.53\linewidth, 0.65\linewidth) node {\includegraphics[width=0.45\linewidth]{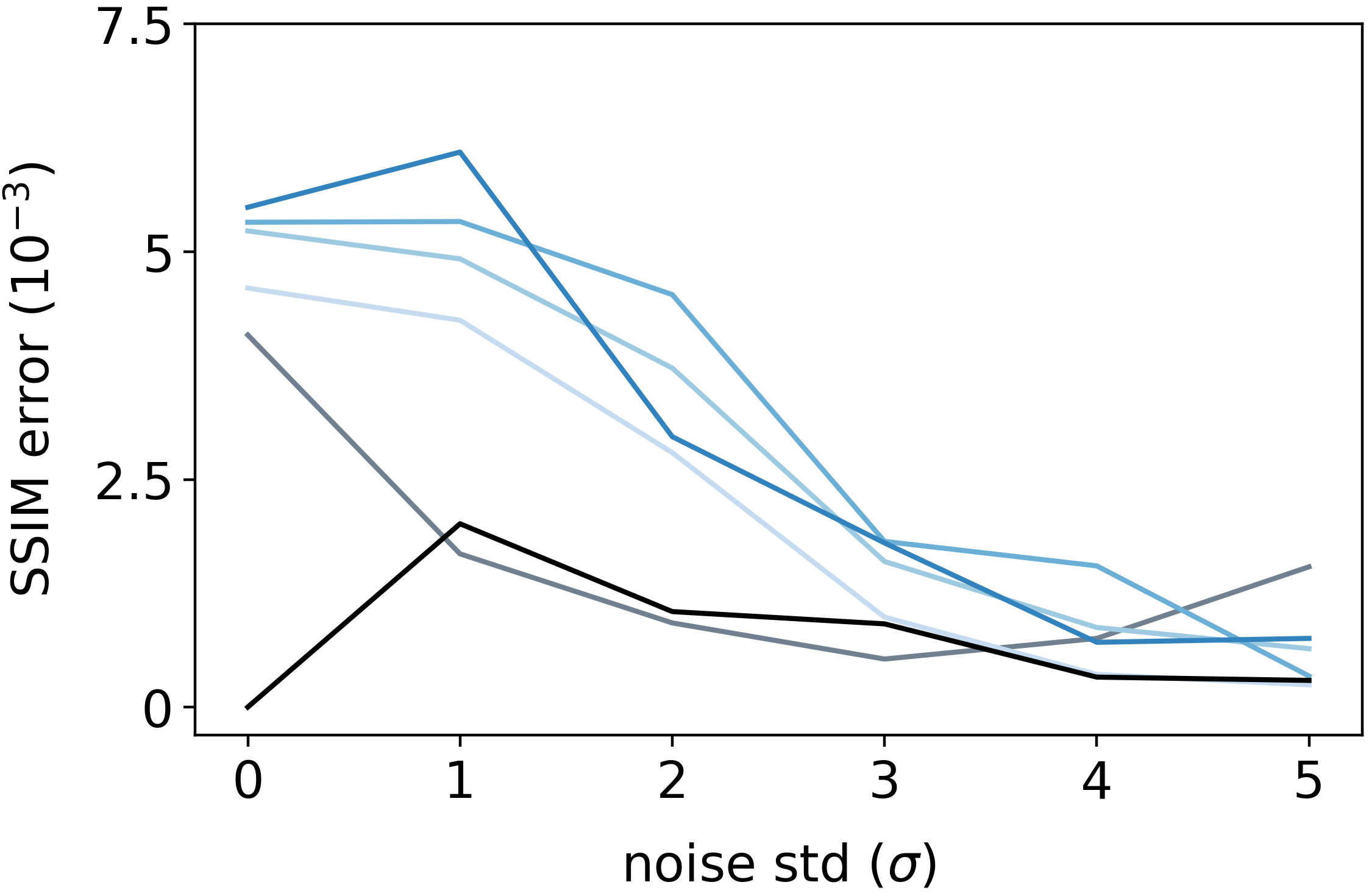} };

\draw [anchor=north west] (0.08\linewidth, 0.32\linewidth) node {\includegraphics[width=0.88\linewidth]{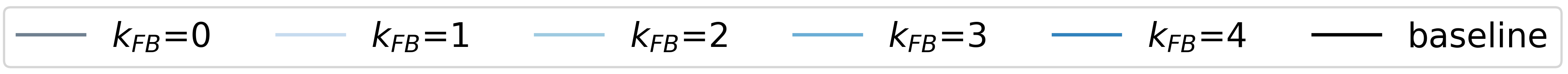}};
\begin{scope}
\draw [anchor=west,fill=white] (0.05\linewidth, 1\linewidth) node {{\bf A}};
\draw [anchor=west,fill=white] (0.55\linewidth, 1\linewidth) node {{\bf B}};
\draw [anchor=west,fill=white] (0.05\linewidth, 0.66\linewidth) node {{\bf C}};
\draw [anchor=west,fill=white] (0.55\linewidth, 0.66\linewidth) node {{\bf D}};

\end{scope}
\end{tikzpicture}
\caption
{{\bf Median Structural Similarity Index (SSIM) and corresponding Median Absolute Deviation (MAD) for degraded images of the STL-10 database}. {\bf(A)} median SSIM index between $1200$ original images and their reconstructions by the first layer of the SDPC. {\bf(B)} SSIM index between original images and their reconstructions by the second layer of the SDPC. {\bf(C)} Error, as computed with the MAD, of the SSIM plotted in {\bf(A)}. {\bf(D)} Error, as computed with the MAD, of the SSIM plotted in {\bf(B)}. The color code corresponds to the feedback strength, from light grey for $k_\mathrm{FB}=0$ to darker blue for higher feedback strength. The black line is the baseline, it is the SSIM between noisy and original input image.}

\label{fig:figSD5}
\end{figure}

\begin{figure}[h]
\begin{tikzpicture}
 \centering
 \draw [anchor=north west] (0.0\linewidth, 0.99\linewidth) node {\includegraphics[width=0.45\linewidth]{Figure/Fig10/CFD_SSIM_L1.png}};
\draw [anchor=north west] (0.50\linewidth, 0.99\linewidth) node {\includegraphics[width=0.45\linewidth]{Figure/Fig10/CFD_SSIM_L2.png} };
 \draw [anchor=north west] (0.02\linewidth, 0.65\linewidth) node {\includegraphics[width=0.45\linewidth]{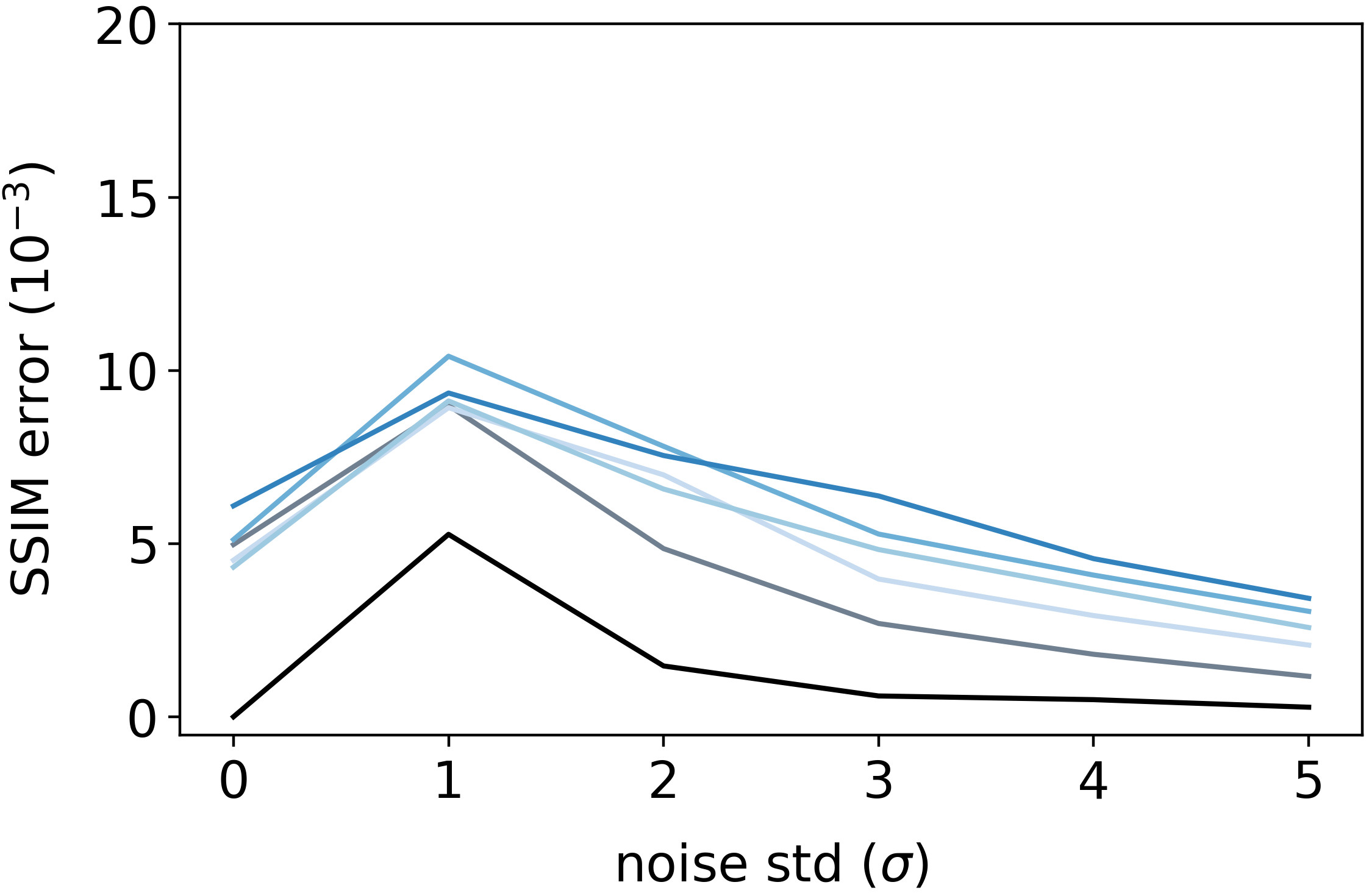}};
\draw [anchor=north west] (0.53\linewidth, 0.65\linewidth) node {\includegraphics[width=0.45\linewidth]{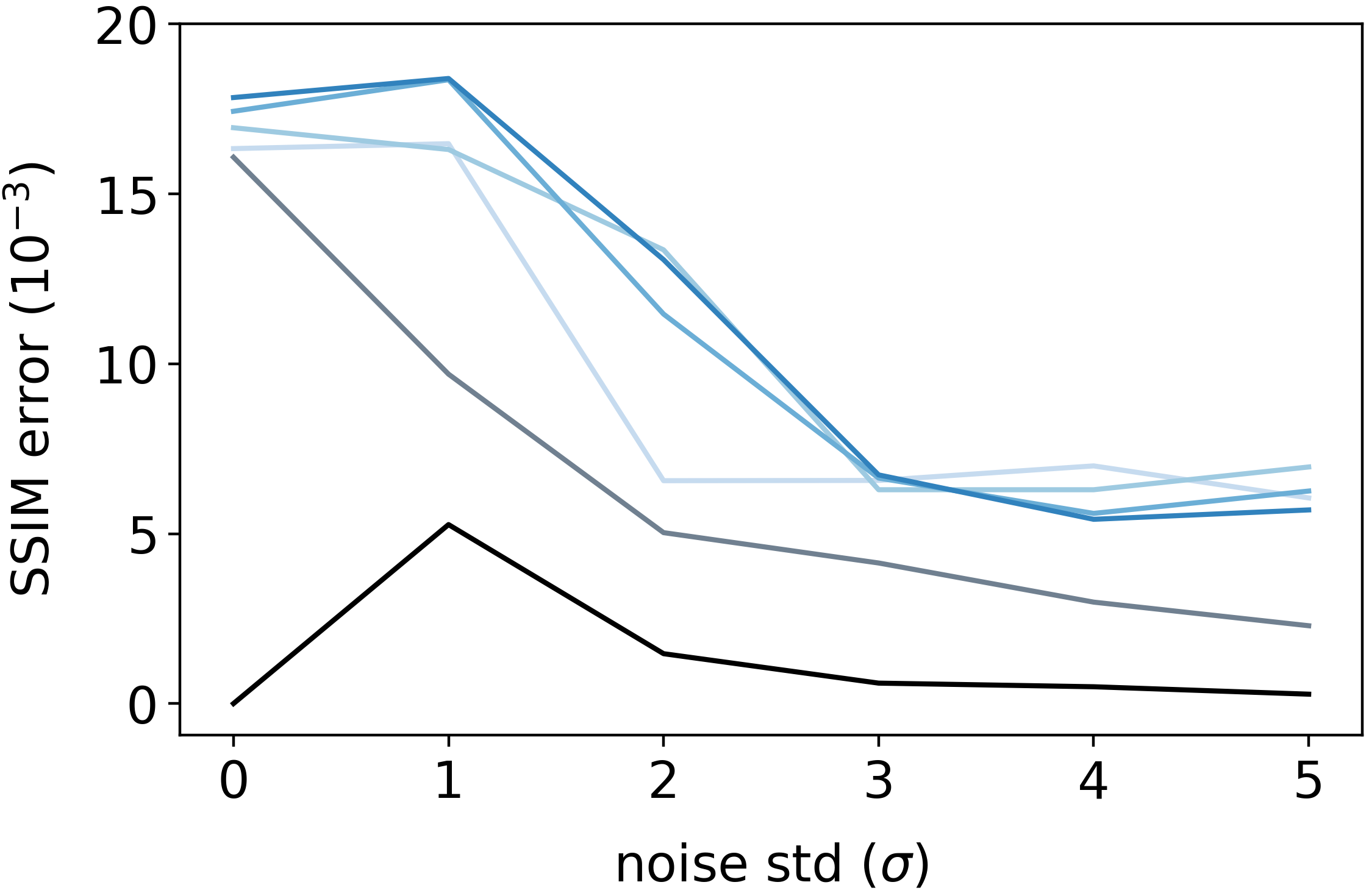} };

\draw [anchor=north west] (0.08\linewidth, 0.32\linewidth) node {\includegraphics[width=0.88\linewidth]{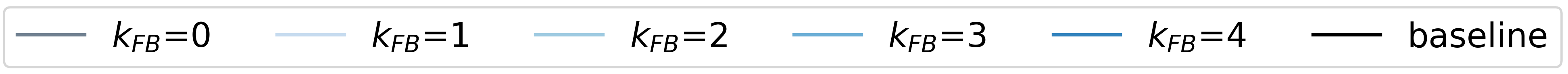}};
\begin{scope}
\draw [anchor=west,fill=white] (0.05\linewidth, 1\linewidth) node {{\bf A}};
\draw [anchor=west,fill=white] (0.55\linewidth, 1\linewidth) node {{\bf B}};
\draw [anchor=west,fill=white] (0.05\linewidth, 0.66\linewidth) node {{\bf C}};
\draw [anchor=west,fill=white] (0.55\linewidth, 0.66\linewidth) node {{\bf D}};

\end{scope}
\end{tikzpicture}
\caption
{{\bf Median Structural Similarity Index (SSIM) and corresponding Median Absolute Deviation (MAD) for degraded images of the CFD database}. {\bf(A)} median SSIM index between $400$ original images and their reconstructions by the first layer of the SDPC. {\bf(B)} SSIM index between original images and their reconstructions by the second layer of the SDPC. {\bf(C)} Error, as computed with the MAD, of the SSIM plotted in {\bf(A)}. {\bf(D)} Error, as computed with the MAD, of the SSIM plotted in {\bf(B)}. The color code corresponds to the feedback strength, from light grey for $k_\mathrm{FB}=0$ to darker blue for higher feedback strength. The black line is the baseline, it is the SSIM between noisy and original input image.}
\label{fig:figSD6}
\end{figure}

\begin{algorithm}[!h]
\caption{SDPC inference algorithm}
\SetKwProg{Fn}{}{\string:}{}
\DontPrintSemicolon
\SetKwInOut{Input}{input}\SetKwInOut{Output}{output}

\Input{~image: $\boldsymbol{x}$, dictionaries: $\{\boldsymbol{D}_{i}\}_{i=1}^{L}$, $\ell_{1}$-penalty parameters: $\{\lambda_{i}\}_{i=1}^{L}$, stability threshold: $T_{stab}$
}
\;
$\boldsymbol{\gamma}_{0}^{t}$ = $\boldsymbol{x}$\;
$\{\boldsymbol{\gamma}_{i}^{0}\}_{i=1}^{L} = \boldsymbol{0}, \mQuad \{\boldsymbol{\gamma}_{m_{i}}^{1}\}_{i=1}^{L} = \boldsymbol{0}$
\textcolor{gray}{\# Initializing layer state variables and FISTA momentum} \;
$\alpha^{1}=1$ \textcolor{gray}{\# Initializing momentum strength}\;

$\eta_{c_{i}}= \displaystyle \frac{1}{max(eigen\_value(\boldsymbol{D}^{T}_{i}\boldsymbol{D}_{i}))}$\;

$Stable = False$ \textcolor{gray}{\# Initializing the stability criterion}\;
$t=0$\;
\;

\While{$Stable == False$ } {
    $t \mathrel{+}= 1$\;
    $\alpha^{t+1} = \displaystyle \frac{1 + \sqrt{1+4(\alpha^{t})^{2}}}{2} $\;

    \For{$i= 1$ \KwTo $L$}{
        \textcolor{gray}{\# Update lower-layer error} \;
        $\boldsymbol{\epsilon}_{LL}$ = $\boldsymbol{\gamma}_{m_{i-1}}^{t} - \boldsymbol{D}_{i}^{T} \boldsymbol{\gamma}_{m_{i}}^{t}$ \;
        \textcolor{gray}{\# Update the upper-layer error } \;
        \If{$i \neq L$}{
            $\boldsymbol{\epsilon}_{UL}$ = $\boldsymbol{\gamma}_{m_{i}}^{t} - \boldsymbol{D}_{i+1}^{T} \boldsymbol{\gamma}_{m_{i+1}}^{t}$ \;
            }
        \Else{
            $\boldsymbol{\epsilon}_{UL} = \boldsymbol{0}$  \;}
       $\boldsymbol{\gamma}_{i}^{t}=\mathcal{T}_{\eta_{c_{i}} \lambda_i}\big( \boldsymbol{\gamma}_{m_{i}}^{t}+ \eta_{c_{i}}\boldsymbol{D}_{i}\boldsymbol{\epsilon}_{LL} - \displaystyle \eta_{c_{i}}\boldsymbol{\epsilon}_{UL}\big) $  \textcolor{gray}{\# Update layer state variables} \;

        $\boldsymbol{\gamma}_{m_{i}}^{t+1} = \mathcal{T}_{0}\Big(\boldsymbol{\gamma}_{i}^{t} + \displaystyle \big( \frac{\alpha^{t} - 1}{\alpha^{t+1}} \big)\big( \boldsymbol{\gamma}_{i}^{t} - \boldsymbol{\gamma}_{i}^{t-1} \big)\Big)$ \textcolor{gray}{\# Update the FISTA momentum}\;
    }

    \If{$\displaystyle \bigwedge\limits_{i=1}^L \Big( \frac{\Vert \boldsymbol{\gamma}_{i}^t - \boldsymbol{\gamma}_{i}^{t-1}\Vert_{2}}{\Vert\boldsymbol{\gamma}_{i}^t\Vert_{2}} < T_{stab} \Big)$}{
        \vspace*{2mm}
        $Stable = True$ \textcolor{gray}{\# Update the stability creterion}\;
    }
}
\KwRet{$\{\boldsymbol{\gamma}_{i}^{t} \}_{i=1}^{L}$}
\algorithmfootnote{\textbf{Note:} $\mathcal{T}_{\alpha} (\cdot)$ denotes the element-wise non-negative soft-thresholding operator. A fortiori, $\mathcal{T}_{0}(\cdot)$ is a rectified linear unit operator. \textcolor{gray}{\# comments} are comments.}
\label{algo:pseudo_code}
\end{algorithm}

\begin{figure}[h]
\begin{tikzpicture}
	\centering
\draw [anchor=north west] (0.05\linewidth, 1\linewidth) node {\includegraphics[width=1\linewidth]{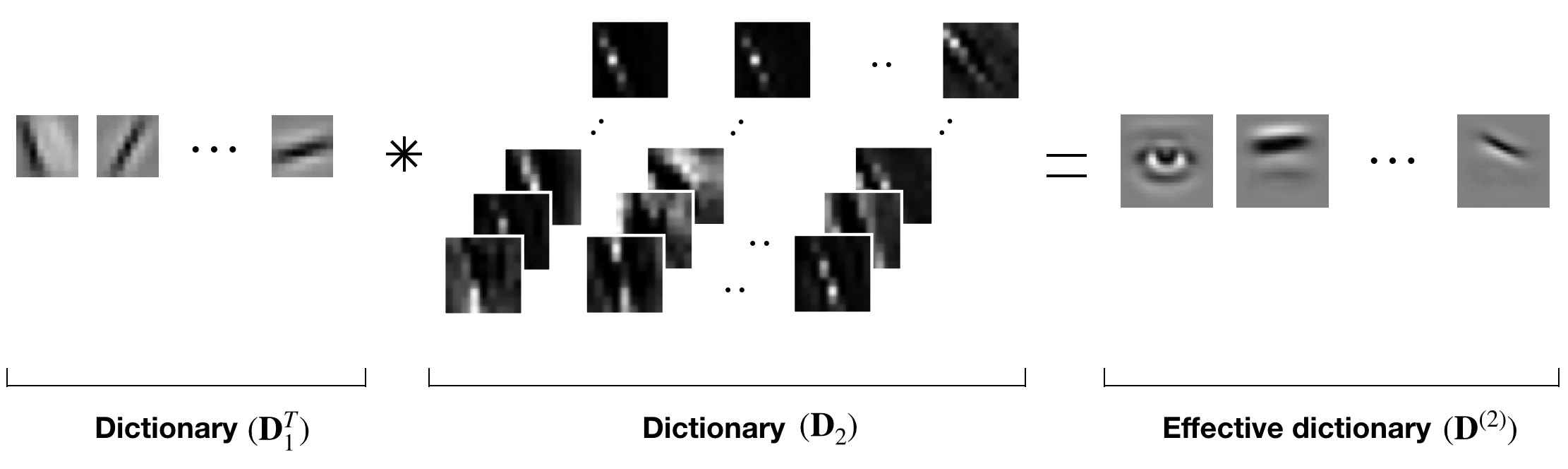}};
\end{tikzpicture}
\caption{{\bf Illustration of the back-projection mechanism.} The projection of the second layer dictionary into the visual space ($\mathbf{D}^{\mathrm{eff}^{T}}_{2}$) is obtained by convolving the transpose of the first layer dictionary ($\mathbf{D}^{{T}}_{1}$) by the second layer dictionary ($\mathbf{D}^{{T}}_{2}$)\cite{Sulam2017}. This mechanism could be also used to back-project any activity map into the visual space (see Eq.~\ref{eq:Reconstruction}).}
\label{fig:fig_backprojection}
\end{figure}

\end{document}